%% file: Thesis.tex
\documentclass[letter,12pt,oneside]{book}
\usepackage{graphicx}

\usepackage[letterpaper, margin=1in]{geometry}

\usepackage{amsmath}
\usepackage{amssymb}
\usepackage{mathtools}
\usepackage{amsthm}
\usepackage{amsfonts}       
\usepackage{bbm}
\usepackage{makecell}
\usepackage{booktabs}
\usepackage{multirow}

\usepackage{subfig}
\usepackage[numbers,compress,sort]{natbib}	

\usepackage{multirow}
\usepackage{multicol}
\usepackage{nicefrac}       
\usepackage{microtype}
\usepackage{booktabs} 
\usepackage{enumitem}

\usepackage{algorithmic}
\usepackage{algorithm}

\usepackage{setspace}
\usepackage{tikz}
\usepackage{wrapfig}
\usepackage[acronym]{glossaries}
\usepackage{subcaption}
\makeglossaries

\usepackage{fix-cm}
\usepackage{xcolor}
\usepackage{xspace}
\usepackage[bf,rm,medium,compact]{titlesec}

\titleformat{\chapter}[display]{\normalfont\Huge\bfseries}{\fontsize{28}{28}\selectfont\textcolor{black}{Chapter \thechapter}}{1ex}{}[]%

\usepackage{blindtext}
\usepackage[hidelinks]{hyperref}

\theoremstyle{plain}
\newtheorem{theorem}{Theorem}[section]

\theoremstyle{definition}
\newtheorem{definition}[theorem]{Definition}

\theoremstyle{remark}

\usepackage{pifont}
\title{Phd Thesis}
\author{Qincheng Lu}
\date{\today}
\input{macros.tex}

\begin{document}
\doublespacing

\newcommand{\mcgillguidelines}{\href{https://www.mcgill.ca/gps/thesis/thesis-guidelines/preparation}{Official McGill Guidelines: }}

\begin{onehalfspacing}
\pagestyle{empty}
\input{content/TitlePage.tex}

\clearpage
\end{onehalfspacing}

\pagenumbering{roman}
\pagestyle{plain}

\input{content/AbstractEnglish}

\input{content/AbstractFrench}

\input{content/Acknowledgements}

\input{content/GenAI_statement}

\tableofcontents

\addcontentsline{toc}{section}{List of Figures}
\listoffigures
\addcontentsline{toc}{section}{List of Tables}
\listoftables
\clearpage

\pagenumbering{arabic}

\input{content/Introduction}
\input{content/Preliminary}

\input{content/RelatedWork}

\input{content/Paper02/Paper02}

\input{content/Paper01/Paper01}

\input{content/Paper03/Paper03}
\input{content/Paper04/Paper04}


\input{content/Conclusions}


\bibliographystyle{ieeetr}
\bibliography{references}

\end{document}

%% file: macros.tex
\usepackage{enumitem}
\allowdisplaybreaks

\newlength{\nunitlength}
\newcommand{\be}{\begin{equation}}
\newcommand{\ee}{\end{equation}}
\newcommand{\beqy}{\begin{eqnarray}}
\newcommand{\eeqy}{\end{eqnarray}}
\newcommand{\beqynn}{\begin{eqnarray*}}
\newcommand{\eeqynn}{\end{eqnarray*}}

\newcommand{\ba}{\begin{array}}
\newcommand{\ea}{\end{array}}
\newcommand{\bmx}{\begin{bmatrix}}
\newcommand{\emx}{\end{bmatrix}}
\newcommand{\bsmx}{\left[\begin{smallmatrix}}
\newcommand{\esmx}{\end{smallmatrix}\right]}
\newcommand{\bmxc}[1]{\left[\begin{array}{@{}#1@{}}}
\newcommand{\emxc}{\end{array}\right]}

\newcommand{\bt}[1]{\begin{tabular}{#1}}
\newcommand{\et}{\end{tabular}}

\newcommand{\bc}{\begin{center}}
\newcommand{\ec}{\end{center}}
\newcommand{\ben}{\begin{enumerate}}
\newcommand{\een}{\end{enumerate}}
\newcommand{\bi}{\begin{itemize}}
\newcommand{\ei}{\end{itemize}}

\newcommand{\Exp}[1]{\mathrm{E}\!\left\{#1 \right\}}

\newcommand{\sign}{\mathrm{sgn}}

\newcommand{\red}{\textcolor{red}}
\newcommand{\blue}{\textcolor{blue}}

\newcommand{\rw}{{\mathrm{rw}}}

\newcommand{\sym}{{\mathrm{sym}}}

\renewcommand{\top}{{\mkern-1.5mu\mathsf{T}}}

\newcommand{\diag}{\mathrm{diag}}

\newcommand{\ReLU}{\mathrm{ReLU}}
\newcommand{\LeakyReLU}{\mathrm{LeakyReLU}}
\newcommand{\softmax}{\mathrm{softmax}}
\newcommand{\erf}{\mathrm{erf}}

\DeclareMathOperator*{\argmax}{argmax}

\newcommand{\Rbb}{{\mathbb{R}}}
\newcommand{\Zbb}{{\mathbb{Z}}}
\newcommand{\Cbb}{{\mathbb{C}}}

\newcommand{\RN}{\Rbb^{N}}

\newcommand{\Rnbn}{\Rbb^{n \times n}}
\newcommand{\RNbN}{\Rbb^{N \times N}}

\newcommand{\Rmbn}{\Rbb^{m \times n}}

\newcommand{\calC}{\mathcal{C}}

\newcommand{\calN}{\mathcal{N}}

\newcommand{\calV}{\mathcal{V}}
\newcommand{\calG}{\mathcal{G}}
\newcommand{\calE}{\mathcal{E}}
\newcommand{\calX}{\mathcal{X}}

\newcommand{\A}{\mathbf{A}}
\newcommand{\B}{\mathbf{B}}

\newcommand{\D}{\mathbf{D}}

\renewcommand{\H}{\mathbf{H}}
\newcommand{\I}{\mathbf{I}}

\newcommand{\LL}{\mathbf{L}}

\newcommand{\N}{\mathbf{N}}
\renewcommand{\P}{\mathbf{P}}
\newcommand{\Q}{\mathbf{Q}}
\newcommand{\R}{\mathbf{R}}

\newcommand{\T}{\mathbf{T}}
\newcommand{\U}{\mathbf{U}}

\newcommand{\W}{\mathbf{W}}
\newcommand{\X}{{\mathbf{X}}}
\newcommand{\Y}{{\mathbf{Y}}}
\newcommand{\Z}{\mathbf{Z}}

\renewcommand{\a}{\mathbf{a}}
\renewcommand{\b}{\mathbf{b}}

\newcommand{\e}{\mathbf{e}}
\newcommand{\f}{\mathbf{f}}
\newcommand{\g}{\mathbf{g}}
\newcommand{\h}{\mathbf{h}}

\newcommand{\n}{\mathbf{n}}

\newcommand{\q}{\mathbf{q}}

\newcommand{\s}{\mathbf{s}}

\renewcommand{\u}{\mathbf{u}}
\newcommand{\noisevector}{{\mathbf{v}}}

\newcommand{\x}{{\mathbf{x}}}
\newcommand{\y}{{\mathbf{y}}}
\newcommand{\z}{{\mathbf{z}}}

\newcommand{\1}{{\mathbf{1}}}

\newcommand{\bmu}{\boldsymbol{\mu}}

\newcommand{\bphi}{\boldsymbol{\phi}}

\newcommand{\bpsi}{\boldsymbol{\psi}}
\newcommand{\btheta}{\boldsymbol{\theta}}
\newcommand{\bgamma}{\boldsymbol{\gamma}}
\newcommand{\blambda}{\boldsymbol{\lambda}}
\newcommand{\bLambda}{{\boldsymbol{\Lambda}}}
\newcommand{\bSigma}{{\boldsymbol{\Sigma}}}

\newcommand{\bPhi}{\boldsymbol{\Phi}}
\newcommand{\bPsi}{\boldsymbol{\Psi}}
\newcommand{\bTheta}{\boldsymbol{\Theta}}

\newcommand{\0}{{\mathbf{0}}}

\newcommand{\bbx}{{\bar{\x}}}

\newcommand{\hbx}{{\hat{\x}}}

\newcommand{\tbQ}{{\widetilde{\Q}}}

\newcommand{\tbq}{{\tilde{\q}}}

\newcommand{\tby}{{\tilde{\y}}}

\newcommand{\hbQ}{{\widehat{\Q}}}
\newcommand{\hbq}{{\widehat{\q}}}

\newcommand\ie{\textit{i.e.,}}

\newcommand{\norm}[1]{\left\lVert#1\right\rVert}

%% file: content/TitlePage.tex


\begin{titlepage}
\begin{center}

\vspace*{0.5cm}

{\bfseries\LARGE Graph Representation Learning and Its }
\vspace{0.15cm}

{\bfseries\LARGE Applications to Integer-valued Parameter}
\vspace{0.15cm}

{\bfseries\LARGE  Estimation in
Linear Models}
\vspace{0.15cm}


{\large Qincheng Lu}
\\Doctor of Philosophy

\vspace{1cm}
\begin{figure}[ht!]
    \centering
    \includegraphics[width=.6\linewidth]{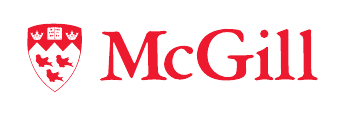}
  \end{figure}
School of Computer Science\\
McGill University\\
Montreal, Quebec, Canada\\

\vspace{1.5cm}
\today

\vspace{1.0cm}
\noindent
A thesis submitted to McGill University in partial\\
fulfillment of the requirements of the degree of\\
Doctor of Philosophy

\vspace{1.0cm}
{\small \copyright Qincheng Lu, \the\year{}}

\end{center}
\end{titlepage}

%% file: content/AbstractEnglish.tex
\chapter*{\rm\bfseries Abstract}
\label{ch:abstraten}

Graph representation learning is a fundamental technique for modeling non-Euclidean graph data that capture complex relational structures.
By encoding graph data into vector representations, 
graph representation learning supports downstream tasks across a wide range of applications.
Beyond that,
recent research has shown that graph representation learning-based methods can approximate solutions to optimization problems that are too costly to solve exactly,
as those optimization problems have inherent graph structures.

Existing graph message passing,
a dominant paradigm for graph representation learning,
relies on local aggregation operators determined by the Laplacian matrix and typically fails to capture information from distant nodes.
This limitation is especially problematic in heterophilic graph learning,
where useful information often lies beyond immediate neighborhoods.
To address this problem,
this thesis proposes methods based on a new class of graph Laplacians that enable adaptive message passing across different levels of graph homophily.
The proposed methods outperform state-of-the-art (SOTA) baselines on six of seven benchmark datasets.

This thesis then develops learning to optimize methods for integer-valued parameter estimation in linear models
that arise in real-world applications,
such as multiple-input multiple-output (MIMO) communication systems,
GPS positioning,
and utilization control.
The first study extends the well-known Babai estimator to some $L_2$-norm regularized Babai estimators,
and proposes a learning-based method to maximize their success probabilities,
which is computationally infeasible to optimize by conventional means. 
We then propose novel estimators based on graph representation learning to leverage the dense graph structure induced by each instance of the estimation problem. 
In the second study, 
we show that the linear model admits a spectral graph convolutional form and propose GCEPNet, 
an efficient graph-based estimator. 
GCEPNet addresses the computational bottleneck caused by the graph neural network module in the existing SOTA estimator, 
thereby improving computational efficiency while matching or outperforming it.
The final study investigates diffusion-based estimation, 
a recently emerging and promising paradigm because a single trained denoising model can refine candidate solutions over a flexible number of inference steps, 
allowing a trade-off between accuracy and runtime at inference time. 
Existing diffusion-based estimators for linear models, 
however, still suffer from high latency and performance degradation when the linear model is underdetermined,
a harder case than the overdetermined case.
To address these limitations, 
this thesis proposes GDIM, 
a graph-based denoising diffusion estimator, 
which achieves a more favorable accuracy--runtime trade-off and is effective for both overdetermined and underdetermined settings.

%% file: content/AbstractFrench.tex
\chapter*{\rm\bfseries Abr\'eg\'e}
\label{ch:abstratfr}

L’apprentissage de représentations de graphes est une technique fondamentale pour modéliser des données de graphes non euclidiennes qui capturent des structures relationnelles complexes.
En encodant les données de graphes sous forme de représentations vectorielles,
l’apprentissage de représentations de graphes permet de réaliser des tâches en aval dans un large éventail d’applications.
Au-delà de cela,
des recherches récentes ont montré que les méthodes fondées sur l’apprentissage de représentations de graphes peuvent approximer les solutions de problèmes d’optimisation trop coûteux à résoudre exactement,
car ces problèmes d’optimisation possèdent des structures de graphes intrinsèques.

Le passage de messages sur graphes existant,
un paradigme dominant de l’apprentissage de représentations de graphes,
repose sur des opérateurs d’agrégation locaux déterminés par la matrice laplacienne et ne parvient généralement pas à capturer l’information provenant de nœuds distants.
Cette limitation est particulièrement problématique dans l’apprentissage sur graphes hétérophiles,
où l’information utile se trouve souvent au-delà des voisinages immédiats.
Pour répondre à ce problème,
cette thèse propose des méthodes fondées sur une nouvelle classe de laplaciens de graphes qui permettent un passage de messages adaptatif selon différents niveaux d’homophilie du graphe.
Les méthodes proposées surpassent les méthodes de référence de l’état de l’art (SOTA) sur six des sept jeux de données de référence.

Cette thèse développe ensuite des méthodes d’apprentissage pour l’optimisation appliquées à l’estimation de paramètres à valeurs entières dans des modèles linéaires
qui apparaissent dans des applications réelles,
telles que les systèmes de communication à entrées multiples et sorties multiples (MIMO),
le positionnement GPS
et le contrôle d’utilisation.
La première étude étend le célèbre estimateur de Babai à certains estimateurs de Babai régularisés en norme $L_2$
et propose une méthode fondée sur l’apprentissage pour maximiser leurs probabilités de succès,
ce qui est impossible à optimiser par des moyens conventionnels en raison du coût de calcul.
Nous proposons ensuite de nouveaux estimateurs fondés sur l’apprentissage de représentations de graphes afin d’exploiter la structure de graphe dense induite par chaque instance du problème d’estimation.
Dans la deuxième étude,
nous montrons que le modèle linéaire admet une forme de convolution spectrale sur graphe et proposons GCEPNet,
un estimateur efficace fondé sur les graphes.
GCEPNet résout le goulot d’étranglement computationnel causé par le module de réseau neuronal de graphes dans l’estimateur existant de l’état de l’art,
améliorant ainsi l’efficacité computationnelle tout en égalant ou en surpassant ses performances.
La dernière étude examine l’estimation fondée sur la diffusion,
un paradigme récent, émergent et prometteur, car un seul modèle de débruitage entraîné peut raffiner des solutions candidates avec un nombre flexible d’étapes d’inférence,
permettant ainsi un compromis entre précision et temps d’exécution au moment de l’inférence.
Cependant,
les estimateurs existants fondés sur la diffusion pour les modèles linéaires souffrent encore d’une latence élevée et d’une dégradation des performances lorsque le modèle linéaire est sous-déterminé,
un cas plus difficile que le cas surdéterminé.
Pour répondre à ces limitations,
cette thèse propose GDIM,
un estimateur de diffusion débruitante fondé sur les graphes,
qui atteint un compromis précision--temps d’exécution plus favorable et reste efficace dans les contextes surdéterminés comme sous-déterminés.

%% file: content/Acknowledgements.tex
\chapter*{\rm\bfseries Acknowledgements}
\label{ch:acknowledgement}

I would like to express my sincere gratitude to my supervisor, Professor Xiao-Wen Chang, 
for his continuous guidance, support, and encouragement throughout my doctoral studies. 
His mentorship has played a central role in shaping the direction of my research, 
guiding me from graph representation learning toward the less explored area of learning to optimize for integer-valued parameter estimation in linear models. 
Under his supervision, 
I have grown both as a researcher and as a person. 
His patience, constructive feedback, and insightful advice have helped me navigate this transition and pursue new research directions with confidence.

I am also grateful to my collaborators, 
especially Sitao Luan, Ziyang Song, Jiaqi Zhu, and Yingzi Xu.
Sitao Luan has provided valuable support and insights based on his expertise in graph representation learning.
I would also like to thank Ziyang Song for his collaboration on our work on time series transformers,
which, although outside the scope of this thesis,
has been an important and rewarding part of my research experience.
Jiaqi Zhu and Yingzi Xu have worked closely with me and have made indispensable contributions to our papers.

Finally, I would like to thank everyone who has supported me throughout this journey. Their encouragement and belief in me have helped me overcome challenges and continue working toward my life goals.

%% file: content/GenAI_statement.tex
\chapter*{\rm\bfseries Generative AI Statement}
\label{ch:gen_ai_statement}

I hereby confirm that I have consulted the \href{https://www.mcgill.ca/gps/students/policies-and-guidelines/guidelines-using-generative-ai-research}{guidelines on the use of 
generative artificial intelligence} issued by McGill's Graduate and Postdoctoral Studies
(GPS) and have familiarized myself with my responsibilities as a student
regarding the use of generative AI in research.
I have consulted with my supervisor and,
where appropriate, my department to ensure that any use of 
generative AI in this thesis complies with both university
policies and the norms of my discipline.
I attest that all use of generative AI in this thesis
adheres to these guidelines.
Where necessary, I reference the use of generative AI in my thesis.
I understand that any misuse of generative AI may result in 
disciplinary action under the \href{https://www.mcgill.ca/secretariat/files/secretariat/code_of_student_conduct_and_disciplinary_procedures.pdf}{Code of Student Conduct}.

%% file: content/Introduction.tex
\chapter{\rm\bfseries Introduction}
\label{ch:introduction}

Graph data are non-Euclidean and can capture complex interdependencies among  entities~\cite{sperduti1997supervised, micheli2005new, gori2005new, bacciu2020gentle}.
Recent years have witnessed rapid progress in graph representation learning, 
which maps graph data into vector representations for downstream tasks~\cite{scarselli2008graph, kipf2016semi, wang2018graphgan, hamilton2017inductive, velivckovic2017graph, xu2018powerful},
such as node classification~\cite{kipf2016semi, hamilton2017inductive, velivckovic2017graph, luan2019break}, 
graph classification~\cite{xu2018powerful, yang2024incorporating},
and link prediction~\cite{zhou2022link}.
Despite these successes,
widely used graph message passing architectures 
suffer from insensitivity to information from distant and poorly connected nodes~\cite{li2018deeper, zhang2018graph, nt2019revisiting, alon2020bottleneck, topping2021understanding, di2023over, arroyo2026vanishing}.
One part of this thesis aims to address this challenge.

In practice,
graph representation learning is effective in many real-world applications,
including anomaly detection~\cite{gao2023addressing}, 
recommendation~\cite{ying2018graph, jiang2024challenging}, 
transportation~\cite{Li2018},
and molecular modeling~\cite{bongini2021molecular, gaudelet2021utilizing, hua2024mudiff}.
It has also emerged as a useful tool in the field of learning to optimize~\cite{chen2022representing},
where machine learning-based methods are developed to efficiently produce approximate solutions
to optimization problems whose exact solutions are computationally expensive to obtain~\cite{monga2021algorithm, chen2022learning, amos2023tutorial, bengio2021machine, mazyavkina2021reinforcement}.
Graph representation learning shows advantages
because many optimization problems, 
such as the traveling salesman problem,
vehicle routing problem, and mixed-integer linear programming,
induce an underlying graph structure~\cite{khalil2017learning, kool2018attention, joshi2019efficient, drori2020learning}.
This thesis considers optimization problems arising in integer parameter estimation in linear model~\cite{bai2014low, damen2003maximum, yao2002lattice, chang2005mlambda, hassibi2002integer, teunissen1998success, teunissen1993least, chen2012syrafa, wang2010pauc},
that comes from applications 
including multiple-input multiple-output (MIMO) communication systems~\cite{bai2014low, damen2003maximum, yao2002lattice}, 
GPS positioning~\cite{chang2005mlambda, hassibi2002integer, teunissen1998success, teunissen1993least} and utilization control~\cite{chen2012syrafa, wang2010pauc}.
For some types of these optimization problems,
each problem instance can be represented as a densely connected graph,
making it naturally suitable for graph representation learning~\cite{scotti2020graph, kosasih2022graph, kosasih2022graph2, he2023gnn}.
For such problems,
this thesis develops graph representation learning-based approaches that leverage the graph structures to better capture the underlying structural information of each problem instance and addresses some limitations of existing methods.

We use MIMO detection as the main application for studying graph representation learning because it provides a practically important family of hard optimization problems that must be solved repeatedly under strict latency requirements. 
As the problem size increases, 
obtaining the optimal solution becomes computationally prohibitive, 
making efficient suboptimal estimation increasingly important.
Moreover, each problem instance contains structured dependencies among the unknown variables, 
making MIMO detection a natural setting for studying how graph representation learning can improve learning-based optimization.

This thesis also studies integer parameter estimation problems in linear models that do not induce graphs.
For these problems, it develops learning-based methods that closely approximate optimal solutions that are computationally infeasible to obtain exactly.




\section{Organization and Contributions}
\input{content/Chapter1/organization}

\section{Statement Regarding to Authorship and Contributions}
\input{content/Chapter1/contribution_of_authers}

\section{Notation}
\input{content/Chapter1/notation}

%% file: content/Chapter1/organization.tex
\begin{figure}[t]
\centering
\includegraphics[scale=0.75]{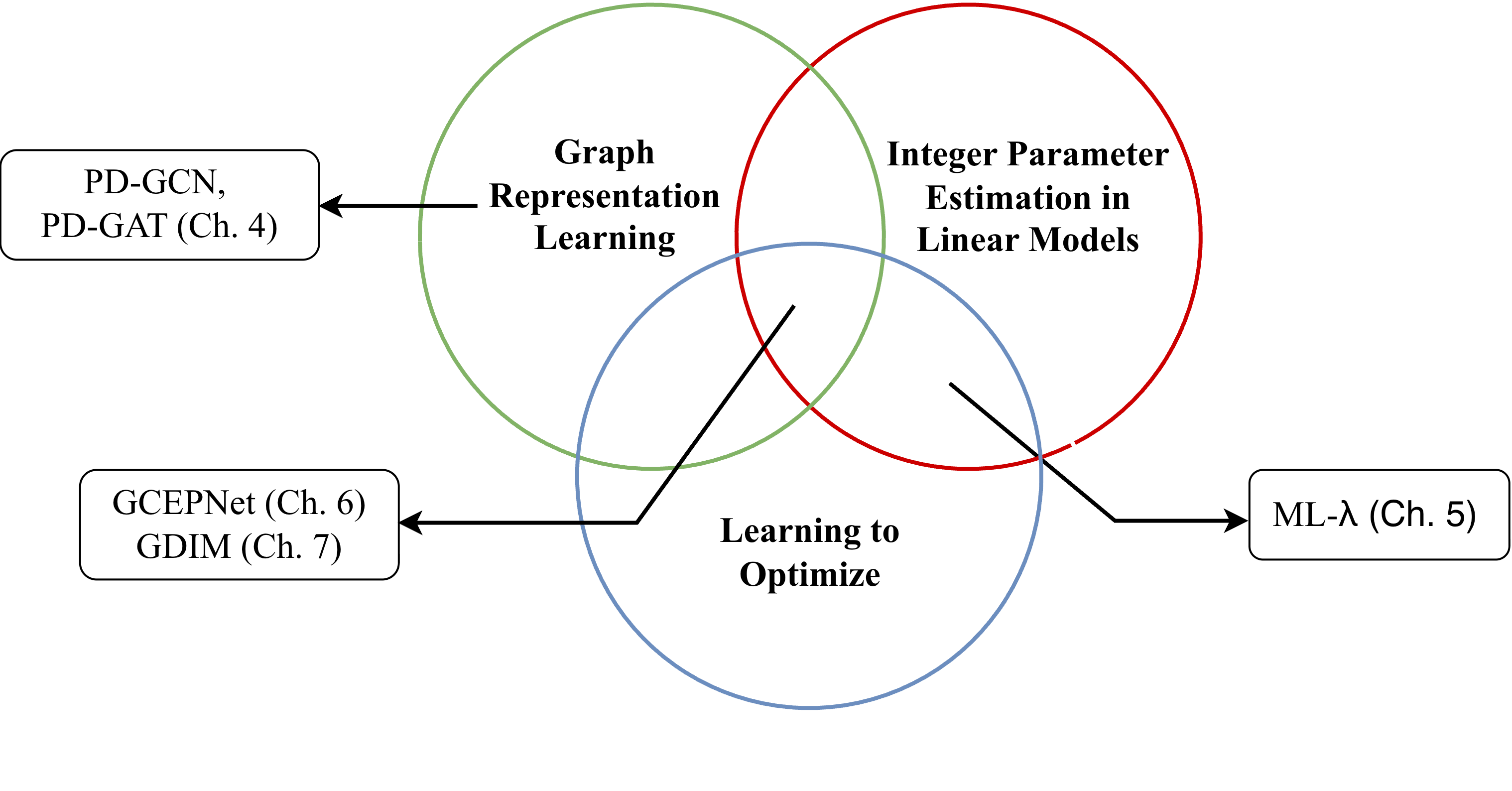}
\caption{Research scope of this thesis and the relationship between the main research areas and proposed methods. PD-GCN and PD-GAT focus on graph representation learning; ML-$\lambda$ combines learning to optimize with integer-valued parameter estimation in linear models; GCEPNet and GDIM lie at the intersection of graph representation learning, learning to optimize, and integer-valued parameter estimation.}
\label{fig:research_scope}
\end{figure}

This thesis aims to address some limitations in graph representation learning and to develop learning-based methods, especially graph representation learning-based methods, for optimization problems arising in integer-valued parameter estimation in linear models.
Fig.\ \ref{fig:research_scope} illustrates the research scope of this thesis and the relationships among its main contributions.
The organization and contributions of this thesis are as follows.

Chapter~\ref{ch:Preliminary} introduces the necessary background, 
including basics of graph, linear models, 
and integer-valued parameter estimation in linear models.
In this thesis, 
we refer to this problem as estimation or detection problem,
and to the corresponding methods as estimators or detectors.

In Chapter~\ref{ch:LiteratureReview},
we review foundational literature related to the topics covered in this thesis,
including graph representation learning for node classification tasks,
learning to optimize,
and commonly used estimators.

Chapter~\ref{ch:Paper02} addresses the limitation of graph message passing mechanism that fails to capture long-range topological information for semi-supervised node classification.
Based on a novel class of parameterized Laplacian matrices that enables adaptive diffusion scopes to accommodate different levels of graph homophily,
we propose Parameterized Diffusion-based Graph Convolutional Networks (PD-GCN) and Parameterized Diffusion-based Graph Attention Networks (PD-GAT)~\cite{lu2024flexible},
and a topology-guided rewiring strategy that facilitates long-range message passing.
The proposed methods outperform SOTA baselines on six out of seven real-world datasets.

The remaining chapters focus on machine learning-based approaches for the estimation problem,
with particular emphasis on methods that employ graph representation learning techniques.

Chapter~\ref{ch:Paper01} extends the well-known Babai estimator,
a commonly used low-complexity suboptimal estimator,
and aims to maximize the success probabilities of the resulting estimators.
This leads to optimization problems that are not graph-structured and are computationally infeasible to solve directly.
Thus graph representation learning is not used in this chapter.
We propose ML-$\lambda$ that learns a structured latent search space in which the success probabilities can be estimated and optimized effectively and efficiently.

Chapter~\ref{ch:Paper03} employs graph representation learning to address computational bottlenecks in the existing SOTA estimator.
We show that the linear model can be represented as a spectral graph convolution on a graph and propose the graph convolution-enhanced expectation propagation network (GCEPNet),
which achieves SOTA performance while enabling significantly faster inference.

Chapter~\ref{ch:Paper04} proposes GDIM,
a graph-based denoising diffusion method,
to address limitations of existing diffusion-based methods for the estimation problem.
The proposed method substantially outperforms existing diffusion-based methods,
especially when the linear model is underdetermined,
while also achieving a better accuracy--runtime trade-off than baseline methods.

Lastly, Chapter~\ref{ch:conclusions} summarizes this thesis and discusses future research directions.

%% file: content/Chapter1/contribution_of_authers.tex
Excluding the Introduction, Preliminary, 
Related Work,
and Conclusion and Future Work chapters,
the remaining chapters of this thesis, 
namely Chapters~4--7, 
are composed of content from previously published work and manuscripts authored and co-authored by Qincheng Lu.
According to the standard publication agreements of these venues,
Qincheng Lu has the right to reproduce these works in this thesis.

\begin{itemize}
\item Chapter~\ref{ch:Paper02} is based on~\cite{lu2024flexible}.
I contributed to the original idea.
I wrote the code, ran the experiments, and wrote most of the published manuscript.
My supervisor, Xiao-Wen Chang, and my collaborator, Jiaqi Zhu, contributed to the original idea and the theoretical foundation of the proposed methods.
Jiaqi Zhu contributed to the development of the directional graph attention network~\cite{Zhu23}, 
an earlier version related to PD-GAT and the rewiring method in Chapter~\ref{ch:Paper02}.
I contributed to improving this earlier method, 
leading to the PD-GAT model and the rewiring method in Chapter~\ref{ch:Paper02}, 
which achieve better performance on broader benchmarks.
My collaborator, Sitao Luan, suggested improvements to the work and assisted with proofreading the paper.
My supervisor, Xiao-Wen Chang, provided supervision and proofreading support.

\item Chapter~\ref{ch:Paper01} is based on~\cite{chang2023success}.
My supervisor, Xiao-Wen Chang, proposed the theoretical foundation of the extended $L_2$-norm regularized Babai estimators and the optimization problem of maximizing its success probabilities, 
which are stated in Section~\ref{sec:ml_lambda_problem}.
I contributed to the original idea, wrote the code, and ran the experiments for the proposed learning-based method to approximate the optimal solution to the optimization problem.
I wrote the sections of the paper corresponding to the learning-based method.
My collaborator, Yingzi Xu contributed to dataset generation, experimental results analysis and proofreading.
My supervisor, Xiao-Wen Chang, provided supervision and proofreading support.

\item Chapter~\ref{ch:Paper03} is based on~\cite{lu2024gcepnet}, 
with some additional results.
I contributed to the original idea, wrote the code, ran the experiments, and wrote the paper.
Sitao Luan contributed to the original idea and assisted with proofreading.
My supervisor, Xiao-Wen Chang, provided supervision and proofreading support.

\item Chapter~\ref{ch:Paper04} is based on the manuscript~\cite{lu2026gd4}, with some additional results.
I contributed to the original idea, wrote the code, ran the experiments, and wrote the paper.
Sitao Luan contributed to the original idea and assisted with proofreading.
My supervisor, Xiao-Wen Chang, provided supervision and proofreading support

\end{itemize}


%% file: content/Chapter1/notation.tex
This section introduces some notation and terms to be used in this thesis.

We use boldface lowercase and uppercase letters to denote column vectors and matrices respectively,
and we use normal lowercase and uppercase letters to denote scalars.
In this thesis, unless specifically stated otherwise, 
we assume that all indices start from 1.

We use $\mathbb{R}$, $\mathbb{C}$ and $\mathbb{Z}$ to denote the set of all real numbers, all complex numbers and all integers respectively.
For a set of scalars $\mathbb{M}$,
we use $\mathbb{M}^n$ to denote the set of all $n$-dimensional column vectors with each entry from $\mathbb{M}$.
And we use $\mathbb{M}^{m \times n}$ to denote the set of all $m$ by $n$ matrices with each entry from $\mathbb{M}$.

For a set $S$, we use $|S|$ to denote the number of elements in $S$.


We use $\mathcal{G} := (\mathcal{V}, \mathcal{E})$ to denote a graph with the vertex set $\mathcal{V}$ and the edge set $\mathcal{E}$.
Unless specially mentioned, all $\mathcal{G}$ are undirected and connected.

For a function $f$ of $\x$, we use $\nabla_{\x} f$ to denote the gradient of function $f$ with respect to $\x$.

We use $i:j$ to denote the sequence of integers $i, i+1, i+2, \ldots, j-1, j$.

For a matrix $\A \in \mathbb{R}^{n \times n}$,
$\A_{i, :}$ denotes the $i^{th}$ row of $\A$.
We use $\a_j$ and $\A_{:, j}$ to denote the $j^{th}$ column of $\A$.
We denote $a_{ij}$ or $\A(i, j)$ or $\A_{ij}$ as the $(i, j)$ element of $\A$.
The transpose of $\A$ is denoted by $\A^\top$.
For a square matrix $\A \in \mathbb{R}^{n \times n}$,
we define its trace as
\begin{equation*}
    \mathrm{trace}(\A) = \sum_{i=1}^n \A_{ii},
\end{equation*}
and use 
\begin{equation*}
\text{diag}(\A) = [a_{11}, \ldots, a_{nn}]^\top \in \mathbb{R}^n \nonumber
\end{equation*}
to denote the column vector containing its diagonal elements.

For a vector $\a \in \mathbb{R}^n$, 
we denote its $i^{th}$ element by either $\a(i)$ or $a_i$.
We use $\mathrm{diag}(\a) \in \mathbb{R}^{n \times n}$ to denote the diagonal matrix whose diagonal entries are the elements of $\a$.

We write $\a \in S^n$ to denote an $n$-dimensional vector whose entries all belong to the set $S$.
Let $s_j$ be the $j^{th}$ smallest element in $S$, $|S| = m$,
we denote by
$\1_{S}(\a) \in \{0,1\}^{n \times m}$ the one-hot encoding matrix of $\a \in S^n$, where
\begin{equation}
\1_{S}(\a)(i, j) =
\begin{cases}
1, & \a(i) = s_j, \\
0, & \text{otherwise}.
\end{cases}
\label{eq:one-hot}
\end{equation}
And for a random vector $\a \in S^n$
with probability matrix
$\P \in \Rbb^{n \times m}$,
where $\P(i, j) = \Pr(a_i = s_j)$,
we write
\begin{equation*}
    \a \sim \mathrm{Cat}(\a; \P).
\end{equation*}
We write $\a \sim \mathrm{Uniform}(S)^n$ if entries of $\a$ are independently and uniformly sampled from $S$.

We use $\A|| \B$ to denote the column concatenation of matrix $\A$ and $\B$,
and $[\A, \B]$ to denote the row concatenation of matrix $\A$ and $\B$.
We use $\odot$ to represent Hadamard (element-wise) matrix product operator.

We also define some special vectors and matrices here. 
We use $\1_n$ to denote the $n$-dimensional vector of ones and $\0_n$ to denote the $n$-dimensional vector of zeros (sometimes the subscript $n$ may be omitted).
We use $\I$ to denote an identity matrix and $\e_k$ to denote the $k^{th}$ column of $\I$.

For a real-valued random vector $\x$,
we use $p(\x)$ to denote its probability density function (PDF).
When $\x$ follows the normal distribution with mean $\bmu$ and covariance matrix $\bSigma$,
we write $\x \sim \calN(\bmu, \bSigma)$ and use $N(\x: \bmu, \bSigma)$ to denote its density function.

The notation $\propto$ stands for equality up to a normalization constant.

The ReLU (rectified linear unit) activation~\cite{fukushima2007visual} is defined as 
\begin{equation}
\mathrm{ReLU}(x) =
\begin{cases}
x, & x > 0, \\
0, & \text{otherwise}.
\end{cases} \label{eq:relu}
\end{equation}
Leaky ReLU allows a small positive gradient when the unit is inactive $(x \leq 0)$~\cite{maas2013rectifier} and is defined as
\begin{equation}
\LeakyReLU(x) =
\begin{cases}
x, & x > 0, \\
0.01x, & \text{otherwise}.
\end{cases} \label{eq:leakyrelu}
\end{equation}
The softmax function for a vector $x$ is defined as
\begin{equation}
\softmax(\x)_i = \frac{\exp(x_i)}{\sum_{k}\exp(x_k)}
\label{eq:softmax}
\end{equation}
The indicator function $\mathbbm{1}\{ B \}$ equals 1 when event $B$ happens and 0 otherwise.

%% file: content/Preliminary.tex
\chapter{\rm\bfseries Preliminary}
\label{ch:Preliminary}

This chapter presents the necessary background for the thesis.
In Section~\ref{sec:pre_grl},
we introduce the basic knowledge of graph.
In Section~\ref{sec:pre_linear_model},
we introduce the linear model and the corresponding optimization problem for integer-valued parameter estimation.

\section{Basics of Graph}
\label{sec:pre_grl}
\input{content/preliminary/spectral_graph_theory}

\section{Linear Model and Optimization for Integer-valued Parameter Estimation}
\label{sec:pre_linear_model}
\input{content/preliminary/linear_model}

%% file: content/preliminary/spectral_graph_theory.tex
For an undirected graph ${\cal G}=({\cal V}, {\cal E})$, 
let $\cal V$ denote the set of $N$ nodes or vertices, indexed by $1,2,\ldots, N$,
and let $\cal E$ denote the set of edges, excluding self-loops.
We use ``node $i$" or $v_i$ to denote the $i^{th}$ node. 
An edge $e_{ij}$ in the edge set $\cal E$ is expressed as a tuple of nodes $(i, j) \in \cal V \times \cal V$.
We associate ${\cal G}$ with a symmetric nonnegative weight matrix $\W=(w_{ij})\in \RNbN$ where
\begin{equation}
\label{eq:weight}
w_{ij} =
\begin{cases}
\mbox{positive weight}, &\mbox{for }\, e_{ij} \in {\cal E}, \\
0, & \mbox{otherwise}.
\end{cases}
\end{equation}
In \eqref{eq:weight}, if $w_{ij}=1$ for $e_{ij} \in {\cal E}$,
then the resulted weight matrix is referred to as 
the adjacent matrix  of  ${\cal G}$ and is often denoted by $\A$.
The degree matrix $\D\in \RNbN$ of ${\cal G}$ is a diagonal matrix, with $d_{ii}=\sum_{j=1}^N w_{ij}$,
and it can be written as $\D= \diag(\W\1)$.  

Let ${\cal N}_i = \{j : e_{ij} \in \cal E \}$ denote the neighborhood set of node $i$,
we have $d_{ii} = |{\cal N}_i|$.

An important matrix in graph theory is the graph Laplacian~\cite{chung1997spectral}, defined as 
\begin{equation*}
\LL = \D - \W.
\end{equation*}
When $\W=\A$, we refer to the graph Laplacian as the combinatorial graph Laplacian.

There are some variants of the graph Laplacian $\LL$~\cite{chung1997spectral}. 
When ${\cal G}$ is connected, i.e., all $d_{ii}\neq 0$,
one of them is the symmetric normalized Laplacian, defined as
\begin{equation*}
\LL_\sym = \D^{-1/2} \LL \D^{-1/2}.
\end{equation*}
Another common variant is the random walk normalized Laplacian, defined as
\begin{equation}
\LL_\rw = \D^{-1} \LL.
\label{eq:rm_laplacian}
\end{equation}

In some graph problems, we have an $N$-dimensional feature/signal vector $\x$
defined on ${\cal V}$.
Its $i^{th}$ element $\x(i)$ is said to be a feature of node $i$.  

The graph Laplacian $\LL$ is symmetric positive semidefinte, because
for any vector $\x$ defined on ${\cal V}$,
\begin{equation*}
\x^\top \LL \x = \frac{1}{2} \sum_{i, j} w_{ij} (x_i - x_j)^2 \geq 0.  
\end{equation*}

Let the eigendecomposition of $\LL$ be
\begin{equation}
\LL = \U \bLambda \U^\top,
\label{eq:eigd}
\end{equation}
where $\bLambda=\diag(\lambda_1,\ldots,\lambda_n)$ 
with the eigenvalues ordered as 
$0=\lambda_1 \leq \lambda_2 \ldots \leq \lambda_N$,
and $\U=[\u_1,\ldots,\u_N]\in \RNbN$ is orthogonal with $\u_i$ being
the eigenvector corresponding to the eigenvalue $\lambda_i$ for $i=1,\ldots,N$. 
It is easy to verify that  $(0, \1)$ is an eigenpair of $\LL$.
We refer to $\u_2$ as the first nontrivial eigenvector, corresponding to the smallest nonzero eigenvalue of $\LL$.

The spectrum of $\LL$ is defined as the set of all the eigenvalues:
\begin{equation*}
\lambda(\LL)=\{\lambda_1,\ldots,\lambda_N\}.
\end{equation*}
Since $\u_i$ is a unit-norm eigenvector, we have
\begin{equation*}
\lambda_i = \u_i^\top \LL \u_i
= \frac{1}{2} \sum_{k,j} w_{kj} (u_{ik} - u_{ij})^2 .
\end{equation*}
A smaller $\lambda_i$ indicates that the graph signal $\u_i$ is smoother~\cite{dakovic2019local}, meaning that any two entries $u_{ik}$ and $u_{ij}$ corresponding to connected nodes $k$ and $j$ tend to have similar values.

Similarly, we define the eigendecomposition of $\LL_\sym$ as 
\begin{equation}
\LL_\sym = \U_\sym \bLambda_\sym \U_\sym^\top,
\label{eq:eigdsym}
\end{equation}
where $\bLambda_\sym=\diag(\lambda_1^{(\sym)},\ldots,\lambda_n^{(\sym)})$ 
with $0=\lambda_1^{(\sym)} \leq \lambda_2^{(\sym)} \leq \ldots \leq \lambda_N^{(\sym)}$,
and $\U_\sym=[\u_1^{(\sym)},\ldots,\u_N^{(\sym)}]\in \RNbN$.
It can be shown that $\lambda_N^{(\sym)} \leq 2$~\cite{chung1997spectral}.

$\LL_\rw$ and $\LL_\sym$ share the same eigenvalues,
and their corresponding eigenvectors satisfy 
$\u_i^{(\rw)} = \D^{-\frac{1}{2}}\u_i^{(\sym)}$~\cite{chung1997spectral},
where $\u_i^{(\rw)}$ denotes the $i^{th}$ eigenvector of $\LL_\rw$ corresponding to the $i^{th}$ smallest eigenvalue. 

The symmetric normalized adjacency matrix corresponding to $\LL_\sym$ is defined as $\A_\sym = \I - \LL_\sym$.
Similarly, we define the random-walk normalized adjacency matrix as $\A_\rw = \I - \LL_\rw$.




%% file: content/preliminary/linear_model.tex
Consider the following linear model which describes the relation between the parameter vector $\x^{*}$ and the observation vector $\y$:
\begin{equation}
\y = \H \x^* + \noisevector, \ \ \noisevector \sim \calN(\0, \sigma_{\noisevector}^2 \I), \label{eq:linear-model}
\end{equation}
where $\H \in \Rbb^{m \times n}$ is a model matrix,
$\x^{*} \in \Zbb^{n}$ is an integer parameter vector and may be subject to some constraint,
and $\noisevector$ is a random noise vector following the Gaussian distribution $\calN(\0, \sigma_{\noisevector}^2 \I)$. 
We say this linear model is overdetermined when $\H$ has full column rank, \ie{} 
$\text{rank}(\H) = n$, otherwise, we call it underdetermined. 
In this thesis, unless specifically mentioned otherwise, we assume that the linear model is overdetermined.

Part of this thesis studies the problem of recovering the unknown true parameter vector $\x^*$,
given the model matrix $\H$ and the observation vector $\y$.
This is an estimation problem.
In communications, $\x^*$ is often subject to some finite constraints and estimating $\x^*$ is often referred to as a detection problem.
We use both terms estimation and detection in this thesis.

The estimation problem comes from various applications,
including multiple-input multiple-output (MIMO) communication systems~\cite{bai2014low, damen2003maximum, yao2002lattice}, 
GPS positioning~\cite{chang2005mlambda, hassibi2002integer, teunissen1998success, teunissen1993least}, utilization control~\cite{chen2012syrafa, wang2010pauc}, etc. 

According to Bayes' rule, the probability density function (PDF) of the posterior of the true parameter vector $\x^*$ conditioned on the observation vector $\y$ satisfies
\begin{equation*}
p(\x^*=\x \,|\, \y) 
= \frac{p(\y \,|\, \x^*=\x) p(\x^*=\x)}{p(\y)} \propto p(\y \,|\, \x^*=\x) p(\x^*=\x), 
\end{equation*}
which is referred to as the posterior distribution in this thesis for simplicity.
Given the model matrix $\H$ and the observation vector $\y$,
the maximum a posteriori (MAP) estimation method solves the optimization problem:
\begin{equation*}
\underset{\x \in \Zbb^n}{\text{max}} \: \text{ln} \left( p(\x^* = \x \,|\, \y) \right). 
\end{equation*}
The optimal solution to this problem is called the MAP estimator to the true parameter $\x^* \in \Zbb^n$ in the linear model~\eqref{eq:linear-model}.

In the absence of a specified prior, or when $\x^*$ is modeled as a deterministic unknown parameter, one can apply the maximum likelihood  estimation method, 
which solves the optimization problem:
\begin{equation}
\underset{\x \in \Zbb^n}{\text{max}} \: \text{ln} \left( p(\y \,|\, \x^* = \x ) \right), \label{eq:linear-model-ml}
\end{equation}
where $p(\y \,|\, \x^* = \x)$ is the PDF of $\y$ given $\x^* = \x$, referred to as the likelihood and is given by:
\begin{equation*}
p(\y \,|\, \x^* = \x) = N(\y: \H \x, \sigma_{\noisevector}^2 \I ) = \frac{1}{\left( \sqrt{2 \pi} \sigma_{\noisevector} \right)^n} e^{-\frac{1}{2 \sigma_{\noisevector}^2}\| \y - \H \x \|_2^2}.
\end{equation*}
The optimal solution to the problem~\eqref{eq:linear-model-ml} is called the maximum likelihood estimator.
It is easy to see that~\eqref{eq:linear-model-ml} is equivalent to the ordinary integer least squares (OILS) problem:
\begin{equation}
\underset{\x \in \Zbb^n}{\text{min}} \: \| \y - \H \x \|^2. \label{eq:linear-model-ils}
\end{equation}

In lattice theory~\cite{lyness1989introduction, proakis2001digital}, 
the model matrix $\H$ is referred to as a lattice generator matrix with its columns as lattice basis vectors, 
and the lattice generated by $\H$ is defined as the set:
\begin{equation*}
\mathcal{L}(\H) := \{ \H \x: \x \in \mathbb{Z}^n \}.
\end{equation*}
Therefore,
the OILS problem~\eqref{eq:linear-model-ils} is equivalent to finding a lattice point $\x \in \mathcal{L}(\H)$ closet to $\y$. 
Thus, the OILS problem is also referred to as the closet vector problem (CVP) in
information theory \cite{agrell2002closest}.
The OILS problem also arises in fields such as lattice cryptograph~\cite{lagarias1985solving},  
number theory~\cite{platonov1993algebraic}, 
and quantization in machine learning~\cite{chen2026the, wang2026ojbkq, birnick2026the},
where no estimation is involved.

A common possible constraint on $\x^*$ in applications is that the entries of $x_i^*$ 
are independently distributed, with $x_i^*$ uniformly distributed over a finite alphabet set 
$\calX^{(i)} \subseteq \mathbb{Z}$ for $i=1:n$. In general, we can assume:
\begin{equation*}
\calX^{(i)} = \{l^{(i)}, l^{(i)}+1, \cdots, u^{(i)} \},
\end{equation*}
where $l^{(i)}$ and $u^{(i)}$ are integers.
Denote
\begin{equation*}
\calX = \calX^{(1)} \times \calX^{(2)} \times
\ldots \calX^{(n)}.   
\end{equation*}
To estimate the true parameter vector $\x^*$ in~\eqref{eq:linear-model},
the maximum likelihood estimation method solves the optimization problem:
\begin{equation*}
\underset{\x \in \calX}{\text{max}} \: \text{ln} \left( p(\y \,|\, \x^* = \x ) \right), 
\nonumber
\end{equation*}
which is equivalent to
\begin{equation}
\underset{\x \in \calX}{\text{min}} \:  \| \y - \H \x \|^2. 
\label{eq:linear-model-bils}
\end{equation}
We refer to~\eqref{eq:linear-model-bils} as a box-constrained integer least squares (BILS) problem.

One important application of the box-constrained linear model based estimation is the multiple-input multiple-output (MIMO) systems in wireless communications. 
We briefly describe the MIMO detection model with quadrature amplitude modulation (QAM) constellation. 

In a MIMO system, 
the transmitter is equipped with $N_T$ antennas and the receiver is equipped with $N_R$ antennas.
The transmitted signal $\x_c^*$ and the received signal $\y_c$ satisfy the following model:
\begin{equation}
\y_c = \H_c \x_c^* + \noisevector_c, \label{eq:mimo-sys-model-complex}
\end{equation}
where $\H_c \in \Cbb^{N_r \times N_t}$ is referred to as the channel matrix of the system model, 
which is assumed to be known to the receiver. 
The noise vector $\noisevector_c \sim \calC \calN (\mathbf{0}, \sigma_{\noisevector_c}^2 \I)$ is a complex Gaussian noise vector.
The transmitted signal $\x_c^*$ is also referred to as the transmitted symbols.
And entries of $\x_c^*$ are assumed to be independently and uniformly distributed over a finite set $\widetilde{\calX}_k$, 
referred to as the constellation set, 
defined by:
\begin{equation}
\widetilde{\calX}_k = \{a + jb : a, b \in \calX_k\},
\qquad
\calX_k = \{\pm 1, \pm 3, \ldots, \pm (2^k-1)\}.
\label{eq:mimo_const}
\end{equation}
Here, $k=1, 2, 3$ correspond to QPSK, \ie{} 4-QAM, 16-QAM, 64-QAM constellations, respectively~\cite{proakis2001digital}.
In this thesis,
$k$ is referred to as the order of constellation.
We denote the constellation size as $ M = |\calX_k| = 2^{k}$.

Given $\y_c$ and $\H_c$, 
the MIMO detection problem is to recover the transmitted vector $\x_c^*$.
The complex model~\eqref{eq:mimo-sys-model-complex} can be easily transformed to the real linear model~\eqref{eq:linear-model}. Define
\begin{align}
&\y = \begin{bmatrix} \Re(\y_c) \\ \Im(\y_c) \end{bmatrix}, \
\H = \begin{bmatrix} \Re(\H_c) & -\Im(\H_c) \\ \Im(\H_c) & \Re(\H_c) \end{bmatrix}, \ \x^* = \begin{bmatrix} \Re(\x_c^*) \\ \Im(\x_c^*) \end{bmatrix}, 
\ 
\noisevector = \begin{bmatrix} \Re(\noisevector_c) \\ \Im(\noisevector_c) \end{bmatrix}, \label{def:ils-real}
\end{align}
with $\Re(\cdot)$ and $\Im(\cdot)$ denoting the real and imaginary parts of a quantity.
Then the complex system model in \eqref{eq:mimo-sys-model-complex} becomes
\begin{equation} 
\y = \H \x^* + \noisevector, \ \ 
\noisevector \sim \calN(\0, \sigma_{\noisevector}^2\I), \label{eq:mimo-sys-model-real}
\end{equation}
where $\x^* \in \calX_k^{n}$, $\sigma_{\noisevector}^2=\sigma_{\noisevector_c}^2/2$, $m := 2N_R$ and $n := 2N_T$.

For MIMO detection, the maximum a posterior estimation finds
\begin{equation*}
\hat{\x}_{\mathrm{MAP}} = \argmax_{\x \in \calX_k^n} p(\x = \x^* \,|\, \y) = \argmax_{\x \in \calX_k^n} p(\y \,|\, \x=\x^*) p(\x = \x^*).  
\end{equation*}
Under the common assumption that  
each element of $\x^*$ is independently and uniformly distributed over $\calX_k$,
the prior probability $p(\x = \x^*)$ is constant for all feasible vectors $\x \in \calX_k^n$,
then the MAP estimation becomes the maximum likelihood estimation,
which solves the integer least squares (ILS) problem
\begin{equation}
\min_{\x \in \calX_k^n} \left\| \y - \H \x \right\|^2_2.
\label{eq:ils-transformed}
\end{equation}

Here we introduce some concepts used in this thesis related to MIMO detection.
The signal-to-noise (SNR) ratio for the system model~\eqref{eq:mimo-sys-model-complex} is defined as:
\begin{equation}
\text{SNR}_{\text{db}} = 10\, \text{log}_{10} \left( \Exp{ \| \H_c \x_c \|_2^2} / \Exp{ \| \noisevector_c \|^2_2} \right).
\label{eq:snr}
\end{equation}

For the constellation set $\calX_k$, 
the mean symbol energy refers to the variance of a uniformly distributed symbol $x \in \calX_k$,
and we also denote it as $\sigma_{\x^*}^2$ in this thesis,
which can be calculated as 
\begin{equation}\label{eq:sigmax}
\sigma_{\x^*}^2 = (M^2 - 1)/3.
\end{equation} 

For the underdetermined MIMO detection, 
we have the number of receive antennas is less than the number of transmit antennas, \ie{} $N_r < N_t$, 
In general, 
an underdetermined ILS problem is much more difficult to solve than an overdetermined problem with the same dimension (\ie{} $2N_t$).

There are different methods to detect $\x^*$.
A review of commonly used detection techniques can be found in~\cite{AlbJS19} and Section~\ref{sec:detection_methods}.
For both the overdetermined ($N_r \geq N_t$) and underdetermined ($N_r < N_t$) problem,
one method is to solve the integer version of
the minimum mean squares error (MMSE) problem
\begin{equation} 
\label{eq:ils-transformed-l2-reg}
    \min_{x \in \calX_k^n} \|\y - \H\x\|_2^2 + \lambda_*^2\norm{\x}_2^2,
\end{equation}
where $\lambda_* = \sigma_{\noisevector} / \sigma_{\x^*}$.
Note that~\eqref{eq:ils-transformed-l2-reg} is an $L_2$-regularized integer least squares problem
and we can rewrite it as 
\begin{equation}
\min_{\x \in \calX_k^n} \left \|\bmx \y \\ \0 \emx - \bmx \H \\ \lambda_* \I_n \emx \x \right\|^2_2,
\label{eq:ils-transformed-l2}
\end{equation}


The ILS problems are NP-hard 
in general~\cite{emde1981another, verdu1989computational, micciancio2002hardness},
meaning that any algorithm guaranteed to find the optimal solution is believed to have exponential worst-case time complexity in the worst cases.
This difficulty becomes especially significant in massive MIMO systems, 
where $N_t$ is large, 
or high-order constellations are used. 
For this reason, 
practical MIMO detection often relies on suboptimal estimators that trade exact optimality for lower computational cost.

We next introduce several metrics used to evaluate detection performance.
Let $\hat{\x}$ denote an estimator of the transmitted vector $\x^*$.
The success probability (SP) measures the probability that the entire transmitted vector is correctly detected and is denoted as 
\begin{equation*}
\mathrm{SP} = \Pr(\hat{\x} = \x^*).
\end{equation*}
In communications literature, 
one often uses the term ``vector error probability",
which is the probability that the estimated vector is not equal to the true transmitted vector, \ie{} 1-SP.

Another commonly used metric is the symbol error rate (SER), 
which measures the probability that a transmitted symbol is incorrectly detected and is defined as
\begin{equation*}
\mathrm{SER} = 1 - \frac{1}{n}\sum_{i=1}^{n} \Pr(\hat{x}_i = x_i^*).
\end{equation*}

Other commonly used metrics include the bit error rate (BER). 
After each symbol is mapped to its corresponding binary representation, 
BER measures the probability that a transmitted bit is incorrectly detected. 

In this thesis, 
SP and SER are used to evaluate the performance of suboptimal estimators in the relevant chapters.

%% file: content/RelatedWork.tex
\chapter{\rm\bfseries Related Work}
\label{ch:LiteratureReview}

This chapter reviews the related literature on several topics covered in this thesis, including graph representation learning for node classification tasks, 
learning-based methods for integer-valued parameter estimation in linear models, 
and the intersection of these two areas.

Section~\ref{sec:literature_grl} reviews graph representation learning for node classification tasks, 
including spectral graph convolution and spatial graph neural networks. 
This section provides background for Chapter~\ref{ch:Paper02}, 
where graph neural networks are studied for node-level tasks, 
as well as for Chapters~\ref{ch:Paper03} and~\ref{ch:Paper04}, 
where graph neural network-based models are used to obtain suboptimal solutions to the estimation problem.

Section~\ref{sec:learning_to_optimize} provides context for learning-based methods used as suboptimal solvers for computationally hard optimization problems in general, 
including settings that are not necessarily graph-based.
This section provides background for the methods proposed for integer-valued parameter estimation in linear models in Chapters~\ref{ch:Paper01}, \ref{ch:Paper03}, and~\ref{ch:Paper04}.

Lastly, Section~\ref{sec:detection_methods} reviews classical and learning-based methods for integer-valued parameter estimation in linear models,
to provide background for Chapters~\ref{ch:Paper01}, \ref{ch:Paper03}, and~\ref{ch:Paper04}.

\section{Graph Representation Learning}
\label{sec:literature_grl}

Graph representation learning aims to learn vector representations of vertices in graph-structured data~\cite{wang2018graphgan}.
These representations are used for downstream tasks including node classification~\cite{kipf2016semi}, 
link prediction~\cite{zhou2022link}, 
and graph classification~\cite{yang2024incorporating},
leading to successful applications in domains such as anomaly detection~\cite{gao2023addressing}, 
recommender systems~\cite{ying2018graph, jiang2024challenging},
transportation~\cite{Li2018}, and biomedical~\cite{Fout2017}.
In this thesis, we focus on graph representation learning for node classification tasks.

Graph Neural Networks (GNNs) were initially introduced in~\cite{gori2005new, scarselli2008graph}
as an extension of recursive neural networks for graph-structured data.
Since then, GNNs have become a general framework for graph representation learning~\cite{scarselli2008graph,kipf2016semi,
hamilton2017inductive,velivckovic2017graph,xu2018powerful,luan2019break,beaini2021directional,zhao2021consciousness, hua2024mudiff}.
The central idea of GNNs is to update node representations by propagating information along the edges of a graph.
At each propagation step, 
the representation of a node is updated using its previous representation together with information from its neighboring nodes.
By repeating this process, 
GNNs allow each node to aggregate information from increasingly larger neighborhoods, 
producing node representations that encode higher-order neighborhood information.

Existing GNN architectures can be divided into two categories, spectral methods and spatial methods.
Spectral methods define graph convolution from the perspective of graph signal processing, 
typically using the graph Laplacian and its spectral decomposition.
Spatial methods, in contrast, define graph convolution directly through local neighborhood aggregation in the vertex domain.
In the following sections, 
we review these two lines of work.

\subsection{Spectral Graph Convolution}
\label{sec:spectral_gnn}
\input{content/relatedworks/graph_convolution}

\subsection{Spatial Graph Neural Network}
\input{content/relatedworks/spatial_gnn}

\subsection{Transductive and Inductive Learning}
\label{sec:grl_setting}
For node classification, 
the loss function is optimized using the labeled training nodes.
The setting can be categorized into transductive and inductive depending on whether the unlabeled nodes to be predicted are observed during training~\cite{hamilton2020graph}.

\paragraph{Transductive Learning}
In the transductive setting, 
the labeled training nodes and the unlabeled validation and testing nodes belong to the same graph.
A single GNN model is trained on this graph using the graph structure and node features of all nodes, 
while labels are available only for the training nodes during optimization.
The node representations are produced by graph message passing over all nodes in the graph.
Therefore, although the labels of the validation and testing nodes are not used during training, 
their features and edges are available and can participate in message passing.
The validation set is used for model selection and hyperparameter tuning.
This setting is often used in semi-supervised learning for node classification~\cite{zhou2003learning} and is studied in Chapter~\ref{ch:Paper02}.


\paragraph{Inductive Learning}
In the inductive setting, 
the features and topology involving unlabeled nodes are completely unseen during training.
That is, the model is trained using only the training nodes and must generalize to new nodes or new graphs at test time~\cite{hamilton2017inductive}.
The node classification tasks studied in Chapter~\ref{ch:Paper03}, and~\ref{ch:Paper04},
where graph representation learning-based methods are used for solving optimization problems, 
belong to this inductive setting.
Specifically, 
networks are trained on graphs constructed from training problem instances and tested on new graphs constructed from unseen testing instances.



\section{Learning to Optimize}
\label{sec:learning_to_optimize}
\input{content/relatedworks/generative_model_for_optimization}

\section{Estimation Methods}
\label{sec:detection_methods}
As stated in Section~\ref{sec:pre_linear_model},
both the OILS problem~\eqref{eq:linear-model-ils} and the BILS problem~\eqref{eq:linear-model-bils} are NP-hard~\cite{emde1981another, verdu1989computational, micciancio2002hardness},
meaning that any algorithm that finds the optimal solution is believed to require exponential time complexity in the worst case.
This thesis focuses on suboptimal approaches to the BILS problem~\eqref{eq:linear-model-bils},
where ``suboptimal" means that the methods do not necessarily attain the optimal solution, and in general do not guarantee optimality.
The thesis mainly considers the BILS problem arising from MIMO detection in communications~\eqref{eq:ils-transformed}.
The methods developed for MIMO detection can be easily extended beyond the MIMO setting and are therefore applicable to a broader class of BILS problems~\cite{goldberger2009gaussian}.

Massive MIMO has become a foundational technology in modern wireless communication,
playing a crucial role in the success of 5G systems by substantially improving system throughput and spectral efficiency~\cite{marzetta2016fundamentals, zhang2020prospective}.
Its importance is expected to grow further in future generations of wireless networks,
where large number of user and base station antennas are expected to be deployed~\cite{letaief2021edge}. 
These gains come with a larger problem size and higher order of constellation for estimation and cause computational challenge.
Since finding optimal estimation in massive MIMO systems becomes prohibitively expensive,
a key research problem is to design efficient suboptimal methods that achieve strong performance while maintaining practical computational complexity for real-world deployment~\cite{zhang2020artificial}.

The key challenge in developing sub-optimal estimation methods lies in the trade-off between computational efficiency and performance.
Achieving near-optimal inference at low computational cost is difficult, 
and methods with lower complexity often suffer from degraded performance. 
More recently,
learning-based methods have emerged as a particularly promising direction for massive MIMO detection,
as they offer the potential to achieve near-optimal inference at low latency in application stage.
These methods leverage training data to learn efficient approximations of the optimal estimator,
shifting much of the computational burden to an offline training stage while enabling low-latency estimation during deployment.
For a high level review of learning-based approaches for solving hard optimization problems, 
refer to Section~\ref{sec:learning_to_optimize}.
Approaches for obtaining the optimal solution to the OILS problem~\eqref{eq:linear-model-ils} and the BILS problem~\eqref{eq:linear-model-bils} are reviewed in Section~\ref{sec:optimal}.
For MIMO detection,
Section~\ref{sec:classical} reviews classical, 
non-learning-based methods and
Section~\ref{sec:learning_based_detection} reviews the landscape of learning-based methods.

\input{content/relatedworks/optimal_ils}
\input{content/relatedworks/classical_suboptimal}
\input{content/relatedworks/learning_based_suboptimal}

%% file: content/relatedworks/graph_convolution.tex
To learn node embeddings that depend on both the graph structure and node features or signals, we have seen substantial research on extending deep learning methods to graph-structured data.
In deep learning, 
Convolutional Neural Networks (CNNs) \cite{LeCun1998} have been successful for data with regular grid structures, such as images. 
However, 
the effectiveness of CNNs relies on data properties such as translation invariant coordinates and local statistical dependencies~\cite{Bruna2014}, 
which do not naturally hold for non-Euclidean graph-structured data.

To apply convolutional ideas to graphs, 
one major research direction defines graph convolution from a spectral perspective through an extension of the Fourier transform to graphs.
Early work on spectral graph convolution can be traced to spectral CNNs~\cite{Bruna2014}, 
which define graph Fourier convolution using the eigenbasis of the graph Laplacian.
Specifically, 
a graph signal is first transformed into the graph Fourier domain, 
and convolution is then performed in this spectral domain.
However, 
this approach is computationally expensive since it requires eigendecomposition of the graph Laplacian to calculate both the graph Fourier transform and the inverse graph Fourier transform~\cite{bronstein2017geometric, monti2017geometric}. 


ChebNet~\cite{defferrard2016convolutional} addresses this limitation by parameterizing graph convolution as a polynomial of the graph Laplacian, computed using Chebyshev polynomials.
This formulation avoids explicit eigendecomposition and substantially reduces computational complexity.
Graph Convolutional Networks (GCNs)~\cite{Kipf2017}, another milestone, further simplify ChebNet by using a first-order approximation of the spectral convolution operator.
This simplification further reduces computational complexity by using simple filters that operate on 1-hop neighborhoods.

\subsubsection{Graph Fourier transform} 
Given a graph signal $\x \in \Rbb^N$, 
whose $i^{th}$ entry $x_i$ is defined on node $i$ of ${\cal G}$, 
the graph Fourier transform of $\x$ is defined as 
\begin{equation*}
\hat{\x} = {\cal F}(\x) := \U^\top \x,
\end{equation*}
where $\U$ is the orthogonal matrix given in the eigendecomposition of the graph Laplacian $\LL$ in \eqref{eq:eigd}.
The inverse graph Fourier transform of $\hbx$ is defined as
\begin{equation*}
\x={\cal F}^{-1}(\hat{\x}) := \U \hat{\x}.
\end{equation*}
The graph convolution between two graph signals $\x, \y \in \Rbb^N$, 
denoted by $*_{{\cal G}}: \mathbb{R}^N \rightarrow \RN$, is defined as
\begin{equation*}
\y *_{{\cal G}} \x := {\cal F}^{-1} \left( {\cal F}(\y) \odot {\cal F}(\x)\right) = \U \left( \U^\top \y \odot \U^\top \x \right). 
\end{equation*}

\subsubsection{Spectral filtering of a graph signal} 
Suppose $g_{\btheta}: \lambda(\LL) \rightarrow \mathbb{R}$ is a function that maps the eigenvalues of $\LL$
to real-valued filter coefficients, 
with learnable parameter vector $\btheta$. 
Define 
\begin{equation*}
\g_{\btheta}(\lambda(\LL)) := [g_{\btheta}(\lambda_1), \cdots, g_{\btheta}(\lambda_N)]^\top.
\end{equation*}
The graph spectral filter associated with $g_{\btheta}(\lambda(\LL))$ is a linear operator $\T_{g_{\btheta}}: \RN \rightarrow \RN$ defined by
\begin{equation}
\T_{g_{\btheta}}  \x   = {\cal F}^{-1} \left( \g_{\btheta}(\lambda(\LL)) \odot {\cal F}(\x)\right) = \U \left( \g_{\btheta}(\lambda(\LL)) \odot \U^\top \x \right) = \U \mbox{diag} \left( \g_{\btheta}(\lambda(\LL)) \right) \U^\top \x.
\label{eq:gcf1}
\end{equation}
Denote $\g_{\btheta}(\bLambda) = \mbox{diag} \left( \g_{\btheta}(\lambda(\LL)) \right)$ and $\g_{\btheta}(\LL) = \U  \g_{\btheta}(\bLambda) \U^\top$.
Then we have
\begin{equation} 
\label{eq:Tg}
\T_{g_{\btheta}}  = \U \g_{\btheta}(\bLambda) \U^\top = \g_{\btheta}(\LL).
\end{equation}
Note that $\g_{\btheta}(\LL)$ is the matrix form of the graph spectral filer. 
Here $\g_{\btheta}(\bLambda)$ is a learnable diagonal matrix that controls the response of the filter at different graph frequencies. 
Different parameterizations of $\g_{\btheta}(\bLambda)$ lead to different spectral graph convolution models. 
Note that $\g_{\btheta}$ can be a non-parametric function, defined as
\begin{equation*}
\g_{\btheta}(\bLambda) = \text{diag} (\btheta), \quad \btheta \in \mathbb{R}^N.
\end{equation*}
In this case, 
the filter has one free parameter for each eigenvalue of the graph Laplacian, 
resulting in $N$ learnable parameters.
It is proved that any graph convolution with a well-defined analytic spectral filter can be written in a truncated block Krylov form~\cite{luan2019break} and polynomial filter is one of the most commonly used instances.
Specifically, consider a polynomial spectral filter of order $K$
\begin{equation}
\g_{\btheta}(\bLambda) = \sum_{k=0}^{K} \theta_{k+1} \bLambda^{k}, \quad \btheta = [\theta_1, \ldots, \theta_{K+1}]^\top \in \Rbb^{K+1}.
\label{eq:chebnet_filter}
\end{equation}
According to~\eqref{eq:gcf1} and~\eqref{eq:Tg}, 
applying this filter to a graph signal $\x$ gives
\begin{equation}
\g_{\btheta}(\LL) \x = \sum_{k=0}^{K} \theta_{k+1} \LL^{k} \x,
\label{eq:ChebNetf}
\end{equation}
which is $K$-localized, 
since it only depends on signal on nodes that are at most $K$ hops away.

The above definitions of the graph Fourier transform and spectral filtering are written in terms of $\LL$, 
but they can also be formulated using other Laplacian variants, such as $\LL_{\sym}$, 
by using the corresponding eigendecomposition.


\subsubsection{ChebNet} 
ChebNet proposes an efficient localized graph convolution that avoids explicitly computing the eigendecomposition of the graph Laplacian~\cite{defferrard2016convolutional}.
Specifically, it considers the polynomial spectral filter in~\eqref{eq:chebnet_filter}.
However, when the polynomial order $K$ is large, directly computing the filtered signal $\g_{\btheta}(\LL)\x$~\eqref{eq:ChebNetf} can be inefficient and may suffer from numerical instability.
To address these issues, 
ChebNet uses Chebyshev polynomials to fit $\g_{\btheta}(\LL) \x$. 

Recall that the Chebyshev polynomial $T_k(x)$ of degree $k$ can be computed recursively by 
\begin{equation*}
T_k(x) = 2 x T_{k-1}(x) - T_{k-2}(x). 
\end{equation*}
with $T_0(x)=1$, $T_1(x) = x$, $x \in [-1, \: 1]$.
We denote $T_k(\bLambda) = \mbox{diag}(T_k(\lambda_1 ), \cdots, T_k(\lambda_N ))$.
It is suggested in \cite{Shuman2011} that the polynomial filter $\g_{\btheta}(\bLambda)$ can be well approximated as
\begin{equation}
\g_{\btheta}(\bLambda) = \sum_{k=0}^{K} \theta_{k+1} \bLambda^{k} \approx \sum_{k=0}^{K} \theta_{k+1} T_{k}(\widetilde{\bLambda}),\ \ 
\widetilde{\bLambda} := \frac{2}{\lambda_{N}}\bLambda - \I_N.
\label{eq:ChebNet1}
\end{equation}
where $T_k(\widetilde{\bLambda}) \in \mathbb{R}^{N \times N}$, $T_0(\widetilde{\bLambda}) = \I$ and $T_1(\widetilde{\bLambda}) = \widetilde{\bLambda}$, 
and the scaling and shifting involved in defining $\widetilde{\bLambda}$
is to ensure that its diagonal elements lie in $[-1, \: 1]$. 

Define $T_k(\widetilde{\LL}) = \U T_k(\widetilde{\bLambda}) \U^\top$, and $\widetilde{\LL} = \U \widetilde{\bLambda} \U^\top$, 
we have 
\begin{equation*}
T_k(\widetilde{\LL}) = \U \left( 2 \widetilde{\bLambda} T_{k-1}(\widetilde{\bLambda}) - T_{k-2}(\widetilde{\bLambda}) \right) \U^\top = 2 \widetilde{\LL} T_{k-1}(\widetilde{\LL}) - T_{k-2}(\widetilde{\LL}).
\end{equation*}
Then we can approximate $\g_{\btheta}(\LL)\x$ by  $\sum_{k=0}^{K} \theta_{k+1} T_{k}(\widetilde{\LL}) \x$.
Denote $\bar{\x}_k := T_k(\widetilde{\LL}) \x$.
Then we have
\begin{equation}
\g_{\btheta}(\LL)\x \approx \sum_{k=0}^{K} \theta_{k+1} \bar{\x}_{k},
\label{eq:gtx}
\end{equation}
and
\begin{equation}
\bar{\x}_k = \left[ 2 \widetilde{\LL} T_{k-1}(\widetilde{\LL}) - T_{k-2}(\widetilde{\LL}) \right] \x = 2 \widetilde{\LL} \bar{\x}_{k-1} - \bar{\x}_{k-2}, \: \: \: \bar{\x}_0 = \x, \: \: \: \bar{\x}_1 = \widetilde{\LL} \x. 
\label{eq:ChebNet3}
\end{equation}
Without computing the eigendecomposition of ${\LL}$, we can find the approximation of 
of $\g_{\btheta}(\LL) \x$ on the right hand side of \eqref{eq:gtx} 
by computing a sequence of $\bar{\x}_k$. 
The main cost of computing $\bar{\x}_k$ in \eqref{eq:ChebNet3}
is the cost of computing $\widetilde{\LL} \bar{\x}_{k-1}$.
Note that 
\begin{equation}
\widetilde{\LL}  
= \U \widetilde{\bLambda} \U^\top 
= \U \left( \frac{2}{\lambda_{N}}\bLambda - \I_N \right) \U^\top
= \frac{2}{\lambda_{N}}\LL - \I_N
\label{eq:GCN1}
\end{equation}
Then, for any vector $\y\in \RN$,
\begin{equation*}
(\widetilde{\LL} \y)_i = \frac{2}{\lambda_{N}}
\sum_{e_{ij} \in \cal E} w_{i, j}(y_i -y_j) - y_i.
\end{equation*}
Suppose $\lambda_{N}$ has been computed, say by the Lanczos method.
Computing $(\widetilde{\LL}\y)_i$ only requires accessing the neighbors of node $i$ and costs $O(d_i)$.
Thus computing $\bar{\x}_k$ in \eqref{eq:ChebNet3} costs $O(|{\cal E}|)$. 
As a result, evaluating the right hand side of \eqref{eq:gtx} costs $O(K|{\cal E}|)$,
which is linear in the number of edges for fixed $K$.
Most applications have large and sparse graphs, i.e., $N^2 \gg  |\cal E|$. 
This makes ChebNet substantially more efficient than spectral CNNs that require explicit eigendecomposition of the graph Laplacian.

\subsubsection{Graph convolutional network (GCN)}

The graph convolutional network (GCN) is a simplified variant of ChebNet~\cite{Kipf2017}. 
Each convolutional layer of GCN uses a first-order polynomial filter, corresponding to $K=1$, 
defined using the symmetric normalized graph Laplacian $\LL_{\sym}$ in~\eqref{eq:ChebNet1}.
GCN further assumes $\lambda_N^{(\sym)} \approx 2$~\cite{Kipf2017}.
Then, according to~\eqref{eq:GCN1}, 
we have
\begin{align*}
\g_{\btheta}(\LL_\sym) \x 
& := \theta_1 T_0 (\widetilde{\LL}_\sym) \x + \theta_2 T_1 (\widetilde{\LL}_\sym) \x = \theta_1 \x + \theta_2 (\LL_\sym - \I_N) \x \\
& = \theta_1 \x - \theta_2 \D^{-\frac{1}{2}} \W \D^{-\frac{1}{2}} \x 
= \theta \left( \I_N + \D^{-\frac{1}{2}} \W \D^{-\frac{1}{2}} \right) \x,
\end{align*}
where the constraint $\theta=\theta_2=-\theta_1$ is imposed for simplification.

GCN stacks multiple such convolutional layers.
After $r$ layers, the representation of a node depends on nodes within its $r$-hop neighborhood.
However, repeated multiplication by 
$\I+\D^{-\frac{1}{2}}\W\D^{-\frac{1}{2}}$ can lead to numerical instability, since its eigenvalues lie in $[0,2]$.
GCN addresses this issue using the renormalization trick:
\begin{equation*}
\I+\D^{-\frac{1}{2}}\W\D^{-\frac{1}{2}}
\quad \longrightarrow \quad
\widetilde{\D}^{-\frac{1}{2}}\widetilde{\W}\widetilde{\D}^{-\frac{1}{2}},
\end{equation*}
where
\begin{equation*}
\widetilde{\W}=\W+\I_N,
\qquad
\widetilde{\D}_{ii}=\sum_j \tilde{w}_{ij}.
\end{equation*}

Let $\X^{(\ell)}\in\mathbb{R}^{N\times d_\ell}$ denote the input feature matrix at layer $\ell$, 
where $N$ is the number of nodes and $d_\ell$ is the number of input channels.
The $i^{th}$ row of $\X^{(\ell)}$ corresponds to the representation of node $i$.
The output feature matrix is $\X^{(\ell+1)}\in\mathbb{R}^{N\times d_{\ell+1}}$.
With a learnable weight matrix $\bTheta^{(\ell)}\in\mathbb{R}^{d_\ell\times d_{\ell+1}}$, 
the GCN layer is defined as
\begin{equation*}
\X^{(\ell+1)}
=
\ReLU \left(
\widetilde{\D}^{-\frac{1}{2}}
\widetilde{\W}
\widetilde{\D}^{-\frac{1}{2}}
\X^{(\ell)}
\bTheta^{(\ell)}
\right), 
\end{equation*}
where $\ReLU$ is the element-wise activation function defined in~\eqref{eq:relu}.
Although GCN is derived from spectral graph convolution, 
its layer-wise update can also be interpreted spatially as aggregating and transforming node features from 1-hop neighbors.

%% file: content/relatedworks/spatial_gnn.tex
While spectral GNNs define graph convolution through the spectrum of the graph Laplacian, 
spatial GNNs define convolution directly in the vertex domain through local neighborhood aggregation.
This perspective avoids explicitly relying on the Laplacian eigendecomposition and naturally connects graph convolution with message passing.
Most modern GNNs use the message-passing framework~\cite{hamilton2020graph}, 
where the representation of each node is updated by aggregating information from its neighbors and combining it with its own representation from the previous layer.

A general spatial aggregation layer can be written as
\begin{equation}
\mathbf{X}^{(\ell)} = \sigma ( \mathbf{S} \mathbf{X}^{(\ell-1)} \mathbf{W}^{(\ell)}),
\label{eq:spatial_gnn}
\end{equation}
where $\mathbf{W}^{(\ell)} \in \Rbb^{d_{\ell} \times d_{\ell-1}}$ is a learnable weight matrix, 
$\sigma (\cdot)$ is a non-linear activation function,
and $\mathbf{S} \in \Rbb^{N \times N}$ is an aggregation matrix that specifies how messages from neighboring nodes are weighted.
The entry $\mathbf{S}_{ij}$ determines the contribution of node $v_j$ to the updated representation of node $v_i$ at each layer.

The term diffusion (or random walk) on a graph refers to the process by which quantities, 
such as real-valued signals, propagate from node to node along the edges of a graph~\cite{gasteiger2019diffusion}.
In~\eqref{eq:spatial_gnn},
$\mathbf{S}$ induces the diffusion process on the graph and is referred to as the diffusion operator.
By stacking multiple layers,
information from high-order neighbors are recursively incorporated.

In many GNNs, $\mathbf{S}$ is fixed and constructed from the graph structure, 
such as adjacency matrix variants $\A$, $\A_\rw$ and $\A_\sym$ corresponding to $\mathbf{L}$, $\mathbf{L}_{\mathrm{rw}}$, 
and $\mathbf{L}_{\mathrm{sym}}$, introduced in Section~\ref{sec:pre_grl}.
A limitation of using a fixed aggregation matrix is that all neighbor weights are determined only by the graph structure and do not depend on node features.
Several GNN architectures have been proposed to address these limitations.

In stead of performing message-passing on the entire graph, GraphSAGE~\cite{hamilton2017inductive} introduces a general inductive framework based on neighborhood sampling.
For each node, GraphSAGE samples a fixed-size set of neighbors and aggregates information only from this sampled neighborhood during message passing.
let $\x_i^{(\ell-1)} \in \Rbb^{d_{\ell - 1}}$ denote the representation for node $v_i$ at the $(\ell-1)^{th}$ layer.
The GraphSAGE update can be written as
\begin{equation*}
\x_i^{(\ell)} = \sigma \left ( (\mathbf{W}^{(\ell)})^\top \text{AGGREGATE}_{\ell} (\x_i^{(\ell-1)}, \{ \x_j^{(\ell-1)}, \forall j \in S_{\calN(i)} \})\right), 
\end{equation*}
where $S_{\calN(i)}$ is a random sample of the neighbors of node $v_i$, and $\text{AGGREGATE}_{\ell}$ is the aggregation function at $\ell^{th}$ layer.
Choices of the aggregation function include the mean aggregator, 
which assigns equal weights to sampled neighbors; 
the LSTM aggregator, which applies an LSTM~\cite{hochreiter1997long} to a random permutation of sampled neighbors; 
and the max-pooling aggregator, which applies a nonlinear transformation followed by element-wise max pooling.

Graph attention networks (GATs)~\cite{velivckovic2017graph} introduce an attention mechanism for aggregation, 
which assigns learnable, feature-dependent weights to neighboring nodes.
Let $i$ and $j$ be two connected nodes.
At $\ell^{th}$ layer, the unnormalized attention score from node $v_i$ to node $v_j$ is defined as
\begin{equation*}
r_{ij}^{(\ell)} = \LeakyReLU\big((\mathbf{a}^{(\ell)})^{\top} [\mathbf{W}^{(\ell)} \mathbf{x}_i^{(\ell-1)} || \mathbf{W}^{(\ell)} \mathbf{x}_j^{(\ell-1)}]\big),
\end{equation*}
where $\mathbf{a}^{(\ell)} \in \Rbb^{2 d_{\ell}}$ is a trainable parameter shared over all nodes.
The score for disconnect nodes is set to 0.
The aggregation at the $\ell^{th}$ layer is weighted by the normalized attention score, where the $(i,j)$ entry of the aggregation matrix is given by
\begin{equation*}
\mathbf{S}_{ij}^{(\ell)} = \softmax([r_{i1}^{(\ell)}, \ldots, r_{iN}^{(\ell)}]^\top)_j   
\end{equation*}
using the softmax normalization~\eqref{eq:softmax}.

%% file: content/relatedworks/generative_model_for_optimization.tex
This section reviews recent work on machine learning-based approaches for solving 
some mathematical optimization problems.

The exact solution of many optimization problems is computationally expensive. 
In fact, many important classes of such problems are NP-hard in the worst case~\cite{bengio2021machine}.
This challenge becomes even more pronounced in some practical settings, 
where one often needs to solve a large number of instances of a specific type of problems.
Classical optimization methods typically solve each instance independently,
even though these instances may differ in data while sharing a common underlying structure.
This observation has motivated growing interest in machine learning-based methods for optimization, known as \textit{learning to optimize (L2O)}~\cite{chen2022learning, song2024towards}.
The central idea is to learn reusable patterns from previously solved instances and use this knowledge to solve new ones more efficiently.
After training, 
machine learning-based methods can quickly generate solutions for new problem instances.
In this sense, 
they replace hand-crafted, 
instance-agnostic heuristics with 
data-driven procedures learned from examples.
However, machine learning-based methods trade optimality for computational efficiency and do not provide the same performance guarantees as exact solvers~\cite{bengio2021machine,amos2023tutorial}.
These methods have led to several lines of research, 
including methods that directly construct solutions, 
methods that improve or guide classical optimization algorithms, 
and methods that learn latent spaces for  search~\cite{amos2023tutorial}.


A first line of work learns to construct solutions directly~\cite{bengio2021machine}.
Early neural network approaches,
such as Pointer Networks,
showed that sequence models can represent combinatorial output spaces~\cite{vinyals2015pointer}.
Subsequent work developed reinforcement learning and graph-based models to generate solutions for routing problems, traveling salesman problems and others NP-hard combinatorial optimization problems~\cite{bello2016neural, khalil2017learning, kool2018attention}.
Representative examples include neural combinatorial optimization based on policy gradients~\cite{bello2016neural},
graph-embedding methods that learn greedy construction heuristics~\cite{khalil2017learning},
and attention-based architectures for routing problems~\cite{kool2018attention}.

A second line of work uses machine learning not to replace optimization entirely, 
but to improve or guide the optimization process itself.
One representative approach is algorithm unrolling, 
also known as deep unfolding or deep unrolling~\cite{gregor2010learning, hershey2014deep}.
The basic idea is to start from a classical iterative algorithm, 
where a solution is progressively refined through a sequence of hand-designed, theory-driven update rules, 
and convert a fixed number of its iterations into a trainable neural architecture~\cite{monga2021algorithm}.
In this architecture, 
each iteration is interpreted as one layer of a neural network, 
and selected components of the original update rule, 
such as step sizes, thresholds, regularization parameters, or correction terms, are made learnable parameters~\cite{monga2021algorithm}.
This design preserves much of the structure and interpretability of the original iterative solver, 
while using training data to improve empirical performance.


Another approach in this category uses machine learning to guide key components of optimization solvers.
For example, 
learning-based methods have been used to improve branching decisions in branch-and-bound algorithms~\cite{tang2020reinforcement, liu2022learning}, 
where graph neural networks can imitate strong expert-designed rules and accelerate the search process~\cite{gasse2019exact}.
This line of work shows that learning-based optimization is not limited to end-to-end solution prediction.
It also includes hybrid methods that enhance classical solvers while retaining their underlying algorithmic frameworks~\cite{d2020learning, cappart2023combinatorial}.

A third line of work uses
deep generative models.
Rather than searching in the original solution space, 
these methods learn the distribution of high quality solutions from previously solved problem instances.
One representative approach uses variational autoencoders (VAEs) to map solutions into a continuous latent space, 
where optimization can be carried out more efficiently~\cite{kingma2013auto}.
VAEs-based methods have been used to learn latent search spaces for problems such as routing~\cite{hottung2021learning},
molecular design~\cite{gomez2018automatic},
neural architecture optimization~\cite{luo2018neural}.
The main idea is that high-quality solutions may lie on a structured low-dimensional manifold, 
making search in the learned latent space easier than search in the original domain.

A related approach uses diffusion models to generate solutions through an iterative denoising process, 
rather than predicting a complete solution in a single step~\cite{giannone2023aligning}.
Diffusion model-based methods are appealing for hard optimization problems because directly generating a high quality solution can be difficult.
By gradually refining noisy or partially corrupted candidates, 
diffusion models provide a structured generative process that can better explore complex solution spaces.
For example, Graikos et al.~\cite{graikos2022diffusion} project Euclidean traveling salesman problem instances into an image space and use a diffusion model to generate solution images.
DIFUSCO~\cite{sun2023difusco} formulates combinatorial optimization problems on graphs and uses diffusion models to generate solution graphs.
These solution images or graphs are subsequently decoded into solutions to the original problems.


In summary, machine learning methods do not remove the intrinsic hardness of optimization problems.
Instead, they exploit shared structure across problem instances to improve efficiency and practical solution quality.
This perspective explains why learning-based methods are especially useful when many similar decisions must be made under limited computational budgets, 
and when modest suboptimality is acceptable in exchange for gains in speed or scalability.

This thesis develops machine learning methods, 
particularly graph representation learning-based approaches, 
for integer-valued parameter estimation problems in linear models, which are NP-hard in general.
The proposed methods correspond to the three lines of work reviewed above.
Chapter~\ref{ch:Paper01} follows the first paradigm, 
in which a structured continuous representation of candidate solutions is learned and optimization is performed in the resulting latent space.
Chapter~\ref{ch:Paper03} follows the second paradigm, 
in which a learned graph convolution module is incorporated into an iterative classical expectation propagation estimator.
Chapter~\ref{ch:Paper04} follows the third paradigm, 
in which a learned denoising process progressively refines candidate solutions.


%% file: content/relatedworks/optimal_ils.tex
\subsection{Optimal Estimation}
\label{sec:optimal}
Several approaches have been developed to find the optimal solution to the OILS problem~\eqref{eq:linear-model-ils}
Among them,
enumeration-based algorithms are the most commonly used. 
These methods differ mainly in how the search region is constructed.
There are two ways of defining a search region for the enumeration. 
One approach, based on the Kannan strategy~\cite{kannan1983improved, kannan1987minkowski}, 
uses a rectangular parallelotope as the search region. 
The other, 
which is more common in practical applications, 
restricts the search to integer vectors lying inside a hyper-ellipsoid region. 
This approach is originally proposed by Fincke and Phost~\cite{fincke1985improved} and is later improved by Schnorr and Euchner~\cite{schnorr1994lattice}.
The Schnorr-Euchner strategy enumerates potential candidates within the search region, while shrinking the search region dynamically, until the optimal solution is found. 
In the communications literature, 
this method is commonly known as sphere decoding. 
Although it was originally developed for solving the
OILS problem, it has been extended to BILS problems~\eqref{eq:linear-model-bils}~\cite{chang2010solving, chang2008solving, damen2003maximum}.
Sphere decoding solves the optimization problem in a sub-set of $\calX_k^n$, 
thus reduces the computational complexity compared to exhaustive search for the optimal solution~\cite{guo2006algorithm}. 
In the context of MIMO detection,
sphere decoding has been found empirically to be computationally efficient for small to moderate problem sizes (e.g., $n \leq 20$ in 16-QAM).
However, its complexity grow rapidly and becomes prohibitive as $n$ increases, the constellation order becomes larger, and/or the SNRs decreases~\cite{jalden2005complexity}.
Therefore, 
finding the optimal solution via sphere decoding is impractical for massive MIMO systems~\cite{hassibi2005sphere}.

%% file: content/relatedworks/classical_suboptimal.tex
\subsection{Non-learning-based Sub-optimal Estimators}
\label{sec:classical}
For MIMO detection, 
this section reviews classical algorithms that obtain suboptimal solutions in polynomial time and have been widely adopted in practice.

\subsubsection{Linear Estimators}
Linear estimation methods first ignore the discrete constraint and compute a continuous estimate 
$\z \in \Rbb^n$ by applying a linear transformation to the observation $\y$.
Each symbol is then estimated independently by projecting the corresponding entry of $\z$ onto the constraint set $\calX_k$:
\begin{equation}
\hat{x}_i = \text{arg} \: \underset{a \in \calX_k} {\text{min}} |z_i - a|, \quad \text{for} \quad i=1:n.
\label{eq:linear_quant}
\end{equation}
One example is the zero-forcing (ZF) estimator, which computes the relaxed least-squares estimate
\begin{equation*}
\z = (\H^\top \H)^{-1} \H^\top \y. 
\end{equation*}
The ZF estimator requires $\H^\top\H$ to be invertible and is therefore not directly applicable when the linear model is underdetermined.

Better performance can often be obtained using the minimum mean square error (MMSE) estimator.
This approach 
assumes a Gaussian prior $\calN(\0, \sigma_{\x^*}^2\I_N)$ on the transmitted symbols, where $\sigma_{\x^*}^2$ is the mean symbol energy. 
The MMSE estimator computes the linear estimate $\z$ in~\eqref{eq:linear_quant} by
\begin{equation*}
\z = \left(\H^\top \H + \frac{\sigma_{\noisevector}^2}{\sigma_{\x^*}^2}\right)^{-1} \H^\top \y.
\end{equation*}
Performance of both ZF and MMSE estimators can degrade significantly when the channel matrix $\H$ is ill-conditioned.

\subsubsection{Babai Estimators}
\label{sec:babai}
In this section, 
we introduce the Babai's nearest plane algorithm~\cite{babai1986lovasz}. 
We refer to the output of Babai's algorithm as the Babai point and the Babai estimator.
In communications,
the Babai point is often referred to as the successive interference cancellation (SIC) estimator or MMSE-SIC estimator.
Although the Babai point is originally for the recovery of integer parameter vector under the ordinary setting,
it has been extended to deal with the box-constrained setting~\cite{chang2008solving, damen2003maximum}.
In~\cite{klein2000finding}, 
Klein proposes a randomized version of Babai's nearest plane algorithm.
Given a fixed observation vector $\y$, the algorithm samples an integer vector from a discrete Gaussian-like probability distribution centering around the Babai point.
The randomness allows us to run the algorithm multiple times with the same $\y$ and use the ``best" output with the smallest residual, 
which may have better performance compared to the deterministic Babai point.
The method can readily be extended to the box-constrained case,
leading to the $K$-best box-constrained randomized Klein-Babai point~\cite{Wan25},
which will hereafter be referred to as the $K$-best Klein-Babai point for brevity.
There are other methods which find sub-optimal estimators such as the FCSD method~\cite{barbero2006performance} and the extended Babai method~\cite{chang2024extended}.

In the following we show how to compute the Babai point for the ILS problem~\eqref{eq:ils-transformed}.
Let the channel matrix $\H$ have the QR factorization $\H = \Q \R$,
where $\Q \in \Rbb^{m \times n}$ has orthonormal columns,
and $\R \in \Rbb^{n \times n}$ is an upper triangular matrix with positive diagonal entries.
With $\tby = \Q^\top \y$, the ILS problem~\eqref{eq:ils-transformed} becomes
\begin{equation*}
    \min_{\x \in \calX_k^n} \left \|\tby -\R \x \right\|^2_2. 
\end{equation*}
The Babai point, denote by $\x^{B}$,
can be computed as follows:
\begin{equation*}
\begin{split}
    & c_n = \tilde{y}_n / r_{n}, \ \
    x_n^{B} = \lfloor c_n \rceil_{{\calX_k}}, \\
    & c_i = ( \tilde{y}_i - \sum_{j = i + 1 }^n r_{ij} 
    x_j^{B} /r_{ii}, \ \ 
    x_i^{B} = \lfloor c_i \rceil_{{\calX_k}}, 
\end{split}
\end{equation*}
where $\lfloor c_i \rceil_{\calX_k}$ means rounding $c_i$ to the nearest integer in $\calX_k$ (if there is a tie, then the one with smaller value is chosen). 
To compute the Babai point corresponding to the $L_2$-regularized ILS problem~\eqref{eq:ils-transformed-l2},
the algorithm is the same, 
except that the QR factorization of the augmented matrix is used.

\subsubsection{Massage Passing Estimators}
In this section, 
we introduce another class of estimators,
known as message passing (MP) estimators.
MP methods formulate the estimation problem as probabilistic inference on the posterior distribution of the transmitted symbols~\cite{etzlinger2011message}.
Since the exact posterior is generally intractable to compute,
these methods aim to approximate the marginal posterior distribution and thereby enable feasible symbol-wise MAP estimation~\cite{cespedes2014expectation}.

A key idea in MP methods is to exploit a factor-graph representation of the posterior and then perform inference through iterative belief propagation to refine symbol estimates~\cite{yoon2013low, fukuda2013low}.
A factor graph provides a graphical representation of the factorization of a probability distribution. 
Based on this representation, 
marginals posterior distributions can be approximated by passing local messages along the edges of the factor graph~\cite{murphy2022probabilistic}.
In this way, 
the original high-dimensional inference problem is decomposed into a sequence of simpler local computations, 
which makes otherwise intractable marginal inference more manageable~\cite{wymeersch1945iterative, loeliger2004introduction}.

We next introduce the original factor graph considered in MP methods, 
which also motivates graph neural network-based approaches for MIMO detection.
Under the uniform prior assumption, 
the posterior distribution corresponds to the linear model~\eqref{eq:mimo-sys-model-real} for MIMO detection is
\begin{equation}
p(\x \,|\, \y) \propto p(\y \,|\, \x) p(\x) \propto \exp\!\left(-\frac{1}{2\sigma_{\noisevector}^2}\|\H\x-\y\|^2\right), \qquad \x \in \calX_k^n,
\label{eq:posterior}
\end{equation}
where we 
we denote $p(\x^*=\x\,|\,\y)$, $p(\y\,|\,\x^*=\x)$ and $p(\x^*=\x)$ 
by $p(\x\,|\,\y)$, $p(\y\,|\,\x)$ and $p(\x)$,
respectively,
to simplify the notation. 
By expanding the quadratic term, the posterior~\eqref{eq:posterior} can be factorized into a product of single-variable and pairwise factors:
\begin{align}
p(\x \,|\, \y) & \propto \prod_i \phi_i(x_i) \prod_{i < j}  \phi_{ij}(x_i, x_j), \nonumber \\  
\phi_i(x_i) &= \text{exp} \left( \frac{1}{\sigma_{\noisevector}^2} \y^\top \h_i x_i - \frac{1}{2 \sigma_{\noisevector}^2} \h_i^\top \h_i (x_i)^2 \right), \nonumber \\ 
\phi_{ij}(x_i, x_j) &= \text{exp} \left( - \frac{1}{\sigma_{\noisevector}^2} \h_i^\top \h_j x_ix_j \right), 
\label{eq:mimo_posterior_mrf}
\end{align}
where $\h_i$ is the $i^{th}$ column of $\H$,
$\phi_i(x_i)$ and $\phi_{ij}(x_i, x_j)$ are referred to as the self-potential and pairwise potential, respectively.
Because the factorization in~\eqref{eq:mimo_posterior_mrf} involves only single and pairwise variable terms, 
it also defines a Markov random field (MRF) representation of the posterior~\cite{yedidia2003understanding} and is referred to as the MRF graph.
In this MRF graph, each variable node represents one transmitted symbol, 
and each edge represents statistical dependence between a pair of symbols.
MP methods operate on this graphical model by iteratively exchanging local summaries, called messages, between neighboring nodes.
The messages are typically initialized using the uniform prior probabilities over $\calX_k$.
These messages are then combined to estimate the marginal posterior probabilities of the transmitted symbols.

Among MP methods, 
belief propagation (BP)
or equivalently the sum-product algorithm on factor graphs, is the most widely used baseline.
Many subsequent MP methods can be viewed as extensions of BP.
For a tree-structured graph (cycle-free), 
BP computes exact marginal distributions~\cite{yedidia2003understanding}.
However, 
a straightforward application of BP to the MIMO detection problem performs poorly, 
because the underlying factor graph contains many short cycles~\cite{shental2008discrete}. 
Therefore, a major line of work on MP-based MIMO detection modifies either the factor graph or the message-update rules to improve performance.

One approach is to find an approximation of the posterior~\eqref{eq:posterior} whose underlying factor graph is tree-structured so that BP can be applied effectively. 
However, 
finding the optimal tree approximation is itself NP hard~\cite{goldberger2011mimo}.
The Gaussian tree approximation (GTA) is proposed to address this issue, 
which replaces the original posterior with a tree-structured Gaussian approximation and then applies BP on the resulting graph~\cite{goldberger2009gaussian}.
This idea was later extended and improved by incorporating successive interference cancellation, 
leading to GTA-SIC~\cite{goldberger2013improved}.
Both GTA and GTA-SIC can be implemented with a complexity of $O(n^3)$.

A second line of work sorts to replace exact BP messages by tractable approximate messages. 
Expectation propagation (EP) provides a general-purpose framework to construct a tractable approximate to an intractable distribution~\cite{minka2013expectation, seeger2007bayesian, opper2005expectation}.
In its application to MIMO detection,
EP approximates the true posterior by a Gaussian distribution whose parameters are refined iteratively~\cite{cespedes2014expectation}. 
Compared with BP, 
EP no longer passes full discrete potential along graph edges. 
Instead, it iteratively updates local approximate factors and use the discrete constraint during updating. 
The details of the EP will be discussed in Chapter~\ref{ch:Paper03}, Section~\ref{sec:ep}.
For EP with $T$ iterations, its total complexity is $O(n^3 T)$.
EP with 10 iterations yields a remarkable performance and outperforms GTA-SIC on high-order and high-dimensional MIMO systems, 
and no improvement can be observed between $T = 10$ and $T = 100$~\cite{cespedes2014expectation}. 

Compared with EP, 
approximate message passing (AMP) and its variants have lower per-iteration complexity.
However, the performance of AMP can be far from optimal in practical finite-dimensional MIMO systems and may exhibit an error floor, 
particularly for high-order constellations~\cite{bayati2011dynamics}.
The performance of AMP can also degrade when the channel matrix is ill-conditioned.
Orthogonal AMP (OAMP)~\cite{ma2017orthogonal} is proposed to improve the robustness and performance of AMP, 
especially for ill-conditioned channel matrices.

Empirical studies have shown that EP achieves strong performance among classical estimators for massive MIMO detection with high-order constellations~\cite{cespedes2014expectation, kosasih2022graph}.
MP methods have also motivated subsequent developments in GNN-based estimators~\cite{scotti2020graph} and GNN-aided MP estimators~\cite{kosasih2022graph, he2023gnn}, 
which are reviewed in Section~\ref{sec:graph_learning_based_detection}.

%% file: content/relatedworks/learning_based_suboptimal.tex
\subsection{Machine Learning-based Approaches for Estimation}
\label{sec:learning_based_detection}
To improve the performance of conventional suboptimal estimators, 
machine learning-based methods have emerged as promising solutions for meeting the high-throughput and low-latency requirements of massive MIMO systems~\cite{scotti2020graph, pratik2020re, he2018model, Neev2019learning, Nhan2020deep, kosasih2022graph, kosasih2022graph2, he2023gnn, liu2023model}.
This section reviews recent advances in machine learning-based methods for MIMO detection.


\subsubsection{Deep unfolding-based methods}
Within the broader family of learning-based methods for optimization problems,
a prominent class of approaches for MIMO detection follows the deep unfolding paradigm, which corresponds to the second paradigm discussed in Section~\ref{sec:learning_to_optimize}.
Representative examples include DetNet~\cite{samuel2017deep} and OAMPNet~\cite{he2018model}.

These methods unfold a classical iterative algorithm into a neural network, 
where each layer corresponds to one iteration of the original algorithm.
Some fixed scalar parameters or algorithmic components in the original method are then replaced by learnable ones.
Specifically,
DetNet~\cite{samuel2017deep} unfolds the iterations of a projected gradient descent algorithm into a neural network.
It has been shown to be effective on i.i.d. Gaussian channels with low-order constellations, such as 4-QAM.
Compared with AMP, DetNet is more robust to ill-conditioned channels.
However, its performance degrades on correlated channels and higher-order constellations~\cite{pratik2020re}.
OAMPNet~\cite{he2018model} is designed by unfolding iterations of the OAMP algorithm~\cite{ma2017orthogonal}.
It improves upon its parent OAMP by introducing learnable parameters into the unfolded iterations.


\subsubsection{GNN-based methods}
\label{sec:graph_learning_based_detection}
This section reviews recent work on GNN-based methods for MIMO detection.

The use of GNNs for MIMO detection was first proposed in~\cite{scotti2020graph}, 
motivated by the pairwise Markov random field (MRF) representation of the posterior in~\eqref{eq:mimo_posterior_mrf}.
This pairwise MRF models the structured dependence among the random variables in $\x^*$ using a fully connected undirected graph with self-loops, 
denoted by $\calG=(\calV,\calE)$.
Each node $v_i \in \calV$ is associated with the $i^{th}$ transmitted symbol variable $x^*_i$.
The self-potential and pairwise-potential terms in~\eqref{eq:mimo_posterior_mrf} are incorporated into the initial node and edge features of the GNN.
Specifically, 
for node $v_i$ and edge $e_{ij}$ between nodes $v_i$ and $v_j$, 
the features are initialized as
\begin{equation}
\noisevector_i = [\y^\top \h_i, \h_i^\top \h_i, \sigma_{\noisevector}^2]^\top, 
\quad 
\e_{ij} = [-\h_i^\top \h_j, \sigma_{\noisevector}^2]^\top.
\label{eq:mrf_mimo_graph_feature}
\end{equation}
The GNN estimator proposed in~\cite{scotti2020graph} outperforms the MMSE estimator and BP, 
but underperforms some MP estimators~\cite{he2023message}.
This idea was later extended to GNN-aided MP estimators.

In GNN-aided MP estimators, 
GNNs are used to improve posterior approximation within various MP algorithms, 
including expectation propagation (EP) \cite{cespedes2014expectation}, 
approximate message passing (AMP) \cite{donoho2009message} and Bayesian parallel interference cancellation \cite{kosasih2021bayesian}.
The GNN-aided EP (GEPNet) \cite{kosasih2022graph} and its variant IEP-GNN \cite{liu2023model} achieve the best performance classical sub-optimal estimators and those which apply machine learning techniques.

We briefly summarize the development of GNN-aided MP estimators.
GEPNet is a pioneering framework that incorporates GNNs into MP~\cite{kosasih2022graph}.
However, its EP module requires matrix inversion, 
which leads to high computational cost with multiple iterations.
To improve efficiency, 
later work replaces the EP module in GEPNet with lower-complexity MP methods. 
For example, GPICNet \cite{kosasih2022graph2} substitutes the EP module with BPIC,
and AMP-GNN \cite{he2023gnn} replaces EP by AMP.
Although these modifications reduce computational complexity, 
they also degrade the performance of GEPNet.
More recently, 
IEP-GNN~\cite{liu2023model} is proposed to improve the computation of local posterior information in the EP iterations.
This approach improves the performance of GEPNet while maintaining the same computational complexity.


%% file: content/Paper02/Paper02.tex
\chapter{\rm\bfseries Flexible Diffusion Scopes with Parameterized Laplacian for Heterophilic Graph Learning}
\label{ch:Paper02}

This chapter focuses on semi-supervised learning with GNNs for node classification.
The motivations and contributions of this chapter are explained as follows.

In GNNs, 
higher-order neighbors are recursively incorporated through graph diffusion across multiple stacked layers. 
At each layer, 
unlike CNNs whose convolutional kernels assign different learnable weights to different spatial positions,
GNNs such as GCN~\cite{kipf2016semi} and many other popular variants~\cite{hamilton2017inductive, xu2018powerful} aggregate neighboring nodes using either uniform weights or fixed weights determined by commonly used normalizations of adjacency or Laplacian matrices.

Recent studies show that the optimal choice of a normalized Laplacian is data-dependent~\cite{dasoulas2021learning, dall2020optimal}. 
Aggregation with learnable weights has also been proposed to improve the expressiveness of GNNs~\cite{eliasof2023improving}. 
However, conventional local aggregation still has limited access to long-range neighbors and global graph structure, which can prevent it from achieving optimal performance~\cite{gasteiger2019diffusion}. 
For example,
although GAT~\cite{velivckovic2017graph} assigns learnable weights through an attention mechanism, 
each layer generates node representations using only direct neighbors. 
This local aggregation mechanism works well on homophilic graphs~\cite{luan2022we, luan2024graph}, 
where connected nodes are likely to share the same label.
However, because of its limited and fixed diffusion scope, GAT can suffer substantial performance degradation on node classification tasks for heterophilic graphs~\cite{zhu2020beyond,luan2022complete, luan2022revisiting}, 
where nodes from different classes are often connected.

Non-local neighborhood information has been shown to be useful for addressing heterophily in GNNs~\cite{abu2019mixhop,liu2020non,zhu2020beyond,he2021bernnet,luan2024heterophilic, li2024permutation, huang2024universal}.
However, its interaction with the design of graph Laplacians and graph diffusion scopes remains insufficiently explored.
We address this issue in this chapter. 

Section~\ref{sec:paper02_pre} provides the preliminaries for this chapter and discusses related work.
The main contributions of this chapter are as follows.  
In Section~\ref{sec:para-lap},
we propose a new class of parameterized normalized graph Laplacian matrices. 
These matrices provide better control over the graph diffusion process and include several widely used normalized Laplacians as special cases. 
Through the parameters of the proposed Laplacian,
we can adjust the diffusion and spectral distances between nodes by altering the spectral properties of the graph. 
This enables flexible message passing that adaptively captures both local and global information.
We establish a theorem showing that spectral distance can serve as a surrogate for diffusion distance, 
which significantly reduces the computational cost of comparing relative distances between nodes to identify long-range neighborhood. 

In Section~\ref{sec:pd}, based on the new graph Laplacian and the theorem, 
we propose two architectures, 
Parameterized Diffusion-based Graph Convolutional Networks (PD-GCN) and 
Parameterized Diffusion-based Graph Attention Networks (PD-GAT).
Both architectures enable flexible diffusion on graph via the proposed Laplacian. 

In Section~\ref{sec:rewire}, we introduce a topology-guided rewiring mechanism to facilitate long-range message passing, 
which further improves learning on heterophilic graphs. 
The method uses the first non-trivial eigenvector of the parameterized Laplacian to identify long-range neighbors and rewires locally disconnected nodes. 

Section~\ref{sec:experiments} presents empirical results demonstrating the effectiveness and superiority of the proposed models compared with state-of-the-art (SOTA) GNNs on semi-supervised node classification tasks across graphs with varying levels of homophily, 
especially on heterophilic graphs. 
The proposed strategy is flexible and can be seamlessly integrated with other types of graph aggregation.



\input{content/Paper02/chapter_preliminary}



\section{Parameterized Normalized Laplacian}
\label{sec:para-lap}
In order to gain a more refined control over the diffusion on a graph, we define a new class of Laplacian matrices, denoted by $\mathbf{L}^{(\alpha,\gamma)}$, where the two parameters $\gamma$ and $\alpha$ determine its spectral properties. Then, we propose an efficient method to compare the relative diffusion distances.

\begin{definition}
\label{def:aug-lap}
A parameterized normalized Laplacian matrix is defined as
\begin{equation*}
\mathbf{L}^{(\alpha,\gamma)} 
= \gamma [\gamma \D + (1-\gamma )\I]^{-\alpha} \mathbf{L}
[\gamma \D + (1-\gamma )\I]^{\alpha-1}
\end{equation*}
and the corresponding parameterized normalized adjacent matrix  
is defined as
\begin{equation} \label{def:adj}
\P^{(\alpha,\gamma)} =  \I-\mathbf{L}^{(\alpha,\gamma)},
\end{equation}
where the parameters $\gamma \in (0, 1]$ and $\alpha \in [0, 1]$. 
\end{definition}

The normalized Laplacian matrices defined in Section~\ref{sec:pre_grl} are special cases of this new class of Laplacian matrices.
Specifically,
when $\alpha=1$ and $\gamma=1$, $\mathbf{L}^{(\alpha,\gamma)}=\mathbf{L}_\rw$; when $\alpha=\frac{1}{2}$ and $\gamma=1$, $\mathbf{L}^{(\alpha,\gamma)}=\mathbf{L}_\sym$.  
Although we cannot choose $\alpha$ and $\gamma$ such that 
$\mathbf{L}^{(\alpha,\gamma)}$ becomes $\mathbf{L}$, 
we have the following result:
\begin{equation*}
\lim_{\gamma \to 0} \frac{1}{\gamma}\mathbf{L}^{(\alpha, \gamma)} = \lim_{\gamma \to 0} (\gamma \D + (1-\gamma )\I)^{-\alpha} \mathbf{L} \times \lim_{\gamma \to 0} (\gamma \D + (1-\gamma )\I)^{\alpha-1} = \mathbf{L}, 
\end{equation*}
which implies that when $\gamma$ is small enough, 
an eigenvector of $\mathbf{L}^{(\alpha,\gamma)}$ is a good approximation to
an eigenvector of $\mathbf{L}$.
The following theorem justifies the definition of $\P^{(1,\gamma)}$ as a random walk matrix,
which generalizes the classic random walk matrix $\D^{-1}\A$.
\begin{theorem}
\label{thm:parameterized-matrix}
The $\P^{(\alpha, \gamma)}$ defined in \eqref{def:adj} is non-negative (\ie{} all of its elements are non-negative), and $\P^{(1, \gamma)}\1= \1$. 
See the proof in Appendix~\ref{appendix:proof_parameterized_matrix}.
\end{theorem}

The properties of eigenvalues of $\mathbf{L}^{(\alpha,\gamma)}$ 
are given in the following theorem.

\begin{theorem}
\label{thm:eigenplaplaican}
    Suppose the graph $\mathcal{G}$ is connected.
    Then the symmetric $\mathbf{L}^{(1/2,\gamma)}\in \Rbb^{N\times N}$ has the eigendecomposition:
    \begin{equation}
    \label{eq:gled}
    \mathbf{L}^{(1/2,\gamma)} = \U \bLambda^{(\gamma)} \U^\top,
    \end{equation}
    where $\U\in \Rbb^{N\times N}$ is orthogonal and $\bLambda^{(\gamma)}=\diag([\lambda_1^{(\gamma)}, \ldots, \lambda_N^{(\gamma)}]^\top)$,
    \begin{equation} \label{eq:glevrange}
    0 =\lambda_1^{(\gamma)} < \lambda_2^{(\gamma)}
    \leq \cdots \leq \lambda_N^{(\gamma)} \leq 2.
    \end{equation}
Each $\lambda_i^{(\gamma)}$ is strictly increasing with respect to $\gamma$ for $i=2:N$.
Furthermore, $\mathbf{L}^{(\alpha,\gamma)}$ has the eigendecomposition:
\begin{equation} \label{eq:gled-alpha}
\mathbf{L}^{(\alpha,\gamma)} 
= \left([\gamma\D+(1-\gamma)\I]^{\frac{1}{2}-\alpha} \U \right) \bLambda^{(\gamma)} 
\left([\gamma\D+(1-\gamma)\I]^{\frac{1}{2}-\alpha}\U\right)^{-1} ,
\end{equation}
\ie{} $\mathbf{L}^{(\alpha,\gamma)}$ share the same eigenvalues as $\mathbf{L}^{(1/2,\gamma)}$ and the columns of $[\gamma\D+(1-\gamma)\I]^{\frac{1}{2}-\alpha}\U$ are the corresponding eigenvectors. See proof in Appendix~\ref{appendix:proof_eigenplaplaican}.
\end{theorem}

The following theorem shows the monotonicity of the diffusion distance with respect to the spectral distance under certain conditions. 
Here, both distances are defined using the eigenvectors of
$\mathbf{L}^{(1,\gamma)}$,
cf.\ \eqref{def:diffusion-dist} and \eqref{def:spectral-dist},
which involve the eigenvectors of $\mathbf{L}_\rw$.

\begin{theorem}
\label{thm:gradient-v2}
Let $v_i$, $v_j$ and $v_m$ be nodes of the graph $\mathcal{G}$.
Suppose that the spectral distance from $v_m$ to $v_j$ is smaller than that from $v_i$ to $v_j$, that is,
\begin{equation*}
d_s(v_m,v_j;\mathbf{L}^{(1,\gamma)})
<
d_s(v_i,v_j;\mathbf{L}^{(1,\gamma)}).
\end{equation*}
Then there exists a constant $C$ such that,
for all $t\geq C$, 
the diffusion distances satisfy
\begin{equation*}
    d_t(v_m, v_j; \mathbf{L}^{(1,\gamma)}) < d_t(v_i, v_j; \mathbf{L}^{(1,\gamma)}),
\end{equation*}
with the reduction in distance being proportional to
to $e^{-\lambda_2^{(\gamma)}}$. 
See proof in Appendix~\ref{appendix:gradient-v2}.
\end{theorem}

Spectral distance and diffusion distance can also be defined using general graph Laplacians that do not necessarily correspond to random-walk dynamics~\cite{lipman2010biharmonic, kreuzer2021rethinking}.
For diffusion distance and spectral distance defined by $\mathbf{L}^{(\alpha,\gamma)}$, 
the monotonicity demonstrated in Theorem \ref{thm:gradient-v2} also holds.
\footnote{Theorem \ref{thm:gradient-v2} substantially extends the scope of~\cite{beaini2021directional}, 
which deals with the diffusion distance defined by $\mathbf{L}_{\rw}$. Furthermore, Theorem \ref{thm:gradient-v2} overcomes some shortcomings of~\cite{beaini2021directional}, see details in Appendix~\ref{appendix:gradient-v2}.}
Therefore, the spectral distance can be used as a good indicator of the diffusion distance.
That is, if node $v_i$ has a larger spectral distance to node $v_j$ than node $v_m$ does, 
then node $v_i$ also has a larger diffusion distance to node $v_j$ after sufficiently many diffusion steps.

According to~\eqref{def:diffusion-dist}, 
computing the diffusion distance requires all eigenvalues and eigenvectors of the Laplacian, 
which is computationally expensive.
In contrast, Theorem~\ref{thm:gradient-v2} shows that the spectral distance, which only depends on the first nontrivial eigenvector, can be used as a surrogate for the diffusion distance.
This provides a more efficient way to encode global structural information into local aggregation.

Since a smaller diffusion distance implies stronger influence between two nodes in graph message passing, 
DGN~\cite{beaini2021directional} defines directional aggregation operators that incorporate diffusion-based structural information.
However, DGN requires the first $k$ eigenvectors of $\mathbf{L}_{\rw}$, 
where a larger $k$ is preferred for better capturing of the actual diffusion process.
By contrast, our approach encodes global structural information using the spectral distance, 
which only involves the first nontrivial eigenvector.

Furthermore,
$\mathbf{L}^{(\alpha,\gamma)}$ provides flexible control over graph message passing.
Graphs with different homophily levels may benefit from neighborhood information at different diffusion ranges.
On homophilic graphs, nearby nodes with small diffusion distance are often more informative.
On heterophilic graphs, useful information may come from more distant nodes, and incorporating long-range neighborhood information can help reduce the effect of noisy local aggregation~\cite{topping2021understanding}.
Adjusting $\alpha$ and $\gamma$ changes the eigenvalues and eigenvectors of $\mathbf{L}^{(\alpha,\gamma)}$, 
thereby modifying the corresponding diffusion and spectral distances.
Therefore, $\alpha$ and $\gamma$ can be tuned as hyperparameters to improve the expressiveness of GNNs according to the homophily level of the graph.

\section{Parameterized Diffusion Augmented GNNs}
\label{sec:pd}
We present the proposed methods, 
which aim to adapt the graph diffusion process to different homophily levels.
Specifically, we introduce two parameterized diffusion-augmented GNNs,
Parameterized Diffusion-based GCN (PD-GCN) and Parameterized Diffusion-based GAT (PD-GAT).
Both models achieve flexible diffusion through the proposed parameterized normalized Laplacian 
$\mathbf{L}^{(\alpha,\gamma)}$.
These models address limitations of message passing based on graph structures encoded by conventional Laplacians, 
and can accelerate the diffusion of long-range information in heterophilic graphs.

\subsection{Parameterized Diffusion-based GCN (PD-GCN)}
As stated in Section~\ref{sec:spectral_gnn},
the one-hop neighborhood aggregation in GCN~\cite{kipf2016semi} is a first-order approximation of the graph convolution using a polynomial filter of the graph Laplacian~\cite{defferrard2016convolutional}.
Accordingly, 
we propose Parameterized Diffusion based Graph Convolutional Networks (PD-GCN),  
where the aggregation utilizes the parameterized normalized adjacent matrix $\P^{(\alpha,\gamma)}$ defined in~\eqref{def:adj} instead.

At the $\ell^{th}$ layer, 
PD-GCN updates node representations as
\begin{equation*}
\mathbf{X}^{(\ell)} = \mathrm{\sigma} (\P^{(\alpha,\gamma)} \mathbf{X}^{(\ell-1)} \mathbf{W}^{(\ell)}),
\end{equation*}
where $\P^{(\alpha,\gamma)}$ provides the aggregation weights for message passing.

While the vanilla GCN sets the weights solely based on paired nodes degrees, 
PD-GCN is able to encode global information into the local aggregation with its flexible diffusion scopes.
This improves the expressiveness of GCN for the following reason.
As shown in Theorem~\ref{thm:gradient-v2}, 
the first nontrivial eigenvector of the graph Laplacian provides information about the diffusion distance between nodes and therefore indicates the dominant direction of diffusion.
However, homophily and heterophily are label-dependent properties.
As a result, a fixed conventional Laplacian may not induce an optimal diffusion process for graphs with different homophily levels.
By introducing the parameterized Laplacian $\mathbf{L}^{(\alpha,\gamma)}$, 
PD-GCN can adjust the diffusion scope through suitable choices of $\alpha$ and $\gamma$, 
thereby enabling more effective message propagation across graphs with different homophily levels.

\subsection{Parameterized Diffusion-based GAT (PD-GAT)}

In GAT~\cite{velivckovic2017graph}, 
the learnable aggregation weights are determined by local node features and do not adapt to global topological characteristics.
PD-GAT addresses this limitation by incorporating parameterized edge features into the attention mechanism.

Let $\u^{(\alpha,\gamma)}$ denote the first nontrivial eigenvector of $\mathbf{L}^{(\alpha,\gamma)}$.
For each edge $e_{ij}\in\mathcal{E}$, 
we define its edge feature as
\begin{equation*}
\f^{(\alpha, \gamma)}_{ij} = \big[ \mathbf{B}_{\mathrm{av}}(\u^{(\alpha, \gamma)})_{ij}, \mathbf{B}_{\mathrm{dx}}(\u^{(\alpha, \gamma)})_{ij} \big]^\top,
\end{equation*}
where 
$\mathbf{B}_{\mathrm{av}}(\u^{(\alpha, \gamma)})$ and $\mathbf{B}_{\mathrm{dx}}(\u^{(\alpha, \gamma)})$ are matrices corresponding to the aggregation and diversification operations defined from the vector field $\nabla\u^{(\alpha, \gamma)}$
(see \eqref{eq:directedagg}).
The entries of these matrices encode diffusion-based relative positional information between nodes $v_i$ and $v_j$.
Instead of directly using these quantities as aggregation weights, as in DGN~\cite{beaini2021directional}, 
we use them as edge features to guide the learning of attention weights between nodes.

At layer $\ell$, we define the unnormalized attention score between nodes $v_i$ and $v_j$ as
\begin{equation}
\label{eq:dgat_attention}
r_{ij}^{(\ell, \alpha, \gamma)}= \text{LeakyReLU}((\mathbf{a}^{(\ell)})^{\top} [\mathbf{W}_n^{(\ell)} \x_i^{(\ell-1)} || \mathbf{W}_n^{(\ell)} \x_j^{(\ell-1)} || \mathbf{W}_e^{(\ell)} \f^{(\alpha, \gamma)}_{ij} ]),
\end{equation}
where
$\mathbf{W}_n^{(\ell)}$ and $\mathbf{W}_e^{(\ell)}$ are learnable weights for node representation and edge features, respectively,
$\mathbf{a}^{(\ell)}$ is a learnable attention vector.

We use $\alpha_{ij}^{(\ell, \alpha, \gamma)}$ to denote the attention coefficient after softmax normalization.
With multi-head attention, PD-GAT computes the updated node representation as
\begin{equation*}
\x_i^{(\ell)} =\left( \bigg\Arrowvert^{M}_{m=1} \sigma \big( 
\sum_{j \in \mathcal{N}(i)} \alpha_{ij}^{(\ell, \alpha, \gamma, m)} \W^{(\ell, m)}_n \x_j^{(\ell-1)}
\big)   \right)\W^{(\ell)}
\end{equation*}
where $M$ is the number of attention heads,
and $\sigma(\cdot)$ is a nonlinear element-wise activation function.
Compared with GAT, the additional number of parameters introduced by PD-GAT in~\eqref{eq:dgat_attention} is negligible, 
since $\f_{ij}^{(\alpha,\gamma)} \in \Rbb^2$.
And $\u^{(\alpha,\gamma)}$ can be computed with complexity $O(|\calE|)$, as detailed in Appendix~\ref{appendix:compute_nontrivial_eigenvector}.

\section{Topology-Guided Rewiring Mechanism for Long-Range Diffusion}
\label{sec:rewire}

This section introduces a graph rewiring technique to further enhance the proposed parameterized diffusion on heterophilic graphs.
Since diffusion distance measures how efficiently information can propagate between nodes,
we regard locally disconnected nodes with large diffusion distance as long-range neighbors.
According to Theorem~\ref{thm:gradient-v2}, large spectral distances can be used to identify such long-range neighbors.
Therefore, the first nontrivial eigenvector $\u^{(\alpha,\gamma)}$ of $\mathbf{L}^{(\alpha, \gamma)}$ can be regarded as a one-dimensional embedding of nodes.
We refer to $\u^{(\alpha,\gamma)}$ as the parameterized spectral embedding.

Let $v^{(\alpha,\gamma)}$ denote the gradient node, 
defined as the node whose corresponding entry in $\u^{(\alpha,\gamma)}$ is the largest that
\begin{equation*}
v^{(\alpha,\gamma)}
=
\argmax_{v_i \in \mathcal{V}}
\u^{(\alpha,\gamma)}(i).
\end{equation*}
Equivalently, $v^{(\alpha,\gamma)}$ is the rightmost node in the parameterized spectral embedding.
For heterophilic graphs, 
the proposed rewiring strategy connects nodes in the graph to the gradient node under a given configuration of $\mathbf{L}^{(\alpha,\gamma)}$.
The motivation for this strategy is described below.

On heterophilic graphs, 
useful information may come from distant nodes rather than immediate neighbors.
Therefore, the proposed rewiring strategy aims to create shortcuts that facilitate long-range diffusion.
We use the one-dimensional spectral embedding $\u^{(\alpha,\gamma)}$ to identify nodes that are far apart in graph diffusion.

Consider the message passing to a target node $v_i$.
We first define its long-range candidate neighbors.
Without loss of generality, 
we consider nodes that lie to the right of $v_i$ in the spectral embedding and restrict the candidates to nodes whose spectral distance from $v_i$ is sufficiently large, that is, nodes $v_j$ satisfying
\begin{equation*}
\u^{(\alpha,\gamma)}(j)-\u^{(\alpha,\gamma)}(i)
\geq
\frac{1}{2}
\left(
\max_m \u^{(\alpha,\gamma)}(m)
-
\min_m \u^{(\alpha,\gamma)}(m)
\right).
\end{equation*}
Such nodes $v_j$ are regarded as long-range neighbors of $v_i$.

The rewiring strategy says,
if $v_i$ and $v^{(\alpha,\gamma)}$ are not connected in the original graph, 
we add an edge between them,
and this new edge creates a shortcut for the long-range message passing from $v_j$ to the target node $v_i$.
Since the gradient node is the rightmost node in the spectral embedding, 
$v_j$ is closer to $v^{(\alpha,\gamma)}$ than to $v_i$ in spectral distance.
By Theorem~\ref{thm:gradient-v2}, for sufficiently large diffusion time $t$, this implies
\begin{equation*}
d_t(v_j,v^{(\alpha,\gamma)};\mathbf{L}^{(\alpha,\gamma)})
<
d_t(v_j,v_i;\mathbf{L}^{(\alpha,\gamma)}).
\end{equation*}
Therefore, after rewiring, information from these long-range nodes can propagate to $v_i$ more efficiently through the newly connected gradient node.
In this way, the rewiring mechanism accelerates long-range diffusion and improves message passing on heterophilic graphs while maintaining the sparsity of the original graph.


\section{Experiments}
\label{sec:experiments}

In this section, we evaluate the effectiveness of our proposed models on synthetic and real-world benchmark datasets. 

All experiments follow the semi-supervised learning setting for transductive node classification (see Section~\ref{sec:grl_setting}). 
For a graph $\mathcal{G}=(\mathcal{V},\mathcal{E})$ with $N$ nodes and labels vector
$\z \in L^N$,
where $L = \{1,\ldots,C\}$ denotes the set of $C$ class labels, 
each node $i\in\mathcal{V}$ has label $z_i$.
We denote the network output by $\Y\in\Rbb^{N\times C}$, 
where $\Y(i, c)$ represents the predicted probability that node $i$ belongs to class $c$.
Let $\mathbf{Z}=\1_L(\z)\in\Rbb^{N\times C}$ denote the one-hot encoding of the labels, as defined in~\eqref{eq:one-hot}.
Let ${\cal V}_{\mathrm{train}}$ denote the set of training nodes.
The training objective is defined as the cross-entropy loss over ${\cal V}_{\mathrm{train}}$:
\begin{equation*}
\mathcal{L}
=
-\frac{1}{|\mathcal{V}_{\mathrm{train}}|}
\sum_{i\in\mathcal{V}_{\mathrm{train}}}
\sum_{c=1}^{C}
\mathbf{Z}_{ic} \log \mathbf{Y}_{ic},
\end{equation*}
where $\log(\cdot)$ is a component-wise logarithm operation.

Since $\alpha$ and $\gamma$ are currently treated as dataset level hyperparameters, selecting them requires an additional grid search. 
This introduces extra tuning cost beyond that of the baseline models and constitutes a limitation of the current approach. 
Jointly learning these parameters is therefore left for future work.

In Section \ref{sec:ablation}, 
we conduct ablation study to validate the effectiveness of each proposed component. 
In Section \ref{sec:synthetic}, 
we generate synthetic graphs to study how the performance of the proposed parameterized diffusion varies with the graph homophily levels. 
In Section \ref{sec:benchmark}, 
we compare the proposed architectures with baselines and state-of-the-art (SOTA) models on real-world datasets.
The results show that parameterized diffusion augmented GNNs outperform the SOTA models on 6 out of 7 node classification tasks under semi-supervised learning. 
These results highlight the explainability of 
proposed parameterized diffusion based on $\mathbf{L}^{(\alpha,\gamma)}$, 
and verify that the proposed GNNs indeed can adaptively capture global information by adjusting parameters in $\mathbf{L}^{(\alpha, \gamma)}$ for graphs with different levels of homophily.

\subsection{Ablation Study}
\label{sec:ablation}

\input{content/Paper02/table/table_result_heter_ablation}
\input{content/Paper02/table/table_virtual_node}

The ablation study aims to investigate the effectiveness of (1) parameterized normalized Laplacian $\mathbf{L}^{(\alpha, \gamma)}$ and its corresponding adjacency matrix $\mathbf{P}^{(\alpha, \gamma)}$, 
(2) the topology-guided graph rewiring strategy,
and (3) the parameterized diffusion based GCN and GAT, referred to as PD-GCN and PD-GAT, respectively.
The ablation results on 7 heterophilic datasets are summarized in Table~\ref{tab:heter_ablation}. Note that both the parameterized diffusion and the rewiring step are influenced by $\alpha$ and $\gamma$. 
The optimal $\alpha$ and $\gamma$ used in the ablation studies are provided in Appendix~\ref{appendix:Hyperparameters}.
Table~\ref{table:heterophily_virtual_node}
provides a comparison with message passing over graphs with virtual nodes~\cite{ishiguro2019graph}, 
as an additional ablation study on the effectiveness of the rewiring strategy.

From Table \ref{tab:heter_ablation}, 
we make the following observations:
\begin{itemize}
\item The rewiring strategy improves both baseline models and the proposed models.
\item Comparing PD-GAT and GAT,
even without leveraging $\mathbf{L}^{(\alpha, \gamma)}$, \ie{} using only the default $\mathbf{L}_{\rw}$ with $\alpha=1$, $\gamma=1$, 
both the rewiring strategy and the parameterized diffusion can individually improve model performance.
\item PD-GCN is more sensitive to the choice of parameters $\alpha$ and $\gamma$.
PD-GCN with the default parameters achieves performance gain over GCN, 
except on the \textit{questions} and \textit{squirrel-filtered}.
However, with optimal parameters,
PD-GCN even outperforms PD-GAT on these two datasets.
\item After incorporating $\mathbf{L}^{(\alpha, \gamma)}$,
\ie{} perform a grid-search for the optimal $\alpha$, $\gamma$ for each graph to find the most suitable relative node positions, 
the effectiveness of both parameterized diffusion and the rewiring strategy is further enhanced.
\item Each component is indispensable for the success of PD-GCN and PD-GAT,
except for the \textit{amazon-ratings} dataset, 
where the rewiring does not provide a benefit\footnote{It can be explained by the results in~\cite{platonov2023critical}, which show that the GNNs yield negligible improvement over the graph-free models on \textit{amazon-ratings}.}.
\end{itemize}

We further demonstrate the effectiveness of parameterized diffusion with rewiring by comparing it with message passing using virtual nodes~\cite{ishiguro2019graph,hu2020open}.
The results are summarized in Table~\ref{table:heterophily_virtual_node}.
A virtual node is an additional node attached to the original graph and connected to all nodes.
In contrast, our rewiring strategy connects all nodes to the gradient node determined by $\mathbf{L}^{(\alpha,\gamma)}$.
The difference between the proposed rewiring strategy and the virtual-node approach is threefold: (1) the gradient node is selected from the original graph, whereas the virtual node is artificially added; 
(2) the selection of the gradient node is determined by $\alpha$ and $\gamma$;
and (3) the virtual node is incompatible with the proposed parameterized diffusion, 
as we cannot define the weights or features for edges connecting the virtual nodes based the first nontrivial eigenvector of $\mathbf{L}^{(\alpha, \gamma)}$ from the original graph topology.

Table~\ref{table:heterophily_virtual_node} shows that the proposed rewiring approach yields a larger performance improvement over the baselines compared to the virtual node method.






\begin{figure*}[t]
\centering
 {  {
 \subfloat[]{
 \captionsetup{justification = centering}
 \includegraphics[width=0.5\textwidth]{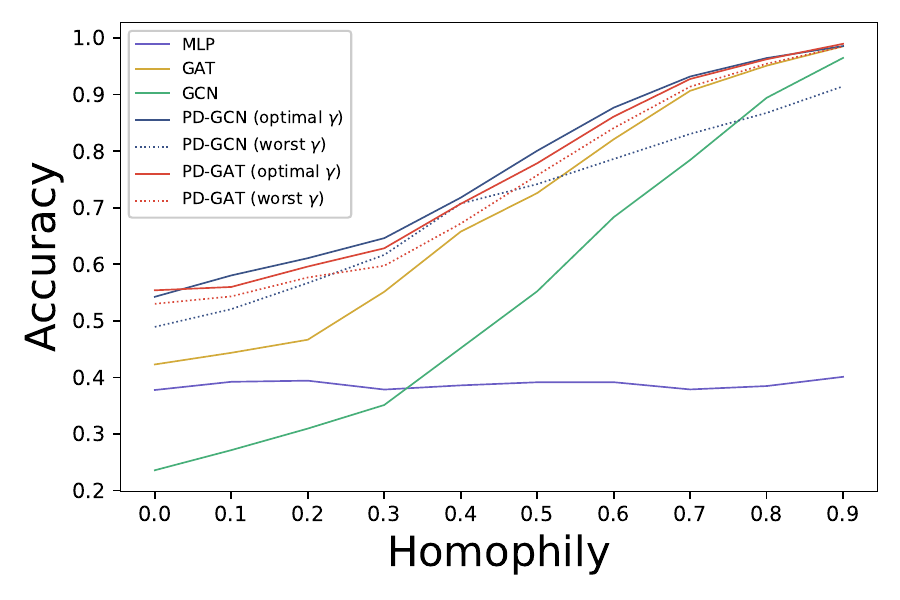} }
 \subfloat[]{
 \captionsetup{justification = centering}
 \includegraphics[width=0.5\textwidth]{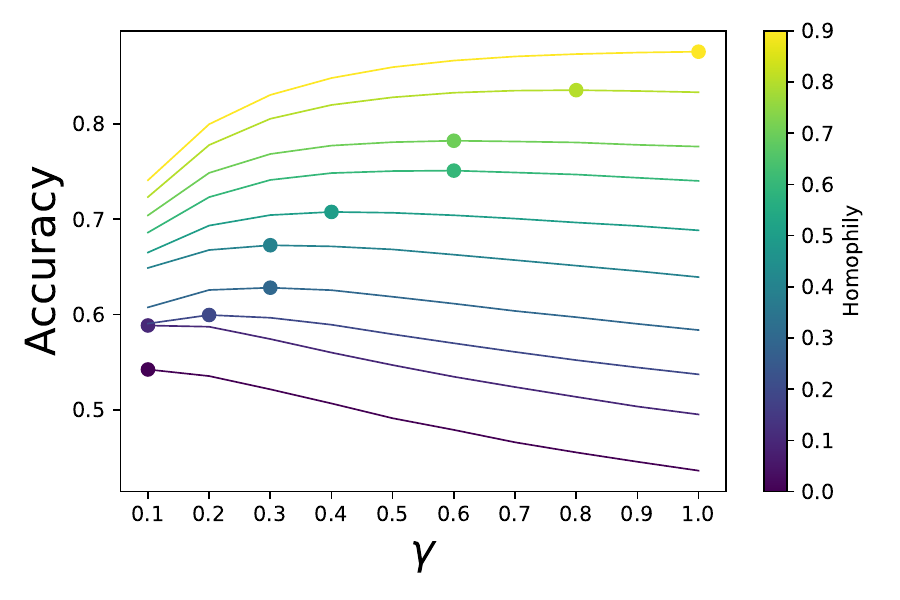} } 
 } }
 \caption{
Experiments on synthetic graphs with varying homophily levels. 
The y-axis represents the average test accuracy.
(a) Performance comparison of PD-GCN and PD-GAT with baseline models across different graph homophily levels.
The x-axis denotes the graph homophily level, where a larger value indicates a more homophilic graph.
For the proposed models, solid lines show the performance with the optimal $\gamma$, while dashed lines show the performance with the worst $\gamma$.
(b) Performance of a one-layer PD-GCN under different values of $\gamma$.
Each line corresponds to synthetic graphs with a specific homophily level.
The line color indicates the homophily coefficient, with blue representing low homophily and yellow representing high homophily.
The dot markers denote the optimal $\gamma$ for each homophily level.
}
\label{fig:syn_res}
\end{figure*}

\subsection{Synthetic Experiments}
\label{sec:synthetic}

In this subsection, 
we evaluate the proposed parameterized diffusion on synthetic graphs with controlled homophily levels.
Following~\cite{Karimi_2018,abu2019mixhop}, 
we generate synthetic graphs characterized by a homophily coefficient 
$\mu \in \{0.0,0.1,\ldots,0.9\}$.
The coefficient $\mu$ represents the probability that a node connects to another node with the same class label.
Thus, a smaller $\mu$ corresponds to a more heterophilic graph.

We now describe the graph generation process.
Given the desired number of nodes $N$, 
the number of classes $C$, 
and the homophily coefficient $\mu$, 
the $N$ nodes are divided into $C$ equal-sized classes.
The graph $\mathcal{G}$ is then generated iteratively, 
starting from an empty graph.
At each step, 
a new node $v_i$ with class label $z_i \in \{1,\ldots,C\}$ is added to the graph.
The class distance $d(z_i, z_j)$ is defined as the shortest distance between the two classes on a circle where classes are indexed from $1$ to $C$.
For example, if $C=6$, $z_i=1$, and $z_j=5$, then the distance between these two classes is $2$.
When connecting $v_i$ to an existing node $v_j$, 
the unnormalized connection probability is defined as
\begin{equation*}
p_{ij} =
\begin{cases}
d_j \mu, & \text{if } z_i=z_j, \\
d_j(1-\mu)w_{d(z_i,z_j)}, & \text{otherwise},
\end{cases}
\end{equation*}
where $d_j$ is the degree of node $v_j$, 
and $w_{d(z_i,z_j)}$ denotes the cost of connecting two nodes from different classes with class distance $d(z_i,z_j)$.
The weight $w_d$ decreases exponentially as the class distance increases and is normalized such that $\sum_d w_d=1$.
The connection probabilities are then normalized over existing candidate nodes:
\begin{equation*}
\bar{p}_{ij} = \frac{p_{ij}}{\sum_{k: v_k\in\mathcal{V}} p_{ik}}.
\end{equation*}
Lastly, node features are sampled from overlapping two-dimensional Gaussian distributions, 
where each class has its own Gaussian distribution.

We generate 5 synthetic graphs for each homophily coefficient $\mu$.
For each generated graph, nodes are randomly partitioned into training, validation, and test sets with a ratio of 60\%/20\%/20\%.
Each model, including MLP, GAT, GCN, PD-GCN and PD-GAT,
is trained under the same hyperparameter setting,
with the learning rate $0.01$, weight decay $0.001$ and dropout rate $0.1$. 
The number of layers is set to be 2 for each model. 
Each model use 64 hidden units and the attention-based model use 8 heads with 8 hidden states per head. 
The rewiring strategy is not applied to the proposed methods. 
For PD-GCN and PD-GAT, 
we fix $\alpha = 1.0$ and only vary $\gamma \in \{0.0, 0.1, \ldots, 0.9 \}$.

As shown in Figure~\ref{fig:syn_res} (a), both PD-GCN and PD-GAT outperform MLP, GCN and GAT across all homophily levels, particularly in heterophilic cases. 
PD-GCN with the optimal $\gamma$ achieves the best overall performance. However, the effectiveness of PD-GCN relies on choosing an appropriate $\gamma$, as its performance drops considerably with the worst $\gamma$, becoming worse than the baseline GNNs on more homophilic graphs. 
In contrast, PD-GAT is more robust to the choice of $\gamma$.
Even with the worst $\gamma$, 
PD-GAT still outperforms the baselines. 
This difference arises because PD-GCN encodes parameterized topological information directly in the aggregation weights, 
whereas PD-GAT incorporates such information into edge features,
which are then used to produce aggregation weights through attention.
As a result, the choice of $\gamma$ has a more direct impact on PD-GCN than on PD-GAT.

Figure~\ref{fig:syn_res} (b) shows the performance of a one-layer PD-GCN and demonstrates the strong correlation between the parameter $\gamma$ and model performance across different graph homophily levels.
We obtain the following observations:
\begin{enumerate}
\item As the graph becomes more heterophilic, 
the optimal $\gamma$, indicated by the dot marker on each curve, decreases, and vice versa.
Empirically, smaller values of $\gamma$ induce a graph geometry that is more suitable for heterophilic graphs.
These results demonstrate that PD-GCN can adapt its diffusion scope by adjusting $\gamma$, providing an interpretable control over the diffusion process.
\item The performance changes smoothly with $\gamma$.
The curves corresponding to larger homophily levels are consistently higher than those corresponding to smaller homophily levels, 
which implies that, for a fixed $\gamma$, PD-GCN generally performs better on more homophilic graphs.
\item Furthermore, the more heterophilic the graph is, the worse the performance becomes under the default setting $\gamma=1$.
This phenomenon highlights that the limitations of the conventional Laplacian are more significant in heterophilic cases, 
and that the proposed parameterized Laplacian effectively addresses this issue through flexible diffusion.
\end{enumerate}

\subsection{Benchmark on Real-world Datasets}
\label{sec:benchmark}
In this subsection, we compare PD-GCN and PD-GAT with 6 baseline models\footnote{Baseline models adopt residual connections at each layer, following the implementation provided by~\cite{platonov2023critical}.}: GAT~\cite{velivckovic2017graph}, 
GAT-sep \footnote{Here, ``sep" means concatenating the ego feature of a node with the aggregated neighborhood information.
This trick is used in~\cite{platonov2023critical}, and we follow the same setting.},
GCN~\cite{kipf2016semi},
SAGE~\cite{hamilton2017inductive},
Graph Transformer (GT)~\cite{shi2020masked} and
GT-sep,
and 11 heterophily-specific SOTA models:
H$_2$GCN~\cite{zhu2020beyond},
CPGNN~\cite{zhu2021graph},
GPR-GNN~\cite{chien2020adaptive},
FSGNN~\cite{maurya2022simplifying},
GloGNN~\cite{li2022finding},
FAGCN~\cite{bo2021beyond},
GBK-GNN~\cite{du2022gbk},
JacobiCov~\cite{wang2022powerful},
BernNet~\cite{he2021bernnet},
LINKX~\cite{lim2021large} and
APPNP~\cite{gasteiger2018predict}.
The comparison further includes GNNs with novel aggregation operators,
including
DGN~\cite{beaini2021directional},
$\omega$GCN~\cite{eliasof2023improving}
and PGSO-GCN~\cite{dasoulas2021learning}.
The experiments are conducted on 7 recently proposed heterophily benchmark datasets~\cite{platonov2023critical},
which address some limitations of previously widely used datasets in~\cite{pei2020geom}:
\textit{roman-empire},
\textit{amazon-ratings},
\textit{minesweeper},
\textit{tolokers},
\textit{questions},
\textit{Chameleon-filtered} and \textit{Squirrel-filtered}.
Additional results on homophilic datasets are provided in Appendix~\ref{appendix:homophily}
\footnote{
The overall statistics of datasets are given in Appendix~\ref{appendix:real-dataset}.}.

\input{content/Paper02/table/table_result_heter}

We evaluate PD-GCN and PD-GAT and their variants under the same experimental settings for the baseline models with fixed splits of training, validation, and test nodes used in~\cite{platonov2023critical}.
Specifically, 
we use a learning rate of $3 \times 10^{-5}$, 
weight decay of $0$, 
dropout rate of $0.2$, 
hidden dimension of $512$,
$8$ attention heads, 
and a number of layers selected from $1$ to $5$.
Following~\cite{platonov2023critical}, 
the proposed models are trained for 1000 steps with the Adam optimizer, and the best checkpoint is selected based on validation performance.
For other GNNs,
we perform a grid search for the learning rate, weight decay and dropout rate, and model-specific hyperparameters, 
with details provided in Appendix \ref{appendix:Hyperparameters}. 

The results are summarized in Table~\ref{table:heterophily_new}.
The GNN baselines are organized in the top block, heterophily-specific GNNs are placed in the second block, 
models with novel aggregation operators are listed in the third block, 
and the proposed methods are shown in the bottom block.
From the results, we make the following observations:
\begin{enumerate}
\item In most cases, the proposed parameterized diffusion-augmented GNNs achieve significant performance improvements over the baseline models.
Specifically, PD-GCN, PD-GAT, and PD-GAT-sep outperform GCN, GAT, and GAT-sep, respectively.
\item The proposed models consistently surpass state-of-the-art GNNs specifically designed for heterophilic graphs,
as well as GNNs utilizing various aggregation paradigms. 
The ``sep'' trick further improves the performance of PD-GAT on \textit{roman-empire}, \textit{minesweeper}, and \textit{tolokers}.
This trick allows the aggregation step to assign negative weights to propagated message, 
enabling node-wise diversification, 
which has been shown to be useful for heterophily data~\cite{luan2021heterophily,luan2022complete}.
\item The superior performance of PD-GAT over GT-based methods suggests that identifying useful sparse message-passing structures is more effective than performing message passing between all pairs of nodes.
\item The topology-guided rewiring mechanism is effective for GAT and GAT-sep on most real-world benchmark datasets.
\end{enumerate}

\section{Conclusion}
In this chapter, 
we address the limitations of GNNs in capturing long-range and global topological information, 
particularly in heterophilic graphs, 
by proposing a novel class of parameterized normalized Laplacian matrices. 
The proposed Laplacian provides greater flexibility in controlling diffusion distance between nodes, 
enabling adaptive diffusion scopes that accommodate different levels of graph homophily.
We further prove an order-preserving relationship between diffusion distance and spectral distance, 
which provides theoretical support for using spectral information to guide graph diffusion.
Based on these results, 
we propose two GNNs with flexible diffusion scopes, PD-GCN and PD-GAT, 
together with a topology-guided rewiring strategy that further enhances long-range information propagation.
The effectiveness of the proposed methods is supported both theoretically and empirically.

One limitation of the current approach is that the parameters in the proposed Laplacian are treated as hyperparameters and selected through validation.
A promising future direction is to train these parameters jointly with GNNs networks.

While this chapter studies how spectral properties of graph Laplacians can improve node representation learning, 
it also motivates Chapter~\ref{ch:Paper03}, 
where we construct a problem-specific graph and its corresponding graph Laplacian, 
and show that spectral graph convolution can be used to approximate optimal solutions to a specific kind of NP-hard optimization problems.
Before moving to graph-structured optimization problems, 
Chapter~\ref{ch:Paper01} first studies learning-based methods for solving a computationally expensive optimization problem arising from integer parameter estimation in a setting that is not graph-structured.

\input{content/Paper02/chapter_appendix}

%% file: content/Paper02/chapter_preliminary.tex
\section{Chapter Preliminaries}
\label{sec:paper02_pre}

\subsection{Spectral-based Nodes Relative Distances}
The relative positions of nodes in a graph can be characterized using eigenvectors of graph Laplacian matrices~\cite{kreuzer2021rethinking}.
Since graph data are non-Euclidean, various measures have been proposed to describe distances between nodes~\cite{nadler2005diffusion, Qiu2007, Belkin2003}.
Among them, diffusion distance is commonly used to characterize how information from node $v_i$ can influence node $v_j$ through random walks along graph edges~\cite{beaini2021directional}.

Starting from an initial node, 
a random walk moves to one of its neighboring nodes at each step.
Intuitively, the diffusion distance between $v_i$ and $v_j$ is proportional to the probability that the random walk starting at node $v_i$ meets the random walk starting at node $v_j$ at step $t$.
Based on $\mathbf{L}_\rw$ defined in~\eqref{eq:rm_laplacian},
the \textit{diffusion distance} can be calculated as~\cite{Coifman2006Diffusion, beaini2021directional}
\begin{equation}
\label{def:diffusion-dist}
d_t(v_i, v_j; \LL_\rw) = \Big( \sum_{k=2}^{N} e^{-2t \lambda_k^{(\rw)}}(\u_k^{(\rw)}(i) - \u_k^{(\rw)}(j))^2 \Big)^\frac{1}{2}.
\end{equation}
Another measure of the distance between $v_i$ and $v_j$ 
is the \textit{spectral distance}, defined as~\cite{Belkin2003}
\begin{equation}
\label{def:spectral-dist}
d_{s}(v_i, v_j; \LL_\rw) = | \u_2^{(\rw)}(i) - \u_2^{(\rw)}(j)|.
\end{equation}
Note that $\u_2^{(\rw)}$ 
is the eigenvector corresponding to the smallest positive eigenvalue of $\LL_{\rw}$, 
and it provides positional information for nodes in the graph~\cite{dwivedi2020generalization, kreuzer2021rethinking, muller2023attending}.
A more general definition of spectral distance can use multiple eigenvectors corresponding to the smallest nonzero eigenvalues.

\subsection{Related Works}
\label{subsec:dgn}
\paragraph{Design of Graph Laplacian and Adjacency Matrices} Efforts have been made to overcome the drawbacks of message passing with the conventional graph Laplacian or adjacency matrices.
This includes works that incorporate novel attention mechanisms into GNNs~\cite{Brody2021How, yun2019graph, hu2020heterogeneous, wang2020direct,luan2022revisiting},
as well as transformer-based GNNs that consider fully connected graphs~\cite{kreuzer2021rethinking, muller2023attending}. 
The parameterized graph shift operator (PGSO)~\cite{dasoulas2021learning} generalizes conventional adjacency matrices, 
while $\omega$GNN~\cite{eliasof2023improving} offers trainable weighting factors for aggregation. 

The Directional Graph Network (DGN)~\cite{beaini2021directional} proposes aggregation matrices, 
namely the directional average matrix $\mathbf{B}_{\text{av}}$ and the directional derivative matrix $\mathbf{B}_{\text{dx}}$.
These matrices assign weights that represent the topological importance based on eigenvectors of some conventional Laplacians. 
Given an eigenvector $\boldsymbol{\phi}$, they are defined as
\begin{equation}
\label{eq:directedagg}
\mathbf{B}_{\mathrm{av}}(\boldsymbol{\phi}) \!=\!  |\nabla \boldsymbol{\phi}|, \; 
\mathbf{B}_{\mathrm{dx}}(\boldsymbol{\phi}) \!=\! \nabla \boldsymbol{\phi} - \text{diag}(\nabla \boldsymbol{\phi} \mathbf{1}), \:
\nabla\phi_{ij} = \phi_j - \phi_i \: \: \text{if} \: \: e_{ij} \in \mathcal{E} \: \: \text{else} \:  \: 0.
\end{equation}

Our proposed flexible diffusion differs from these works by introducing a new class of parameterized Laplacian matrices that provides a principled way to control graph spectral properties and adapt to different levels of homophily.
The proposed method enhances conventional message passing and is independent of the specific local aggregation framework or attention mechanism with which it is combined.
Therefore, it has the potential to improve a broad range of GNN architectures, including graph transformers.

\paragraph{Heterophily and Long-range Topology Information} 
In heterophilic graphs, nodes tend to connect to other nodes with different labels~\cite{luan2022revisiting}.
For heterophilic graph learning, 
recent studies have found that global receptive fields are often more useful than purely local information~\cite{zheng2023node, xing2024less, luan2024heterophilic, cheng2024resurrecting}, 
because information from nodes of the same class is more likely to come from distant nodes. 
However, increasing the number of diffusion or message passing steps in GNNs can cause information loss, 
a phenomenon known as over-squashing~\cite{alon2020bottleneck}.
Thus, the difficulty of capturing long-range neighborhood information without degrading its quality partly explains the unsatisfactory performance of many GNNs on heterophilic graphs.
Meanwhile, transformer-based GNNs
still fall short on existing heterophilic benchmarks~\cite{muller2023attending},
suggesting that naively creating edges between distant nodes is not sufficient to address this challenge.
Various graph rewiring methods have also been proposed~\cite{rong2019dropedge, papp2021dropgnn, abu2019mixhop, arnaiz2022diffwire, alon2020bottleneck}. 

Our work addresses the learning challenge on heterophilic graphs by accelerating diffusion from long-range neighbors, 
and is characterized by both flexible parameterized diffusion and a topology-guided rewiring strategy.

%% file: content/Paper02/table/table_result_heter_ablation.tex
\begin{table*}[ht!]
\resizebox{1\hsize}{!}{
  \begin{tabular}{c|cccccccccc}
  \toprule
    Method&
    $\mathbf{L}^{(\alpha, \gamma)}$&
    $\mathcal{PD}$&
    $\mathcal{R}(\mathcal{G})$&
    roman-empire&
    amazon-ratings&
    minesweeper&
    tolokers&
    questions&
    squirrel filtered&
    chameleon filtered
    \\
    \midrule
    \multirow{1}[0]{*}
    GAT&
    &
    &
    &
    80.87 $\pm$ 0.30&
    49.09 $\pm$ 0.63&
    92.01 $\pm$ 0.68&
    83.70 $\pm$ 0.47&
    77.43 $\pm$ 1.20&
    35.62 $\pm$ 2.06&
    39.21 $\pm$ 3.08
    \\

    \multirow{1}[0]{*}
    &
    &
    &
    $\checkmark$ &
    81.21 $\pm$ 0.71&
    47.27 $\pm$ 0.52&
    92.86 $\pm$ 0.53&
    84.28 $\pm$ 0.68&
    78.79 $\pm$ 1.01&
    40.45 $\pm$ 1.36&
    43.51 $\pm$ 5.06
    \\

        &
    $\checkmark$ &
        &
    $\checkmark$ &
    82.27 $\pm$ 0.60&
    47.69 $\pm$ 0.64&
    92.94 $\pm$ 0.55&
    84.59 $\pm$ 0.53&
    78.97 $\pm$ 1.04&
    41.49 $\pm$ 2.72&
    43.89 $\pm$ 4.67
    \\
    
    \midrule
    \multirow{1}[0]{*}
    &
    &
    $\checkmark$ &
    &
    82.50 $\pm$ 0.60&
    49.29 $\pm$ 0.57&
    92.07 $\pm$ 0.69&
    84.11 $\pm$ 0.41&
    77.82 $\pm$ 0.78&
    36.50 $\pm$ 1.37&
    41.84 $\pm$ 2.76
    \\

    \multirow{1}[0]{*}
    PD-GAT &
       &
    $\checkmark$ &
    $\checkmark$ &
    86.82 $\pm$ 0.60&
    47.81 $\pm$ 0.54&
    93.15 $\pm$ 0.80&
    84.39 $\pm$ 0.49&
    79.40 $\pm$ 0.79&
    41.48 $\pm$ 2.34&
    43.09 $\pm$ 4.13
    \\
    
    &
    $\checkmark$ &
    $\checkmark$ &
      &
    83.46 $\pm$ 0.39&
    \textbf{49.69 $\pm$ 0.51}&
    92.15 $\pm$ 0.71&
    84.11 $\pm$ 0.41&
    78.66 $\pm$ 0.97&
    37.83 $\pm$ 1.54&
    43.37 $\pm$ 3.01
    \\
    &
    $\checkmark$ &
    $\checkmark$ &
    $\checkmark$ &
    \textbf{87.27 $\pm$ 0.64}&
    48.03 $\pm$ 0.58&
    \textbf{93.27 $\pm$ 0.56}&
    \textbf{84.74 $\pm$ 0.59}&
    \textbf{79.55 $\pm$ 0.81}&
    \textbf{42.09 $\pm$ 2.65}&
    \textbf{44.16 $\pm$ 4.20}
    \\
    
    \midrule
    \midrule
    \multirow{1}[0]{*}
    &
    &
    &
    &
    73.69 $\pm$ 0.74& 
    48.70 $\pm$ 0.63&
    89.75 $\pm$ 0.52&
    83.64 $\pm$ 0.67&
    76.09 $\pm$ 1.27&
    39.47 $\pm$ 1.47&
    40.89 $\pm$ 4.12
    
    \\

    \multirow{1}[0]{*}
    GCN&
    &
    &
    $\checkmark$&
    74.50 $\pm$ 0.58&
    48.32 $\pm$ 0.54&
    89.93 $\pm$ 0.60&
    83.41 $\pm$ 0.93&
    77.24 $\pm$ 1.14&
    40.92 $\pm$ 1.49&
    43.60 $\pm$ 1.91
    
    \\

    \multirow{1}[0]{*}
    &
    $\checkmark$&
    &
    $\checkmark$&
    74.60 $\pm$ 0.65&
    48.45 $\pm$ 0.48&
    89.94 $\pm$ 0.61&
    83.46 $\pm$ 0.92&
    77.27 $\pm$ 1.14&
    41.12 $\pm$ 1.29&
    45.34 $\pm$ 4.53
    
    \\

    \midrule

    &
    &
    $\checkmark$&
    &
    73.97 $\pm$ 0.46&
    49.38 $\pm$ 0.80&
    91.60 $\pm$ 0.62&
    81.35 $\pm$ 0.66&
    74.80 $\pm$ 0.76&
    36.19 $\pm$ 2.45&
    41.65 $\pm$ 2.81
    
    \\

    \multirow{1}[0]{*}
    PD-GCN&
    $\checkmark$&
    $\checkmark$&
    &
    \textbf{78.05 $\pm$ 0.49}&
    \textbf{49.49 $\pm$ 0.67}&
    \textbf{91.60 $\pm$ 0.62}&
    \textbf{83.83 $\pm$ 0.86}&
    \textbf{78.04 $\pm$ 0.91}&
    \textbf{43.31 $\pm$ 1.92}&
    \textbf{46.67 $\pm$ 3.56}
    
    \\

  \bottomrule
  \end{tabular}
  }
\caption{Ablation study on real-world heterophilic datasets proposed by \cite{platonov2023critical}.
Here, GCN and GAT are baseline models. 
A checkmark on $\mathbf{L}^{(\alpha, \gamma)}$ indicates the use of the parameterized normalized Laplacian (or the corresponding parameterized normalized adjacency $\mathbf{P}^{(\alpha, \gamma)}$),
with optimally tuned $\alpha$ and $\gamma$.
An unchecked $\mathbf{L}^{(\alpha, \gamma)}$ indicates the use of the default setting $\mathbf{L}^{(1, 1)} = \mathbf{L}_{\rw}$ and $\mathbf{P}^{(1, 1)} = \mathbf{P}_{\rw}$. 
The term $\mathcal{PD}$ stands for ``parameterized diffusion".
For PD-GCN, it refers to weighted aggregation using $\mathbf{P}^{(\alpha,\gamma)}$. 
For PD-GAT, it refers to incorporating edge features defined by $\mathbf{L}^{(\alpha,\gamma)}$ into the graph attention mechanism.
An unchecked $\mathcal{PD}$ means that the vanilla GCN or GAT is used.
Lastly, $\mathcal{R}(\mathcal{G})$ stands for the graph rewiring strategy.
Results for baselines with $\mathcal{R}(\mathcal{G})$ are obtained by applying the baseline models to the rewired graph.
Results for baselines with both $\mathbf{L}^{(\alpha,\gamma)}$ and $\mathcal{R}(\mathcal{G})$ are obtained by applying the rewiring strategy with optimally tuned $\alpha$ and $\gamma$.
Note that the rewiring technique is not applicable to PD-GCN, as PD-GCN assigns parameterized edge weights based on the original graph structure.
The best results are highlighted in \textbf{bold} separately for GCN-based and GAT-based architectures.}
\label{tab:heter_ablation}
\end{table*}

%% file: content/Paper02/table/table_virtual_node.tex
\begin{table*}[!ht]
 \resizebox{1\hsize}{!}
 {
  \begin{tabular}{ccccccccccc}
  \toprule

    &
    roman-empire&
    amazon-ratings&
    minesweeper&
    tolokers&
    questions&
    squirrel-filtered&
    chameleon-filtered
    \\

    \midrule
    \multirow{1}[0]{*}
    GCN&
    73.69 $\pm$ 0.74& 
    48.70 $\pm$ 0.63&
    89.75 $\pm$ 0.52&
    83.64 $\pm$ 0.67&
    76.09 $\pm$ 1.27&
    39.47 $\pm$ 1.47&
    40.89 $\pm$ 4.12
    \\

    \multirow{1}[0]{*}
    GAT&
    80.87 $\pm$ 0.30&
    49.09 $\pm$ 0.63&
    92.01 $\pm$ 0.68&
    83.70 $\pm$ 0.47&
    77.43 $\pm$ 1.20&
    35.62 $\pm$ 2.06&
    39.21 $\pm$ 3.08
    \\

    \multirow{1}[0]{*}
    GAT-sep&
    88.75 $\pm$ 0.41&
    52.70 $\pm$ 0.62&
    93.91 $\pm$ 0.35&
    83.78 $\pm$ 0.43&
    76.79 $\pm$ 0.71&
    35.46 $\pm$ 3.10&
    39.26 $\pm$ 2.50
    \\
    
    \midrule
    \multirow{1}[0]{*}
    GCN-Virtual Nodes&
    74.47 $\pm$ 0.67& 
    48.18 $\pm$ 0.64&
    89.92 $\pm$ 0.59&
    83.43 $\pm$ 0.92&
    77.26 $\pm$ 1.14&
    40.71 $\pm$ 2.67&
    43.94 $\pm$ 4.15
    \\

    \multirow{1}[0]{*}
    GAT-Virtual Nodes&
    79.35 $\pm$ 0.26& 
    46.85 $\pm$ 0.56&
    92.62 $\pm$ 0.77&
    84.17 $\pm$ 0.70&
    78.47 $\pm$ 0.81&
    39.07 $\pm$ 1.81&
    43.47 $\pm$ 3.59
    \\

    \multirow{1}[0]{*}
    GAT-sep-Virtual Nodes&
    88.07 $\pm$ 0.61& 
    48.92 $\pm$ 0.50&
    93.79 $\pm$ 0.46&
    84.11 $\pm$ 0.50&
    78.00 $\pm$ 0.96&
    37.76 $\pm$ 1.57&
    40.96 $\pm$ 3.54
    \\

    \midrule
    \multirow{1}[0]{*}
    PD-GAT ($\mathcal{R}(\mathcal{G})$)&
    87.27 $\pm$ 0.64&
    48.03 $\pm$ 0.58&
    93.27 $\pm$ 0.56&
    84.74 $\pm$ 0.59&
    79.55 $\pm$ 0.81&
    42.09 $\pm$ 2.65&
    44.16 $\pm$ 4.20
    \\

    \multirow{1}[0]{*}
    PD-GAT-sep ($\mathcal{R}(\mathcal{G})$)&
    89.23 $\pm$ 0.56 &
    50.96 $\pm$ 0.43 &
    94.03 $\pm$ 0.45&
    84.83 $\pm$ 0.40&
    78.88 $\pm$ 0.94&
    39.69 $\pm$ 2.28&
    41.15 $\pm$ 4.66
    \\

  \bottomrule
  \end{tabular}
  }
  \caption{
  Ablation study comparing baseline models with virtual-node methods~\cite{ishiguro2019graph,hu2020open} and the proposed parameterized diffusion with topology-guided rewiring on heterophilic datasets proposed by~\cite{platonov2023critical}.
  The ``sep" refers to the trick proposed in~\cite{zhu2020beyond} which concatenates node's and the mean of neighbours' embedding in each aggregation step, instead of adding them together.}
  \label{table:heterophily_virtual_node}
\end{table*}

%% file: content/Paper02/table/table_result_heter.tex
\begin{table*}[t]
 \resizebox{1\hsize}{!}
 {
  \begin{tabular}{lccccccccccc}
  \toprule
  
    &
    roman-empire&
    amazon-ratings&
    minesweeper&
    tolokers&
    questions&
    squirrel-filtered&
    chameleon-filtered&
    \multicolumn{1}{|c}{Rank}
    \\

    \midrule
    \multirow{1}[0]{*}
    GCN&
    73.69 $\pm$ 0.74& 
    48.70 $\pm$ 0.63&
    89.75 $\pm$ 0.52&
    83.64 $\pm$ 0.67&
    76.09 $\pm$ 1.27&
    39.47 $\pm$ 1.47&
    40.89 $\pm$ 4.12&
    \multicolumn{1}{|c}{12.14}
    \\
    
    \multirow{1}[0]{*}
    SAGE&
    85.74 $\pm$ 0.67&
    \red{53.63 $\pm$ 0.39}&
    93.51 $\pm$ 0.57&
    82.43 $\pm$ 0.44&
    76.44 $\pm$ 0.62&
    36.09 $\pm$ 1.99&
    37.77 $\pm$ 4.14&
    \multicolumn{1}{|c}{10.29}
    \\

    \multirow{1}[0]{*}
    GAT&
    80.87 $\pm$ 0.30&
    49.09 $\pm$ 0.63&
    92.01 $\pm$ 0.68&
    83.70 $\pm$ 0.47&
    77.43 $\pm$ 1.20&
    35.62 $\pm$ 2.06&
    39.21 $\pm$ 3.08&
    \multicolumn{1}{|c}{11.57}
    \\

    \multirow{1}[0]{*}
    GAT-sep&
    \blue{88.75 $\pm$ 0.41}&
    \textcolor{violet}{52.70 $\pm$ 0.62}&
    \blue{93.91 $\pm$ 0.35}&
    83.78 $\pm$ 0.43&
    76.79 $\pm$ 0.71&
    35.46 $\pm$ 3.10&
    39.26 $\pm$ 2.50&
    \multicolumn{1}{|c}{8.71}
    \\

    \multirow{1}[0]{*}
    GT&
    86.51 $\pm$ 0.73&
    51.17 $\pm$ 0.66&
    91.85 $\pm$ 0.76&
    83.23 $\pm$ 0.64&
    77.95 $\pm$ 0.68&
    36.30 $\pm$ 1.98&
    38.87 $\pm$ 3.66&
    \multicolumn{1}{|c}{10.43}
    \\
    
    \multirow{1}[0]{*}
    GT-sep&
    87.32 $\pm$ 0.39&
    52.18 $\pm$ 0.80&
    92.29 $\pm$ 0.47&
    82.52 $\pm$ 0.92&
    78.05 $\pm$ 0.93&
    36.66 $\pm$ 1.63&
    40.31 $\pm$ 3.01&
    \multicolumn{1}{|c}{8.43}
    \\

    \midrule
    \multirow{1}[0]{*}
    H$_2$GCN&
    60.11 $\pm$ 0.52&
    36.47 $\pm$ 0.23&
    89.71 $\pm$ 0.31&
    73.35 $\pm$ 1.01&
    63.59 $\pm$ 1.46&
    35.10 $\pm$ 1.15&
    26.75 $\pm$ 3.64&
    \multicolumn{1}{|c}{22.14}
    \\
    
    \multirow{1}[0]{*}
    CPGNN&
    63.96 $\pm$ 0.62&
    39.79 $\pm$ 0.77&
    52.03 $\pm$ 5.46&
    73.36 $\pm$ 1.01&
    65.96 $\pm$ 1.95&
    30.04 $\pm$ 2.03&
    33.00 $\pm$ 3.15&
    \multicolumn{1}{|c}{22.29}
    \\

    \multirow{1}[0]{*}
    GPR-GNN&
    64.85 $\pm$ 0.27&
    44.88 $\pm$ 0.34&
    86.24 $\pm$ 0.61&
    72.94 $\pm$ 0.97&
    55.48 $\pm$ 0.91&
    38.95 $\pm$ 1.99&
    39.93 $\pm$ 3.30&
    \multicolumn{1}{|c}{18.57}
    \\

    \multirow{1}[0]{*}
    FSGNN&
    79.92 $\pm$ 0.56&
    \blue{52.74 $\pm$ 0.83}&
    90.08 $\pm$ 0.70&
    82.76 $\pm$ 0.61&
    \textcolor{violet}{78.86 $\pm$ 0.92}&
    35.92 $\pm$ 1.32&
    40.61 $\pm$ 2.97&
    \multicolumn{1}{|c}{9.86}
    \\

    \multirow{1}[0]{*}
    GloGNN&
    59.63 $\pm$ 0.69&
    36.89 $\pm$ 0.14&
    51.08 $\pm$ 1.23&
    73.39 $\pm$ 1.17&
    65.74 $\pm$ 1.19&
    35.11 $\pm$ 1.24&
    25.90 $\pm$ 3.58&
    \multicolumn{1}{|c}{22.86}
    \\

    \multirow{1}[0]{*}
    FAGCN&
    65.22 $\pm$ 0.56&
    44.12 $\pm$ 0.30&
    88.17 $\pm$ 0.73&
    77.75 $\pm$ 1.05&
    77.24 $\pm$ 1.26&
    41.08 $\pm$ 2.27&
    41.90 $\pm$ 2.72&
    \multicolumn{1}{|c}{13.71}
    \\

    \multirow{1}[0]{*}
    GBK-GNN&
    74.57 $\pm$ 0.47&
    45.98 $\pm$ 0.71&
    90.85 $\pm$ 0.58&
    81.01 $\pm$ 0.67&
    74.47 $\pm$ 0.86&
    35.51 $\pm$ 1.65&
    39.61 $\pm$ 2.60&
    \multicolumn{1}{|c}{15.29}
    \\

    \multirow{1}[0]{*}
    JacobiCov&
    71.14 $\pm$ 0.42&
    43.55 $\pm$ 0.48&
    89.66 $\pm$ 0.40&
    68.66 $\pm$ 0.65&
    73.88 $\pm$ 1.16&
    29.71 $\pm$ 1.66&
    39.00 $\pm$ 4.20&
    \multicolumn{1}{|c}{19.86}
    \\

    \multirow{1}[0]{*}
    BernNet&
    65.56 $\pm$ 1.34&
    44.64 $\pm$ 0.56&
    77.99 $\pm$ 0.95&
    77.00 $\pm$ 0.65&
    70.43 $\pm$ 1.38&
    \textcolor{violet}{41.18 $\pm$ 1.77}&
    40.90 $\pm$ 4.06&
    \multicolumn{1}{|c}{15.71}
    \\

    \multirow{1}[0]{*}
    LINKX&
    56.15 $\pm$ 0.93&
    52.66 $\pm$ 0.64&
    56.78 $\pm$ 2.47&
    81.15 $\pm$ 1.23&
    71.96 $\pm$ 2.07&
    40.10 $\pm$ 2.21&
    42.34 $\pm$ 4.13&
    \multicolumn{1}{|c}{13.14}
    \\

    \multirow{1}[0]{*}
    APPNP&
    65.87 $\pm$ 0.53&
    46.02 $\pm$ 0.73&
    69.62 $\pm$ 2.11&
    76.98 $\pm$ 1.03&
    64.77 $\pm$ 1.32&
    35.12 $\pm$ 1.12&
    37.50 $\pm$ 3.69&
    \multicolumn{1}{|c}{20.00}
    \\
    
    \midrule
    \multirow{1}[0]{*}
    DGN&
    83.12 $\pm$ 0.47&
    47.65 $\pm$ 0.71&
    90.64 $\pm$ 0.61&
    OOM&
    OOM&
    38.56 $\pm$ 1.84&
    41.24 $\pm$ 3.62&
    \multicolumn{1}{|c}{11.20}
    \\

    \multirow{1}[0]{*}
    \text{$\omega$}GCN&
    74.79 $\pm$ 0.46&
    51.79 $\pm$ 0.74&
    90.71 $\pm$ 0.67&
    80.96 $\pm$ 1.04&
    71.10 $\pm$ 1.45&
    35.65 $\pm$ 2.06&
    40.93 $\pm$ 3.48&
    \multicolumn{1}{|c}{13.00}
    \\

    \multirow{1}[0]{*}
    PGSO-GCN&
    74.72 $\pm$ 0.59&
    48.43 $\pm$ 0.41&
    87.84 $\pm$ 1.95&
    78.58 $\pm$ 1.73&
    OOM&
    41.06 $\pm$ 3.05&
    \textcolor{violet}{43.91 $\pm$ 3.04}&
    \multicolumn{1}{|c}{11.83}
    \\
    \midrule


    \multirow{1}[0]{*}
    PD-GCN&
    78.05 $\pm$ 0.49&
    49.49 $\pm$ 0.67&
    91.60 $\pm$ 0.62&
    83.83 $\pm$ 0.86&
    78.04 $\pm$ 0.91&
    \red{43.31 $\pm$ 1.92}&
    \red{46.67 $\pm$ 3.56}&
    \multicolumn{1}{|c}{6.57}
    \\


    \multirow{1}[0]{*}
    PD-GAT&
    83.46 $\pm$ 0.39&
    49.69 $\pm$ 0.51&
    92.15 $\pm$ 0.71&
    \textcolor{violet}{84.11 $\pm$ 0.41}&
    78.66 $\pm$ 0.97&
    37.83 $\pm$ 1.54&
    43.37 $\pm$ 3.01&
    \multicolumn{1}{|c}{6.71}
    \\
    
    \multirow{1}[0]{*}
    PD-GAT ($\mathcal{R}(\mathcal{G})$)&
    87.27 $\pm$ 0.64&
    48.03 $\pm$ 0.58&
    93.27 $\pm$ 0.56&
    \blue{84.74 $\pm$ 0.59}&
    \red{79.55 $\pm$ 0.81}&
    \blue{42.09 $\pm$ 2.65}&
    \blue{44.16 $\pm$ 4.20}&
    \multicolumn{1}{|c}{4.57}
    \\

    \multirow{1}[0]{*}
    PD-GAT-sep&
    \textcolor{violet}{88.46 $\pm$ 0.58}&
    52.57 $\pm$ 0.95&
    \textcolor{violet}{93.81 $\pm$ 0.42}&
    84.04 $\pm$ 0.51&
    77.29 $\pm$ 0.73&
    37.01 $\pm$ 2.97&
    41.40 $\pm$ 4.74&
    \multicolumn{1}{|c}{6.14}
    \\

    \multirow{1}[0]{*}
    PD-GAT-sep ($\mathcal{R}(\mathcal{G})$)&
    \red{89.23 $\pm$ 0.56} &
    50.96 $\pm$ 0.43 &
    \red{94.03 $\pm$ 0.45}&
    \red{84.83 $\pm$ 0.40}&
    \blue{78.88 $\pm$ 0.94}&
    39.69 $\pm$ 2.28&
    41.15 $\pm$ 4.66&
    \multicolumn{1}{|c}{4.29}
    \\
  \bottomrule
  \end{tabular}
  }
  \caption{Experiment results on heterophily datasets proposed by~\cite{platonov2023critical}.
  Values report the mean and standard deviation of the evaluation metrics on the test datasets.
  Here
  \textit{roman-empire}, \textit{amazon-ratings}, \textit{squirrel-filtered} and \textit{chameleon-filtered} are evaluated using accuracy, 
  while \textit{minsweeper}, \textit{tolokers} and \textit{questions} are evaluated using ROC AUC. 
  The ``sep'' trick refers to the method proposed in~\cite{zhu2020beyond},
  which concatenates each node's embedding with the mean embedding of its neighbors at each aggregation step, 
  instead of adding them together. 
  For a fair comparison, 
  we use the same experimental setup as in~\cite{platonov2023critical} for the proposed models. 
  Results for heterophily-specific GNNs, 
  except for BernNet, LINKX and APPNP, are reported from~\cite{platonov2023critical}. 
  Here, OOM means out of memory.
  The top three results are highlighted in \red{red}, \blue{blue}, and \textcolor{violet}{violet}, respectively.
  The last column reports the average rank.}
  \label{table:heterophily_new}
\end{table*}

%% file: content/Paper02/chapter_appendix.tex
\section{Chapter Appendix}
\subsection{Proof of Theorem \ref{thm:parameterized-matrix}}
\label{appendix:proof_parameterized_matrix}
\begin{proof}
 By Definition \ref{def:aug-lap}, we have
\begin{align*}
& \P^{(\alpha, \gamma)} = \I - \mathbf{L}^{(\alpha,\gamma)}
    = \I - \gamma [\gamma \D + (1-\gamma)\I]^{-\alpha} \mathbf{L}[\gamma\D + (1-\gamma)\I]^{\alpha-1} \\
& =[\gamma \D + (1-\gamma)\I]^{-\alpha}
  [\gamma \D + (1-\gamma)\I - \gamma \mathbf{L}]
 [\gamma \D + (1-\gamma)\I]^{\alpha-1} \\
& =[\gamma \D + (1-\gamma)\I]^{-\alpha}
  [\gamma \A + (1-\gamma)\I]
 [\gamma \D + (1-\gamma)\I]^{\alpha-1}
\end{align*}
It is easy to see that all elements in $\P^{(\alpha,\gamma)}$ are non-negative. 
Since $\A\1=\D\1$, we have
$$
\P^{(1,\gamma)} \1 = (\gamma\D + (1-\gamma)\I)^{-1}
        \left(\gamma\A+(1-\gamma)\I\right) \1 = \1,
$$
completing the proof.
\end{proof}

\subsection{Proof of Theorem \ref{thm:eigenplaplaican}}
\label{appendix:proof_eigenplaplaican}
\begin{proof}
For any nonzero $\x\in \Rbb^N$, 
write $\y:=[\gamma \D + (1-\gamma)\I]^{-1/2}\x$.
Then we have
\begin{align*}
& \frac{\x^\top \mathbf{L}^{(1/2, \gamma)}\x}{\x^\top\x} = \frac{\x^\top\gamma[\gamma \D + (1-\gamma)\I]^{-1/2} \mathbf{L}
[\gamma \D + (1-\gamma)\I]^{-1/2}\x}{\x^\top\x} \\
& = \frac{\gamma\y^\top \mathbf{L} \y}{\y^\top [\gamma \D + (1-\gamma)\I] \y} =\frac{\frac{\gamma}{2} \sum_{ij} w_{ij}(y_i-y_j)^2}
{\gamma \sum_{ij} w_{ij}y_i^2 + (1 - \gamma) \sum_{i}y_i^2}
\end{align*}
By the Rayleigh quotient theorem,  
\begin{equation*}  
\lambda_1^{(\gamma)}
=\min_{\y \neq \0}  \frac{ \frac{\gamma}{2}\sum_{ij} w_{ij}(y_i-y_j)^2}
{\gamma\sum_{ij} w_{ij}y_i^2+ (1 - \gamma) \sum_{i}y_i^2} = 0,
\end{equation*}
where the minimum is reached when $\y$ is a multiple of $\1$
and
\begin{equation*}
\lambda_N^{(\gamma)}
= \max_{\y \neq \0} \frac{ \frac{\gamma}{2}\sum_{ij} w_{ij}(y_i-y_j)^2}
{\gamma\sum_{ij} w_{ij}y_i^2+ (1 - \gamma)\sum_{i}y_i^2}
\leq  \max_{\y \neq \0} \frac{ \sum_{ij} w_{ij}  (y_i^2 +y_j^2)}
 {\sum_{ij} w_{ij}y_i^2}  
 \leq 2,
\end{equation*}
leading to Eq.~\eqref{eq:glevrange}.
The proof of showing $\lambda_2^{(\gamma)}\neq 0$ if and only if ${\cal G}$ is connected is similar to~\cite{Belkin2003}, 
thus we omit the details here.

By the Courant-Fischer min-max theorem,  
for $\gamma\neq 0$,
\begin{align*}
    \lambda_i^{(\gamma)}
& = \min_{\{S:\text{dim}(S)=i\}} \max_{\{\x:\0\neq \x\in S\}} 
\frac{\x^\top \mathbf{L}^{(1/2, \gamma)}\x}{\x^\top\x} \\
& = \min_{\{S:\text{dim}(S)=i\}}\max_{\{\y:\0\neq \y\in S\}} 
\frac{\y^\top\mathbf{L}\y}{\y^\top[\D-\I+ (1/\gamma)\I]\y}.
\end{align*}
It is obvious that the Rayleigh quotient
$\frac{\y^\top\mathbf{L}\y}{\y^\top [\D-\I+ (1/\gamma)\I] \y}$
is strictly increasing with respect to $\gamma\in (0,1]$ if $\mathbf{L}\y\neq 0$,
\ie{} $\y$ not a multiple of $\1$. 
Note that $\lambda_1^{(\gamma)}=0$ and it is reached when 
$\mathbf{L}\y=\0$, or equivalently $\y$ is a multiple of $\1$.
Thus, $\lambda_i^{(\gamma)}$ is strictly increasing with respect to $\gamma$
for $i=2:N$.

From the eigendecomposition of the symmetric $\mathbf{L}^{(1/2,\gamma)}$ in~\eqref{eq:gled},
we can find the eigendecomposition of $\mathbf{L}^{(\alpha,\gamma)}$
as follows:
\begin{align*}
&\mathbf{L}^{(\alpha,\gamma)} = [\gamma\D+(1-\gamma)\I]^{1/2-\alpha}\mathbf{L}^{(1/2, \gamma)} [\gamma\D+(1-\gamma)\I]^{\alpha -1/2} \\
& = [\gamma\D+(1-\gamma)\I]^{1/2-\alpha}(\U \bLambda^{(\gamma)} \U^\top)
[\gamma\D+(1-\gamma)\I]^{\alpha -1/2}\\
& = \left([\gamma\D+(1-\gamma)\I]^{1/2-\alpha}\U\right) \bLambda^{(\gamma)}  
\left([\gamma\D+(1-\gamma)\I]^{1/2-\alpha}\U\right)^{-1}
\end{align*}
Thus,  $\lambda_i^{(\gamma)}$ is also 
an eigenvalue of $\mathbf{L}^{(\alpha,\gamma)}$ for $i=1:N$,
and the $i^{th}$ column of $[\gamma\D+(1-\gamma)\I]^{1/2-\alpha}\U$ is
a corresponding eigenvector.
\end{proof}

\subsection{Proof of Theorem \ref{thm:gradient-v2}}
\label{appendix:gradient-v2}

\begin{proof}
The proof is similar to the proof of~\cite{beaini2021directional}.
According to the definition~\eqref{def:diffusion-dist}, 
the diffusion distance at time $t$ between node $v_i$ and $v_j$ is
\begin{equation*}
   d_t(v_i, v_j; \mathbf{L}^{(1,\gamma)}) = \left( \sum_{k=2}^{N} e^{-2t \lambda_k^{(\gamma)}}(\u_k^{(\gamma)}(i) - \u_k^{(\gamma)}(j))^2 \right)^\frac{1}{2},
\end{equation*}
where we denote $\u_k^{(\gamma)}$ as the $k^{th}$ eigenvector corresponding to the eigenvalue $\lambda_k^{(\gamma)}$ of $\mathbf{L}^{(1,\gamma)}$.
The inequality $d_t(v_m, v_j; \mathbf{L}^{(1,\gamma)}) < d_t(v_i, v_j; \mathbf{L}^{(1,\gamma)})$ is then equivalent to 
\begin{equation*}
    \left( \sum_{k=2}^{N} e^{-2t \lambda_k^{(\gamma)}}(\u_k^{(\gamma)}(m) - \u_k^{(\gamma)}(j))^2 \right)^\frac{1}{2}
    < \left( \sum_{k=2}^{N} e^{-2t \lambda_k^{(\gamma)}}(\u_k^{(\gamma)}(i) - \u_k^{(\gamma)}(j))^2
    \right)^\frac{1}{2}.
\end{equation*}
We can take out $\lambda_2^{(\gamma)}$ and $\u_2^{(\gamma)}$ and rearrange the above inequality as:
\begin{align}
    &\sum_{k=3}^{N} e^{-2t \lambda_k^{(\gamma)}}
    \left(
    (\u_k^{(\gamma)}(m) - \u_k^{(\gamma)}(j))^2 - 
    (\u_k^{(\gamma)}(i) - \u_k^{(\gamma)}(j))^2 
    \right) \nonumber \\ 
    &< e^{-2t \lambda_2^{(\gamma)}}\left(
    (\u_2^{(\gamma)}(i) - \u_2^{(\gamma)}(j))^2 - 
    (\u_2^{(\gamma)}(m) - \u_2^{(\gamma)}(j))^2
    \right).\label{eq:pf-ineq}
\end{align}
The left-hand side of Eq.~\eqref{eq:pf-ineq} 
has an upper bound:
\begin{align*}
    &\sum_{k=3}^{N} e^{-2t \lambda_k^{(\gamma)}}
    \left|
    (\u_k^{(\gamma)}(m) - \u_k^{(\gamma)}(j))^2 - 
    (\u_k^{(\gamma)}(i) - \u_k^{(\gamma)}(j))^2 
    \right|\\
    &\leq
    e^{-2t \lambda_3^{(\gamma)}}
    \sum_{k=3}^{N} 
    \left|
    (\u_k^{(\gamma)}(m) - \u_k^{(\gamma)}(j))^2 - 
    (\u_k^{(\gamma)}(i) - \u_k^{(\gamma)}(j))^2  
    \right|.
\end{align*}
Then Eq.~\eqref{eq:pf-ineq} holds if:
\begin{align*}
    &e^{-2t \lambda_3^{(\gamma)}}
    \sum_{k=3}^{N} 
    \left|
    (\u_k^{(\gamma)}(m) - \u_k^{(\gamma)}(j))^2 - 
    (\u_k^{(\gamma)}(i) - \u_k^{(\gamma)}(j))^2  
    \right|\\
    &< e^{-2t \lambda_2^{(\gamma)}}\left(
        (\u_2^{(\gamma)}(i) - \u_2^{(\gamma)}(j))^2 - 
        (\u_2^{(\gamma)}(m) - \u_2^{(\gamma)}(j))^2
    \right),
\end{align*}
which is equivalent to 
\begin{equation}
\label{eq:pf-ineq3}
\log\left(
    \frac
    {(\u_2^{(\gamma)}(i) - \u_2^{(\gamma)}(j))^2 -(\u_2^{(\gamma)}(m) - \u_2^{(\gamma)}(j))^2 
    }
    {\sum_{k=3}^{N} 
    \left|
    (\u_k^{(\gamma)}(m) - \u_k^{(\gamma)}(j))^2 - 
    (\u_k^{(\gamma)}(i) - \u_k^{(\gamma)}(j))^2  
    \right|}
    \right) \times \frac{1}{2(\lambda_2^{(\gamma)} - \lambda_3^{(\gamma)})} < t.
\end{equation}     
Let the constant $C$ be the left-hand side of Eq.~\eqref{eq:pf-ineq3}, 
then if we take $t \geq \left \lfloor C \right \rfloor + 1$, 
we have $d_t(v_m, v_j; \mathbf{L}^{(1,\gamma)}) < d_t(v_i, v_j; \mathbf{L}^{(1,\gamma)})$.    
Note that $C$ exits if
\begin{equation*}
\frac
{(\u_2^{(\gamma)}(i) - \u_2^{(\gamma)}(j))^2 - (\u_2^{(\gamma)}(m) - \u_2^{(\gamma)}(j))^2}
{\sum_{k=3}^{N} 
\left|
(\u_k^{(\gamma)}(m) - \u_k^{(\gamma)}(j))^2 - 
(\u_k^{(\gamma)}(i) - \u_k^{(\gamma)}(j))^2  
\right|} > 0.
\end{equation*}
This condition is satisfied because we assume $d_{s}(v_m, v_j; \mathbf{L}^{(1,\gamma)}) < d_{s}(v_i, v_j; \mathbf{L}^{(1,\gamma)})$ which implies
$|(\u_2^{(\gamma)}(m) - \u_2^{(\gamma)}(j))^2| < 
 |(\u_2^{(\gamma)}(i) - \u_2^{(\gamma)}(j))^2|$.
This assumption is not imposed in Theorem~2.3 from~\cite{beaini2021directional}, 
but it is necessary for the existence of $C$.
Therefore, Theorem~\ref{thm:gradient-v2} extends Theorem~2.3 from~\cite{beaini2021directional} to a more general class of Laplacians and overcomes some shortcomings.

%
\end{proof}

\subsection{Computational Complexity Analysis}
\label{appendix:compute_nontrivial_eigenvector}

The first nontrivial eigenvector of a graph Laplacian can be computed without performing a full eigendecomposition.
Using iterative methods such as the power method, the per-iteration cost is $O(|\calE|)$, 
which is linear in the number of edges and consistent with prior work~\cite{beaini2021directional,kreuzer2021rethinking}.
Here, we explain the procedure using the symmetric normalized Laplacian as an example for simplicity in notation.
The same idea can be applied to $\LL^{(1/2,\gamma)}$, and the eigenvectors of the general $\LL^{(\alpha,\gamma)}$ can then be obtained using the relation in Theorem~\ref{thm:eigenplaplaican}.

Consider the symmetric normalized Laplacian $\LL_{\sym}$ with eigendecomposition defined in~\eqref{eq:eigdsym}.
The eigenvector corresponding to the smallest eigenvalue $0$ is $\mathbf{u}_1^{(\sym)} = \mathbf{D}^{\frac{1}{2}} \1$.
To compute the first nontrivial eigenvector $\mathbf{u}_2^{(\sym)}$ of $\mathbf{L}_{\sym}$,
we define $\bar{\mathbf{L}} = 2 \mathbf{I} - \mathbf{L}_{\sym}$ and denote the eigenvalues of $\bar{\mathbf{L}}$ as $\bar{\lambda}_i = 2 - \lambda^{(\sym)}_i$.
The $\bar{\mathbf{L}}$ and $\mathbf{L}_{\sym}$ shares eigenvectors,
and $\bar{\lambda}_1 > \bar{\lambda}_2 \geq \cdots \geq \bar{\lambda}_N$.
Thus, computing $\mathbf{u}_2^{(\sym)}$ is equivalent to computing the eigenvector associated with $\bar{\lambda}_2$.
Then we define the deflated matrix $\widetilde{\mathbf{L}} =  \bar{\mathbf{L}} - 2 \mathbf{u}_1^{(\sym)} (\mathbf{u}_1^{(\sym)})^\top / \| \mathbf{u}_1^{(\sym)} \|_2^2$.
The largest eigenvalue of $\widetilde{\mathbf{L}}$ is $\bar{\lambda}_2$, and the corresponding eigenvector is $\mathbf{u}_2^{(\sym)}$.
Therefore, $\mathbf{u}_2^{(\sym)}$ can be obtained by applying the power method to $\widetilde{\mathbf{L}}$.

Note that during power iteration, 
we compute the multiplication of $\widetilde{\mathbf{L}}$ with a vector $\mathbf{b}$.
We do not need to form $\widetilde{\mathbf{L}}$ explicitly,
and the complexity is $O(|\calE|)$ under a reasonable stopping criterion.
In each iteration, 
(1) the term $\bar{\mathbf{L}} \mathbf{b}$ takes $O(|\calE|)$, considering the sparsity of the graph;
(2) the multiplication of $2 \mathbf{u}_1^{(\sym)} (\mathbf{u}_1^{(\sym)})^\top / \| \mathbf{u}_1^{(\sym)} \|_2^2$ with $\mathbf{b}$ takes $O(N)$, 
since we first compute $(\mathbf{u}_1^{(\sym)})^\top \mathbf{b}$.
Thus, the overall complexity of computing $\mathbf{u}_2$ is in $O(|\calE|)$ since $N < |\calE|$.

This procedure also applies to $\LL^{(1/2,\gamma)}$, whose eigenvalues lie in $[0,2]$ as shown in Theorem~\ref{thm:eigenplaplaican}.
For the general $\LL^{(\alpha,\gamma)}$, the corresponding eigenvectors can be recovered from those of $\LL^{(1/2,\gamma)}$ using the relation in~\eqref{eq:gled-alpha}.

\subsection{Real-world Datasets}
\label{appendix:real-dataset}
The overall statistics of the real-world datasets are presented in Table \ref{table:chars} and
Table \ref{table:heter_metric} provides their homophily levels calculated using various homophily metrics.
\input{content/Paper02/table/table_dataset_full_list}
\label{table:chars}
\input{content/Paper02/table/table_heter_metric}
\label{table:heter_metric}

\subsection{Hyperparameters}
\label{appendix:Hyperparameters}
Table~\ref{table:optimal_gamma_alpha} lists the optimal $\gamma$ and $\alpha$ used in baselines and models with the proposed methods.
\input{content/Paper02/table/table_alpha_gamma.tex}

For BernNet, LINKX, APPNP and $\omega$GCN in real-world benchmark,
we perform a grid search for learning rate $\in \{0.01, 0.05, 0.1\}$, weight decay $\in \{0, 5e-7, 5e-6, 1e-5, 5e-5, 1e-4, 5e-4, 1e-3, 5e-3, 1e-2 \}$, dropout $\in \{0, 0.1, 0.3, 0.5, 0.7\}$. Model specific parameters are: (1) BernNet: the propagation steps $K = 10$; (2) LINKX: the numbder of layers of $\text{MLP}_{\text{A}}$ and $\text{MLP}_{\text{X}}$ in $\{1, 2\}$;
(3) APPNP: $\alpha \in \{ 0.1, 0.2, 0.5, 0.9 \}$ and up to $10^{th}$ power of the adjacency is used; (4) PGSO: initialization $\in$ \{ ``GCN", ``all zeros", ``SymLaplacian", ``RWLaplacian", ``Adjacency"\}. According to~\cite{dasoulas2021learning}, the weight decay is $5e-4$, the learning rate is 0.005 for the exponential parameters and 0.01 for all other model parameters. We perform a grid search for dropout $\in \{0, 0.1, 0.3, 0.5, 0.7\}$;
(5) DGN: We compute the first two non-trivial eigenvectors of $\mathbf{L}_{\text{rw}}$ for DGN. We perform a grid search for the hyperparameters of DGN according to~\cite{beaini2021directional} for the learning rate in $\{ 10^{-5}, 10^{-4} \}$, the weight decay in $\{ 10^{-6}, 10^{-5}\}$, the dropout rate in $\{0.3, 0.5\}$, the aggregator in \{``mean-dir1-av", ``mean-dir1-dx", ``mean-dir1-av-dir1-dx" \}, the net type in \{``complex", ``simple" \}.

\subsection{Additional Benchmark Results}
\label{appendix:homophily}
\input{content/Paper02/table/table_homo.tex}

This subsection provides additional results on homophilic datasets.
These experiments are included for completeness, 
while the main focus of this work is on heterophilic datasets, where the proposed parameterized diffusion is expected to be most beneficial.
As shown in Table~\ref{table:pd_homo}, the proposed methods achieve performance comparable to the baseline models on homophilic datasets.

For these experiments, we perform a grid search over the learning rate $\in \{0.01, 0.05, 0.1\}$, weight decay $\in \{0, 5e-7, 5e-6, 1e-5, 5e-5, 1e-4, 5e-4, 1e-3, 5e-3, 1e-2 \}$, dropout $\in \{0, 0.1, 0.3, 0.5, 0.7\}$, and 64 hidden states. Note that the implementation of baselines follows~\cite{platonov2023critical}, where residual connections are adopted in each layer.

%% file: content/Paper02/table/table_dataset_full_list.tex
\begin{table}[htbp]
  \centering  
  \begin{tabular}{lccccc}
  \toprule
  &
  \#Nodes&
  \#Edges&
  \#Features&
  \#Classes&
  Metric
  \\

  \midrule






  cora&
  2,708&
  5,278&
  1,433&
  7&
  Accuracy
  \\

  citeseer&
  3,327&
  4,552&
  3,703&
  6&
  Accuracy
  \\

  pubmed&
  19,717&
  44,324&
  500&
  3&
  Accuracy
  \\

  roman-empire&
  22,662&
  32,927&
  300&
  18&
  Accuracy
  \\

  amazon-ratings&
  24,492&
  93,050&
  300&
  5&
  Accuracy
  \\

  minesweeper&
  10,000&
  39,402&
  7&
  2&
  ROC AUC
  \\

  tolokers&
  11,758&
  519,000&
  10&
  2&
  ROC AUC
  \\

  questions&
  48,921&
  153,540&
  301&
  2&
  ROC AUC
  \\

  squirrel-filtered&
  2,223&
  46,998&
  2,089&
  5&
  Accuracy
  \\

  chameleon-filtered&
  890&
  8,854&
  2,325&
  5&
  Accuracy
  \\

  \bottomrule
  \end{tabular}
    \caption{
    Statistics of the real-world benchmark datasets.
    After preprocessing, all graphs are undirected graphs without self-loops or multiple edges.}
\end{table}

%% file: content/Paper02/table/table_heter_metric.tex
\begin{table}[htbp]
  \centering  
  \begin{tabular}{lcccccc}

  \toprule
  &
  $H_{\mathrm{node}}$&
  $H_{\mathrm{edge}}$&
  $H_{\mathrm{class}}$&
  $H_{\mathrm{agg}}^{\mathrm{M}}(\mathcal{G})$&
  $H_{\mathrm{edge}}^{*}$&
  LI
  \\

  \midrule

  









  cora&
  0.83& 
  0.81& 
  0.77& 
  0.99& 
  0.77& 
  0.59 

  \\

  citeseer&
  0.71& 
  0.74& 
  0.63& 
  0.97& 
  0.67& 
  0.45 

  \\

  pubmed&
  0.79& 
  0.80& 
  0.66& 
  0.94& 
  0.69& 
  0.41  

  \\
  roman-empire&
  0.05& 
  0.05& 
  0.02& 
  1.00& 
  -0.05& 
  0.11  
  \\

  amazon-ratings&
  0.38& 
  0.38& 
  0.13& 
  0.60& 
  0.14& 
  0.04  
  \\

  minesweeper&
  0.68& 
  0.68& 
  0.01& 
  0.61& 
  0.01& 
  0.00  
  \\

  tolokers&
  0.63& 
  0.59& 
  0.18& 
  0.00& 
  0.09& 
  0.01  
  \\

  questions&
  0.90&
  0.84&
  0.08&
  0.00&
  0.02&
  0.00
  \\

  squirrel-filtered&
  0.19& 
  0.21& 
  0.04& 
  0.00& 
  0.01& 
  0.00  
  \\

  chameleon-filtered&
  0.24& 
  0.24& 
  0.04& 
  0.25& 
  0.03& 
  0.01  
  \\
  
  \bottomrule
  \end{tabular}

\caption{Heterophily levels of benchmark datasets.
Here, $H_{\mathrm{node}}$ denotes node homophily~\cite{pei2020geom}, 
$H_{\mathrm{edge}}$ denotes edge homophily~\cite{abu2019mixhop,zhu2020beyond}, 
and $H_{\mathrm{class}}$ denotes class homophily~\cite{lim2021new}.
The metric $H_{\mathrm{agg}}^{\mathrm{M}}(\mathcal{G})$ denotes aggregation homophily~\cite{luan2022revisiting}, 
calculated using $\hat{\mathbf{A}}=\mathbf{A}+\mathbf{I}$.
The metric $H_{\mathrm{edge}}^{*}$ denotes adjusted edge homophily~\cite{platonov2023critical}, 
and LI denotes label informativeness~\cite{platonov2023characterizing}.
A smaller value indicates that the graph is more heterophilic.
A comparison of these metrics can be found in~\cite{luan2024heterophilic}.}
\end{table}

%% file: content/Paper02/table/table_alpha_gamma.tex
\begin{table}[!ht]
    \centering
    \resizebox{1\hsize}{!}{
    \begin{tabular}{lcccccccc}
    \toprule
        ~ & ~ & roman-empire & amazon-ratings & minesweeper & tolokers & questions & squirrel-filtered & chameleon-iltered \\ \hline
        GCN ($\mathcal{R}(\mathcal{G})$) & $\gamma$ & 0.1 & 0.3 & 0.1 & 0.9 & 0.1 & 0.1 & 0.1 \\ 
        ~ & $\alpha$ & 0.6 & 0.3 & 0.3 & 0.9 & 0.2 & 0 & 0 \\ \hline
        GAT ($\mathcal{R}(\mathcal{G})$) &$\gamma$ & 0.8 & 0.2 & 0.2 & 0.5 & 0.5 & 0.2 & 0.1 \\ 
        ~ & $\alpha$ & 0.7 & 0 & 0.3 & 0.9 & 1 & 0.9 & 1 \\ \hline
        PD-GCN &$\gamma$ & 1 & 0.9 & 1 & 0.6 & 0.7 & 1 & 0.9 \\ 
        ~ & $\alpha$ & 0 & 0.9 & 1 & 0.4 & 0.8 & 0.1 & 0 \\ \hline
        PD-GAT &$\gamma$ & 0 & 0.9 & 0.4 & 1 & 0.1 & 0.4 & 0.6 \\ 
        ~ & $\alpha$ & 0 & 0.9 & 0.2 & 1 & 0.4 & 0.3 & 0.2 \\ \hline
        PD-GAT ($\mathcal{R}(\mathcal{G})$) &$\gamma$ & 0.5 & 0.4 & 0.3 & 1 & 0.5 & 0.7 & 0.8 \\ 
        ~ & $\alpha$ & 0.9 & 1 & 0 & 0.7 & 1 & 0.4 & 0.5 \\ \hline
        PD-GAT-sep &$\gamma$ & 0.9 & 0.1 & 0.9 & 0.5 & 0.2 & 0.5 & 0.1 \\ 
        ~ & $\alpha$ & 0.4 & 0.9 & 0 & 1 & 0.2 & 0.3 & 0.9 \\ \hline
        PD-GAT-sep ($\mathcal{R}(\mathcal{G})$) &$\gamma$ & 0.6 & 0.6 & 0.3 & 0.5 & 0.3 & 0.5 & 0.2 \\ 
        ~ & $\alpha$ & 0.8 & 0.2 & 0 & 0.9 & 0.4 & 0.9 & 0.9 \\ \bottomrule
    \end{tabular}
    }
    \caption{Optimal $\gamma$ and $\alpha$ for real-world heterophilic datasets.}
    \label{table:optimal_gamma_alpha}
\end{table}

%% file: content/Paper02/table/table_homo.tex
\begin{table}[!ht]
    \centering
    \begin{tabular}{lccc}
    \toprule
        ~ & Cora & Citeseer & Pubmed \\ \midrule
        ResNet & 72.03 $\pm$ 0.24 & 70.77 $\pm$ 1.81 & 88.01 $\pm$ 0.41 \\ 
        GCN & 86.15 $\pm$ 1.24 & 74.58 $\pm$ 0.97 & 89.63 $\pm$ 0.44 \\ 
        SAGE & 85.12 $\pm$ 1.64 & 74.47 $\pm$ 1.93 & 89.69 $\pm$ 0.51 \\ 
        GAT & 86.46 $\pm$ 1.02 & 74.13 $\pm$ 1.85 & 88.87 $\pm$ 0.65 \\ 
        GAT-sep & 84.24 $\pm$ 1.72 & 73.93 $\pm$ 1.93 & 89.14 $\pm$ 0.57 \\ 
        GT & 86.19 $\pm$ 0.99 & 74.23 $\pm$ 1.12 & 89.48 $\pm$ 0.52 \\ 
        GT-sep & 86.3 $\pm$ 1.13 & 74.14 $\pm$ 0.91 & 89.63 $\pm$ 0.52 \\ 
        \midrule
        DGN & 85.16 $\pm$ 1.17 & 72.70 $\pm$ 1.17 & 87.35 $\pm$ 0.53 \\ 
        $\omega$GCN & 86.13 $\pm$ 1.38 & 74.74 $\pm$ 1.39 & 88.65 $\pm$ 0.42 \\ 
        PGSO & 88.66 $\pm$ 0.94 & 76.55 $\pm$ 1.12 & 89.44 $\pm$ 0.53 \\ 
        \midrule
        PD-GCN & 86.91 $\pm$ 1.45 & 75.17 $\pm$ 1.24 & 89.70 $\pm$ 0.45 \\ 
        PD-GAT & 85.40 $\pm$ 1.41 & 74.83 $\pm$ 1.75 & 89.48 $\pm$ 0.45 \\ \bottomrule
    \end{tabular}
    \caption{Experiments on homophily datasets proposed in~\cite{pei2020geom}.}
    \label{table:pd_homo}
\end{table}

%% file: content/Paper01/Paper01.tex
\chapter{\rm\bfseries Learning a Latent Search Space for Maximizing the Success Probabilities of Extended L2-norm Regularized Babai Estimators}
\label{ch:Paper01}

This chapter studies the success probabilities of some $L_2$-norm regularized Babai estimators 
and proposes a machine learning-based approach to optimize these success probabilities, 
as conventional optimization methods are computationally infeasible.
Specifically, the proposed method learns a latent search space in which the success probability can be estimated and optimized more efficiently.

As discussed in Section~\ref{sec:pre_linear_model},
the real-valued MIMO detection problem in~\eqref{eq:ils-transformed}, which is a special case of the BILS problem in~\eqref{eq:linear-model-bils}, is NP-hard.
To improve performance, 
one may consider an $L_2$-norm regularized version of the integer least squares problem defined in~\eqref{eq:ils-transformed-l2},
which is also NP-hard.
The Babai estimator provides a low-complexity suboptimal solution to this $L_2$-norm regularized problem.
The resulting estimators are also referred to as $L_2$-norm regularized Babai estimators.
For a review of the Babai estimator, 
please refer to Section~\ref{sec:babai}.

Section~\ref{sec:ml_lambda_problem} extends the standard Babai estimator to two $L_2$-norm regularized Babai estimators.
For these extended estimators, 
we derive their success probabilities as functions of the regularization parameters.
However, 
conventional optimization methods for selecting the regularization parameters are computationally inefficient.

To address this issue, Section~\ref{sec:ml_lambda} proposes a learning-based method, called ML-$\lambda$, 
to improve the success probabilities of the extended regularized Babai estimators.
ML-$\lambda$ shifts the major computational cost to an offline training phase and computes an approximation to the optimal regularization parameter vector in polynomial time during inference.

The numerical results in Section~\ref{sec:ml_lambda_exp} demonstrate the superior performance of the proposed estimators and the effectiveness of ML-$\lambda$.

\input{content/Paper01/chapter_preliminary}

\section{Learning to Optimize the Success Probability}
\label{sec:ml_lambda}
In this section 
we consider maximizing $f(\blambda;\H,\sigma_\noisevector) = \Pr(\x^{(\blambda)}=\x^*)$
and $\bar{f}(\blambda;\H,\sigma_\noisevector) = \Pr(\bbx^{(\blambda)}=\x^*)$
with respect to $\blambda$.
Because the two functions have similar forms, 
we focus on $f(\blambda;\H,\sigma_\noisevector)$.
The proposed method can be applied to maximize $\bar{f}(\blambda;\H,\sigma_\noisevector)$ as well.

The proposed approach is motivated by amortized optimization through \textit{learning to optimize} introduced in Section~\ref{sec:learning_to_optimize}, 
which trains a neural network to efficiently infer solutions for repeated optimization problems with similar structures~\cite{amos2022tutorial,monga2021algorithm}.
In particular, 
methods that learn a continuous latent space of solutions using variational autoencoders (VAEs)~\cite{kingma2013auto} 
have been successfully applied in various domains~\cite{hottung2020learning, gomez2018automatic, luo2018neural}.
We use a VAE in this chapter to transform the difficult search over $\blambda$ into an optimization problem in a structured continuous latent space.
Since VAE-based methods are problem-dependent, 
we design a specific VAE-based method, 
called ML-$\blambda$, 
for the success-probability optimization problem considered here.
Essentially our learning-based approach transforms the difficult 
optimization problem to a much easier one.
After training, given $\H$, $\sigma_\noisevector$, and the default regularization parameter vector $\blambda^*$, 
ML-$\blambda$ produces a good approximation to the optimal $\hat{\blambda}$ at low inference cost.

The goal of our learning-based method is to learn a meaningful mapping between an $\blambda$ and its latent representation $\z \in \Rbb^h$ for a given problem instance $(\H$, $\sigma_\noisevector)$.
In this problem-specific latent space, 
we also train a predictor that estimates the success probability from $\z$.

In this data-driven approach,
a training dataset is generated to train the neural network parameters.
Denote the generated training set by $T_{\text{train}} = \{ (\hat{\blambda}_i, f(\hat{\blambda}_i; \H^{(i)},\sigma_{\noisevector}^{(i)}), \H^{(i)}, \sigma_{\noisevector}^{(i)}) \}_{i}$.
This offline data-generation phase can be time-consuming because,
for each randomly sampled problem instance 
$(\H^{(i)}, \sigma_{\noisevector}^{(i)})$,
we compute the optimal regularization parameter vector $\hat{\blambda}_i$ by maximizing 
$f(\blambda; \H^{(i)},\sigma_{\noisevector}^{(i)})$ 
using a conventional optimization algorithm.
However, once the model is trained, 
the inference phase has time complexity $\mathcal{O}(hmn + h^2 n)$.

The proposed method adopts the conditional VAE (CVAE) framework~\cite{kingma2014semi, sohn2015learning}.
Given a problem instance $(\H, \sigma_\noisevector)$,
we introduce a latent variable $\z$ with prior distribution $p(\z)$ to model the regularization parameter vector $\blambda$.
Assume that the conditional likelihood
$p_{\btheta}(\blambda \mid \z, \H, \sigma_\noisevector)$ is parameterized by $\btheta$.
For inference, we would like to compute the posterior distribution $p_{\btheta}(\z| \blambda, \H, \sigma_\noisevector)$.
By Bayes' rule,
\begin{equation}
p_{\btheta}(\z \mid \blambda,\H,\sigma_\noisevector)
=
\frac{
p_{\btheta}(\blambda \mid \z,\H,\sigma_\noisevector)p(\z)
}{
p_{\btheta}(\blambda \mid \H,\sigma_\noisevector)
},
\label{eq:cvae_posterior}
\end{equation}
where the marginal likelihood is
\begin{equation}
p_{\btheta}(\blambda \mid \H,\sigma_\noisevector)
=
\int p(\z)p_{\btheta}(\blambda \mid \z,\H,\sigma_\noisevector)\,d\z .
\label{eq:cvae_marginal_likelihood}
\end{equation}
However, the posterior~\eqref{eq:cvae_posterior} is generally intractable because evaluating the marginal likelihood~\eqref{eq:cvae_marginal_likelihood} requires integration over the latent variable.
Therefore, we finds an approximate posterior $q_{\bphi}(\z | \blambda, \H, \sigma_\noisevector)$ parameterized by $\bphi$.

To learn $\btheta$ and $\bphi$, 
we use the evidence lower bound (ELBO) \cite{kingma2013auto} as part of the training loss:
\begin{align}
\operatorname{ELBO}(\blambda,\H,\sigma_\noisevector)_{\btheta,\bphi}
&=
\mathbb{E}_{\z \sim q_{\bphi}(\z \mid \blambda,\H,\sigma_\noisevector)}
\left[
\log p_{\btheta}(\blambda \mid \z,\H,\sigma_\noisevector)
\right] \nonumber \\
&\quad
-
D_{\mathrm{KL}}
\left(
q_{\bphi}(\z \mid \blambda,\H,\sigma_\noisevector)
\| p(\z)
\right),
\label{eq:elbo}
\end{align}
where $D_{\mathrm{KL}}(\cdot\|\cdot)$ denotes the Kullback-Leibler divergence.
It is known that the maximization of \eqref{eq:elbo} is equivalent to the minimization of $D_{\scriptscriptstyle \text{KL}}(q_{\bphi}(\z | \blambda, \H, \sigma_\noisevector) || p_{\btheta}(\z| \blambda, \H, \sigma_\noisevector))$. 

\begin{figure}[t]
\centering
\includegraphics[scale=1]{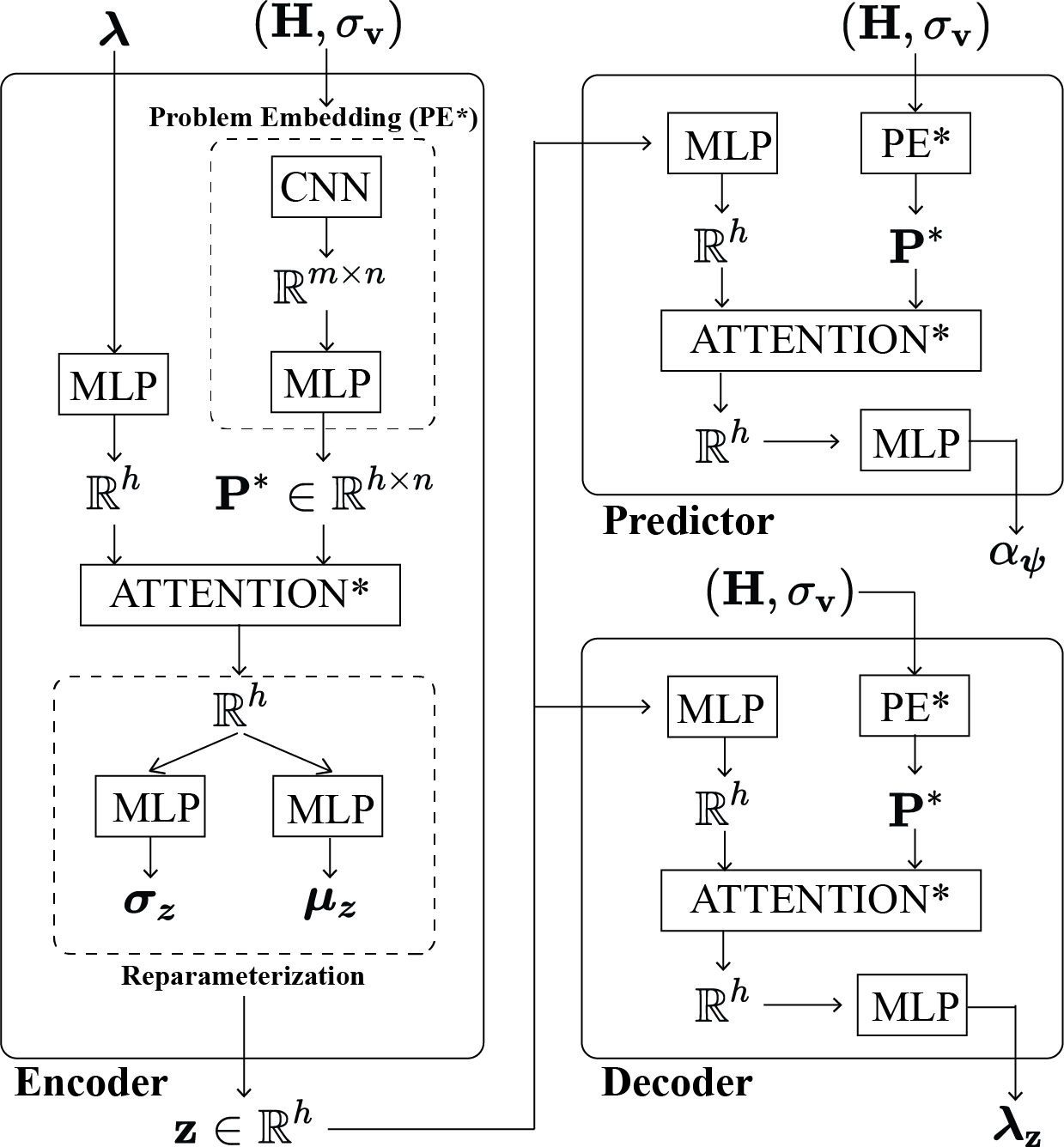}
\caption{The architecture of ML-$\blambda$}
\label{fig:mlmodel}
\end{figure}

The overall architecture of the proposed learning-based method, ML-$\blambda$,
is shown in Fig.\ \ref{fig:mlmodel}.
It consists of three main components,
an encoder, a decoder, and a predictor.

The encoder produces the approximate posterior $q_{\bphi}(\z | \blambda, \H, \sigma_\noisevector)$,
which represents the latent distribution of $\blambda$ for the problem instance $(\H,\sigma_\noisevector)$.
The latent variable is sampled using the reparameterization trick~\cite{kingma2013auto}.
Specifically, if
\begin{equation*}
q_{\bphi}(\z \mid \blambda,\H,\sigma_\noisevector)
=
\mathcal{N}(\bmu_\z,\operatorname{diag}(\boldsymbol{\sigma}_\z^2)),
\end{equation*}
then we sample:
\begin{equation*}
\z = \bmu_\z + \boldsymbol{\sigma}_\z \odot \boldsymbol{\epsilon},
\qquad
\boldsymbol{\epsilon}\sim \mathcal{N}(\0,\I).
\end{equation*}
This reparameterization enables the calculation of the gradient 
$\partial \text{E}_{\z \sim q_{\bphi}(\z | \blambda, \H, \sigma_\noisevector)}{\log p_{\btheta}(\blambda | \z, \H, \sigma_\noisevector)}/\partial \bphi$
in training.
We implement the encoder using convolutional neural networks (CNNs)~\cite{LeCun1998Gradient}, 
multilayer perceptrons (MLPs), and an attention mechanism~\cite{vaswani2017attention}.


The decoder maps a latent representation $\z$, which is the output of the encoder, 
together with the problem instance $(\H,\sigma_\noisevector)$,
back to the $\blambda$-space.  
It is trained to generate an output $\blambda_{\z}$ that closely reconstructs in original $\blambda$.
We define the log-likelihood term as
\begin{equation*}
\log p_{\btheta}(\blambda | \z, \H, \sigma_\noisevector)
= -\| \blambda - \blambda_{\z} \|_2^2,
\end{equation*}
which is used in the ELBO in~\eqref{eq:elbo}.

In addition, 
The predictor estimates the success probability from the latent representation.
Specifically, it outputs
$\alpha_{\bpsi}(\z, \H, \sigma_\noisevector)$ 
which approximates $f(\blambda;\H,\sigma_\noisevector)$.
Since the parameter vector $\bpsi$ is learned during training,
this predictor avoids the need to evaluate the exact success probability during inference.


To ensure that the encoder, decoder, and predictor embed the problem instance consistently,
the problem embedding (PE) component  is shared across the three modules.
The attention parameters in the three modules are also shared to reduce the total number of trainable parameters.

The training objective is to optimize the parameters $\{\bphi,\btheta,\bpsi\}$ by minimizing the average loss over the training set $T_{\mathrm{train}}$.
For the $i^{th}$ training sample, the loss is defined as
\begin{align}
\ell_i(\bphi,\btheta,\bpsi)
&:=
\beta
\left(
\alpha_{\bpsi}(\z^{(i)},\H^{(i)},\sigma_\noisevector^{(i)})
-
f(\hat{\blambda}_i;\H^{(i)},\sigma_\noisevector^{(i)})
\right)^2
\nonumber \\
&\quad
-
(1-\beta)
\operatorname{ELBO}
(\hat{\blambda}_i,\H^{(i)},\sigma_\noisevector^{(i)})_{\btheta,\bphi},
\label{eq:sampleloss}
\end{align}
where $\z^{(i)}$ is the output from the encoder using input 
$(\hat{\blambda}_i,\H^{(i)},\sigma_\noisevector^{(i)})$, 
and $\beta\in(0,1)$ is a hyperparameter that balances the prediction loss and the negative ELBO.


At the inference stage,
the learned parameters $\{\bphi,\btheta,\bpsi\}$ are fixed.
Given a new problem instance $(\H, \sigma_\noisevector)$, 
we first use the trained encoder to obtain the latent representation $\z^*$ corresponding to the default regularization parameter vector $\blambda^*$. 
Then we apply gradient ascent to maximize 
$\alpha_{\bpsi}(\z, \H, \sigma_\noisevector)$ with respect to $\z$,
using $\z^*$ as the initial point.
The gradient $\partial \alpha_{\bpsi}(\z, \H, \sigma_\noisevector) / \partial \z$ 
can be computed efficiently by automatic differentiation with complexity bounded by $\mathcal{O}(h^2n)$~\cite{Griewank2008}.
The number of gradient-ascent iterations is set to a small constant.
Let the resulting latent vector be denoted by $\z_{\mathrm{ML}}$.
The decoder then maps $(\z_{\mathrm{ML}},\H,\sigma_\noisevector)$ to the final regularization parameter vector $\blambda_{\mathrm{ML}}$, 
which is expected to improve upon the default choice $\blambda^*$.


\section{Numerical Results}
\label{sec:ml_lambda_exp}
In our experiments, 
we set $k=2$ in~\eqref{eq:mimo_const} which corresponds to the 16QAM constellation in MIMO detection.
We use the formulas in~\eqref{eq:l2rbsp} and~\eqref{eq:dl2rbsp} 
to compute the theoretical success probabilities of the two estimators 
$\x^{(\blambda)}$ and $\bar{\x}^{(\blambda)}$.
We generate two types of matrices, 
which correspond to the real-valued channel matrices used in communications.

\ben 
\item  $\H \in \Rbb^{m\times n}$ where the entries of $\H$ are sampled independently sampled from $\calN(0,1)$.
Equivalently, in the implementation, $\H$ is generated by the \textsc{matlab}'s built in function $\text{randn}(m, n)$. 
We take $\sigma_\noisevector = |\text{randn}|$. 
If $\Pr(\x^{(\hat{\blambda})}=\x^*)-\Pr(\x^{(\blambda^*)}=\x^*)\geq 0.04$,
then the corresponding instance is taken as a testing sample. 

\item  $\H=\bPsi^{1/2} \A \bPhi^{1/2}$,
where 
$\A=\text{randn}(m,n)$,
$\psi_{ij}= a^{\lvert i-j \rvert}$ and $\phi_{ij}=b^{\lvert i-j \rvert}$ 
with $a,b \in [0,1]$, $\bPsi^{1/2}\bPsi^{1/2}=\bPsi$
and $\bPhi^{1/2}\bPhi^{1/2}=\bPhi$;
see \cite{ChuTKV02, ShiWC08, loyka2002channel}.
In our tests, we choose $a=0.95, b = 0.97$, which makes these matrices significantly
more ill-conditioned than the $\H$ in Type 1.
We take $\sigma_\noisevector = |\text{randn}|*0.08$.
If $\Pr(\x^{(\hat{\blambda})}=\x^*)-\Pr(\x^{(\blambda^*)}=\x^*)\geq 0.02$,
then the corresponding instance is taken as a testing sample. 
\een

Here we explain the selection criterion for testing samples.
When
$\Pr(\x^{(\hat{\blambda})}=\x^*)-\Pr(\x^{(\blambda^*)}=\x^*)$
is small, the default parameter vector $\blambda^*$ is already close to the optimal parameter vector $\hat{\blambda}$.
In this case, 
the remaining room for improvement is very limited, 
making it difficult to observe the the improvement $\blambda_{\text{ML}}$ can bring. 
The imposed success probability gap ensures that the testing samples are informative for comparing 
$\blambda_{\mathrm{ML}}$ with the default parameter vector $\blambda^*$ and the optimized parameter vector $\hat{\blambda}$.

To compute the experimental success probability of an estimator, 
for each instance of $\H$,
we generate 200 $\x^*$, 
whose entries are uniformly and independently sampled from $\calX_k$ in~\eqref{eq:mimo_const}.
For each $\x^*$, 
we randomly generate 200 $\noisevector \sim \calN(\0,\sigma_\noisevector^2 \I)$.
This gives $4\times 10^4$ instances of $\y$ in~\eqref{eq:ils-transformed-l2-v2}. 
The experimental success probability is the ratio of 
the number of occurrences of 
the estimator being equal to $\x^\ast$ 
to $4\times 10^4$. 

\begin{figure}[t]
    \centering
    \captionsetup{justification=centering}
    \subfloat[Type 1 matrices, Babai estimator $\x^{(\blambda)}$.]
    {\includegraphics[width=0.8\linewidth]{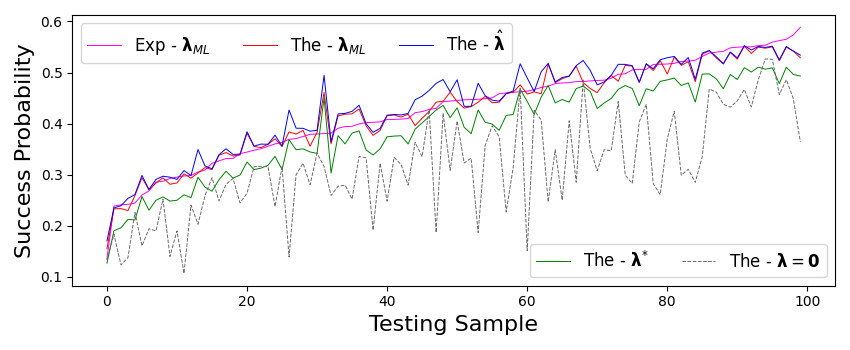}\label{fig:sptest_normal_cor}}\\
    \subfloat[Type 1 matrices, Babai estimator $\bbx^{(\blambda)}$.]{\includegraphics[width=0.8\linewidth]{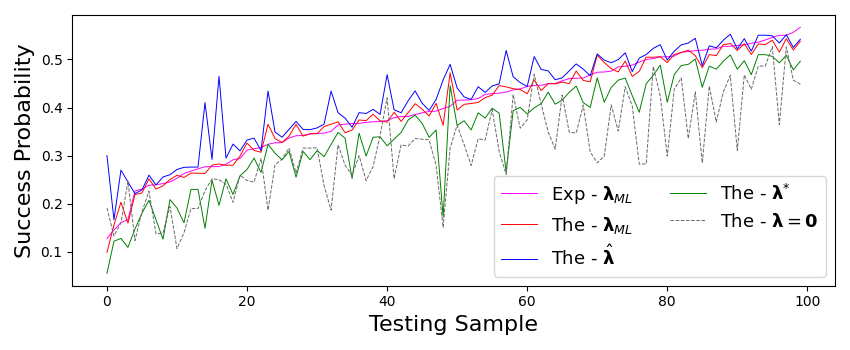}\label{fig:sptest_normal_decor}}
    \caption{Success probabilities of the $L_2$-regularized Babai points for Type 1 matrices.}
    \label{fig:sptest1} 
\end{figure}

\begin{figure}[t]
    \centering
    \captionsetup{justification=centering}
    \subfloat[Type 2 matrices, Babai estimator $\x^{(\blambda)}$.]{\includegraphics[width=0.8\linewidth]{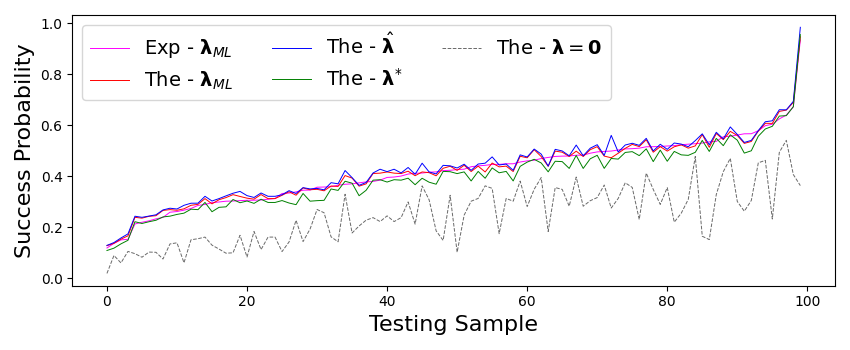}\label{fig:sptest_ill_cor}}\\
    \subfloat[Type 2 matrices, Babai estimator $\bbx^{(\blambda)}$.]{\includegraphics[width=0.8\linewidth]{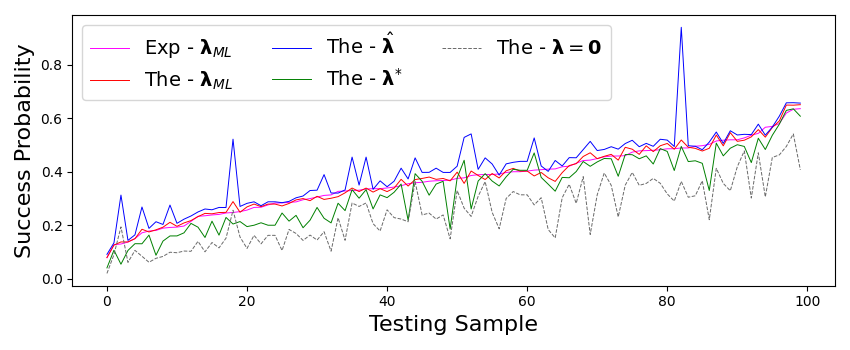}\label{fig:sptest_ill_decor}}
    \caption{Success probabilities of the $L_2$-regularized Babai points for Type 2 matrices.}
    \label{fig:sptest2} 
\end{figure}

Due to the exponential computational complexity of success probability evaluation and conventional optimization,  
we do not perform simulations with large dimensions. 
In our experiments, we set $m=n=8$.
For each type of matrices,
we train two separate networks to maximize $f(\blambda; \A, \sigma_\noisevector)$
and $\bar{f}(\blambda; \A, \sigma_\noisevector)$, 
respectively.
Thus, there are a total of four settings.
For each one, 
we generate a training set with 800 entries,
a validation set of size 100 for selecting the parameters $(\bphi,\btheta,\bpsi)$.
We evaluate each trained network on 100 testing instances.

For all generated data, 
we use \textsc{matlab}'s built-in optimizer \texttt{fminunc} to find the optimal $\hat{\blambda}$.
Since the optimizer may terminate at a local maximum, 
the performance of $\blambda_{\text{ML}}$ 
may possibly exceed that of $\hat\blambda$. 

The data generation process takes roughly 12 hours 
using around 150 CPU cores on the B{é}luga cluster 
of The Digital Research Alliance of Canada.
In the implementation of ML-$\blambda$, 
we set the latent dimension to $h=8$,
use the standard Gaussian prior
$p(\z)=(2\pi)^{-h/2}\exp(\|\z\|_2^2/2)$
and set $\beta = 0.3$ in \eqref{eq:sampleloss},
and use the Adam optimizer with learning rate $0.001$ to minimize the loss function. 

The network is trained for 100 epochs with batch sizes of 20 training samples.
In each training epoch,
we evaluate the network on the validation set using the corresponding success-probability formula.
We select the iterate of $\{ \bphi,\btheta,\bpsi \}$ obtained by Adam that achieves the best validation performance.
The training is completed within 2 hours 
on an NVIDIA GeForce RTX 3080 GPU.

The test results for the success probabilities 
of the two Babai estimators
corresponding to different $\blambda$
are given in Figure~\ref{fig:sptest1} for Type~1 matrices and in Figure~\ref{fig:sptest2} for Type~2 matrices.
In the figure,
the prefixes ``Exp" and ``The" represent ``experimental" and ``theoretical" success probabilities, respectively.
In each plot, 
the 100 testing samples,
each corresponding to one pair $(\H, \sigma_\noisevector)$,
are ordered according to the experimental 
success probability of $\x^{(\blambda_{\text{ML}})}$.
Therefore, 
the same data samples in plots (a) and (b) in Figure~\ref{fig:sptest1} may appear in different orders; the same is true for Figure~\ref{fig:sptest2}.
From Figure~\ref{fig:sptest1} and Figure~\ref{fig:sptest2}, 
we make the following observations:
\begin{enumerate}
\item Although the formula for the success probability $\Pr(\x^{(\blambda)}=\x^*)$ is an approximation,
the experimental and theoretical success probabilities of $\x^{(\blambda_{\text{ML}})}$
are close.
\item $\x^{(\blambda^*)}$ performs (much) better better than $\x^{(\0)}$,
and our extended estimator $\x^{(\hat{\blambda})} $ performs (much) better than $\x^{(\blambda^*)}$,
especially for Type 1 matrices.
These indicate that the conventional $L_2$-norm regularization is helpful
and that the proposed extension can bring significant improvement.
\item In general, the success probability of $\x^{(\blambda_{\text{ML}})}$ 
obtained by our ML-$\lambda$ is 
a reasonable approximation to the success probability of $\x^{(\hat{\blambda})}$
and is higher than the success probability of $\x^{(\blambda^*)}$.
This shows that ML-$\blambda$ can efficiently approximate the optimized regularization parameter vector.
\item There is no significant difference between
the performance of $\x^{(\blambda)}$ and $\bbx^{(\blambda)}$.
Therefore, comments (2) and (3) above can be applied to $\bbx^{(\blambda)}$ as well.
\end{enumerate}

\section{Conclusion}
In this chapter,
we have extended the MMSE-SIC estimator to two $L_2$-regularized Babai estimators by replacing the regularization constant scalar $\lambda^*$ with an $n$-dimensional vector $\blambda$. 
We have also presented two success probability formulae for the two estimators.
Optimizing these success probabilities with respect to $\blambda$ by a conventional optimization method 
would have exponential time complexity.

To overcome this issue, 
we have proposed a machine learning-based method, 
ML-$\blambda$,
which learns a latent search space for efficiently estimating the optimal regularization parameter vector $\hat{\blambda}$ with polynomial-time complexity.
Considering both its performance and computational efficiency, 
ML-$\blambda$ provides a practical approach for improving the success probabilities of $L_2$-norm regularized Babai estimators.

As future work based on this chapter,
we plan to replace the model matrix $\H$ in the training input with the $\R$ factor from its QR factorization.
This modification may allow a learning-based method trained for a fixed value of $n$ to be applied to different values of $m$.

%% file: content/Paper01/chapter_preliminary.tex
\section{Problem Statement}
\label{sec:ml_lambda_problem}
We denote the Babai point for the standard $L_2$-norm regularized ILS problem~\eqref{eq:ils-transformed-l2} by $\x^{(\lambda^*)}$.
Although $\x^{(\lambda^*)}$ has been widely used in MIMO detection, 
to our knowledge, 
there is no systematic studies of its performance.
Part of the goal of this chapter is to study its success probability $\Pr(\x^{(\lambda^*)}=\x^*)$.

To obtain an $L_2$-norm regularized Babai point with better estimation performance than $\x^{(\lambda^*)}$, 
we first extend the regularized estimation formulation.
Specifically,
we consider the following problem: 
\begin{equation} \label{eq:mmse_extended}
    \min_{\x \in \calX_k^{n}} \left \| \y - \H \x \right\|^2_2 + \sum_{i=1}^n \lambda_i^2 x_i^2,
\end{equation}
which replaces the regularization term ${\lambda^*}^2\|\x\|_2^2$
in~\eqref{eq:ils-transformed-l2-reg} (equivalent to~\eqref{eq:ils-transformed-l2}), 
where $\lambda^*=\sigma_{\noisevector}/\sigma_{\x^*}$, 
with the more general coordinate-wise regularization term $\sum_{i=1}^n \lambda_i^2 x_i^2$. 
Here, 
$\lambda_i\geq 0$ for $i=1,\ldots,n$.

Let 
\begin{equation*}
\bLambda = \diag(\blambda), \quad \blambda = [\lambda_1, \ldots, \lambda_n]^\top \nonumber
\end{equation*}
Then~\eqref{eq:mmse_extended} is equivalent to  
\begin{equation} \label{eq:ils-transformed-l2-v2}
    \min_{\x \in \calX_k^{n}} \left \|\bmx \y \\ \0 \emx - \bmx \H \\ \bLambda \emx \x \right\|^2_2.
\end{equation}

We denote the Babai point corresponding to~\eqref{eq:ils-transformed-l2-v2} by $\x^{(\blambda)}$.
Let $\blambda^* = \lambda^*\1_n$.
If $\blambda=\blambda^*$, 
then $\x^{(\blambda)}$ reduces to the standard regularized Babai point $\x^{(\lambda^*)}$ for~\eqref{eq:ils-transformed-l2}.
We therefore call $\blambda^*$ the default value of $\blambda$.
This extension introduces coordinate-wise regularization parameters, 
allowing us to optimize $\blambda$ and potentially obtain a Babai point with higher success probability.

The detailed derivation of an approximate formula for the success probability of $\x^{(\blambda)}$, 
i.e.,
$\Pr(\x^{(\blambda)}=\x^*)$ can be found in our paper~\cite{chang2023success}.
We do not obtain an exact formula because we ignore the correlation between 
the elements of the noise vector in the linear model which is used to define $\x^{(\blambda)}$.
But our numerical experiments indicate the formula is often a good approximation
to the experimental success probability.
To deal with the theoretical shortcoming, 
we further transform the model to a new one so that the resulting noise vector
has uncorrelated entries. 
We then define the corresponding Babai point, denoted by $\bbx^{(\blambda)}$, 
and obtain an exact formula for its success probability, 
$\Pr(\bbx^{(\blambda)}=\x^*)$.
The proof details are provided in~\cite{chang2023success}.
The following presents formulas for both success probabilities, $\Pr(\x^{(\blambda)}=\x^*)$ and  $\Pr(\bbx^{(\blambda)}=\x^*)$.

For \eqref{eq:ils-transformed-l2-v2}, we have the following QR factorization:
\begin{equation*}
\bmx \H \\ \bLambda \emx 
= \bmx \tbQ \\ \hbQ \emx \R, \ \ 
    \tbQ \in \Rmbn, \ \ \hbQ  \in \Rnbn. 
\end{equation*}
Assume $\H$ has full column rank.
Therefore $\tbQ$ has full column rank and we compute its QL factorization $\tbQ = \mathbf{\bar{Q}} \LL$ where $\mathbf{\bar{Q}}^\top \mathbf{\bar{Q}} = \I_n$ and $\LL \in \Rbb^{n \times n}$ is lower triangular with positive diagonal entries.
Denote $\B=\hbQ \LL^{-1}$, $\mathbf{\bar{R}} = \LL^{-\top} \mathbf{R}$,
we obtain
\begin{align}
  \Pr(\x^{(\blambda)}=\x^*) \approx
   f(\blambda;\H,\sigma_\noisevector):=
    \frac{1}{2^{kn}} \sum_{\x \in \calX^n_k} \prod_{i=1}^n \rho_i(\x), \label{eq:l2rbsp} \\
  \Pr(\bbx^{(\blambda)}=\x^*) =
    \bar{f}(\blambda;\H,\sigma_\noisevector):=
        \frac{1}{2^{kn}} \sum_{\x \in \calX^n_k} \prod_{i=1}^n \bar{\rho}_i(\x). \label{eq:dl2rbsp}
\end{align}
where
\begin{align*}
\rho_i(\x) = & \begin{cases} 
 \frac{1}{2} \left ( 1 + \erf \left( \frac{r_{ii} -  (\hbq_i^\top\bLambda\x)\sign(x_i)}{\sqrt{2}\sigma_\noisevector\|\tbq_i\|_2} \right) \right ), \!\! \!\! & \!\! x_i=\pm (2^k-1), \\
\frac{1}{2} \left ( 
 \erf \left( \frac{r_{ii} + \hbq_i^\top \bLambda\x}
  {\sqrt{2}\sigma_\noisevector\|\tbq_i\|_2} \right) 
+ \erf \left( \frac{r_{ii} -  \hbq_i^\top\bLambda\x}
  {\sqrt{2}\sigma_\noisevector\|\tbq_i\|_2} \right) \right ), \!\! \!\! & \mathrm{otherwise}
\end{cases} \\
\end{align*}
and
\begin{align*}
\bar{\rho}_i(\x) = & \begin{cases} 
 \frac{1}{2} \left ( 1 + \erf \left( \frac{\bar{r}_{ii} -  (\b_i^\top\bLambda\x)\sign(x_i)}{\sqrt{2}\sigma_\noisevector} \right) \right ), \!\! \!\! & \!\! x_i=\pm (2^k-1), \\
\frac{1}{2} \left ( 
 \erf \left( \frac{\bar{r}_{ii} + \b_i^\top \bLambda\x}
  {\sqrt{2}\sigma_\noisevector} \right) 
+ \erf \left( \frac{\bar{r}_{ii} -  \b_i^\top\bLambda\x}
  {\sqrt{2}\sigma_\noisevector} \right) \right ), \!\! \!\! & \mathrm{otherwise}
\end{cases}
\end{align*}
with the error function 
$\erf(\zeta)=\frac{2}{\sqrt{\pi}}\int_0^\zeta\exp(-t^2)dt$.

To maximize the success probabilities of the extended $L_2$-norm regularized Babai estimators,
we consider maximizing $f(\blambda; \H, \sigma_\noisevector)$
and  $\bar{f}(\blambda; \H, \sigma_\noisevector)$ with respect to $\blambda$. 
However, the computational cost of evaluating $f(\blambda; \H,\sigma_\noisevector)$ and $\bar{f}(\blambda; \H,\sigma_\noisevector)$ grows exponentially with $n$.
Therefore, conventional optimization methods may be too time-consuming for solving these problems, 
as they require repeated evaluations of the success probability.

In the following sections, 
we propose a learning-based method called ML-$\blambda$, 
which aims to approximate the optimal regularization parameter vector in polynomial time.
Numerical results demonstrate the effectiveness of the proposed method and the performance of the two extended estimators.

%% file: content/Paper03/Paper03.tex
\chapter{\rm\bfseries Efficient and Effective Graph Convolution-Enhanced Expectation
Propagation for Estimation}
\label{ch:Paper03}

Chapter~\ref{ch:Paper02} extends the graph Laplacian to improve the expressiveness of GNNs.
Chapter~\ref{ch:Paper01} proposes a learning-based method to increase the success probabilities of some extended Babai estimators for MIMO detection.
This chapter continues the study of learning-based methods for MIMO detection, 
with a particular focus on graph representation learning-based methods.
The goal is to develop a suboptimal estimator that achieves lower SER across a wide range of SNR levels for massive MIMO detection.

As discussed in Section~\ref{sec:grl_setting},
Chapter~\ref{ch:Paper02} studies transductive node classification,
where the training and testing nodes belong to the same graph.
In contrast, 
this chapter treats each MIMO detection problem instance as a graph and therefore belongs to the inductive setting.
Here the network is trained on graphs constructed from training instances and must generalize to unseen graphs constructed from testing instances.
Specifically, 
we investigate a graph convolution formulation of the linear model for MIMO detection and propose an efficient graph convolution-enhanced EP estimator.

The motivation for the proposed method is twofold.
First,
as discussed in Section~\ref{sec:learning_based_detection},
GNN-aided MP methods originating from GEPNet~\cite{kosasih2022graph}, 
have outperformed both classical and other learning-based estimators.
Based on GEPNet,
several methods have been proposed to improve the efficiency of its MP modules~\cite{kosasih2022graph2, he2023gnn}.
However, 
to the best of our knowledge, 
there is no existing work that aims to improve the efficiency of the GNN module.
In this chapter, 
we focus on addressing the inefficiency issue of the GNN module in GEPNet, 
while maintaining performance and, 
in some cases, further improving it.
The proposed method can be seamlessly integrated with other GNN-aided MP methods by replacing their GNN modules.

Second, the theoretical justification and expressiveness of the GNN frameworks currently used for MIMO detection remain underexplored.
As stated in Section~\ref{sec:learning_based_detection},
existing GNN-based MIMO estimators are largely motivated by the pairwise Markov random field (MRF) factorization of the posterior~\cite{scotti2020graph, kosasih2022graph, kosasih2022graph2, he2023gnn}.
In these methods, spatial GNNs are used to parameterize and learn representations of the variable nodes in the MRF graph.
However, this formulation is mainly inherited from general approaches that combine deep learning with MRFs~\cite{yoon2019inference, scotti2020graph},
rather than being explicitly motivated by any unique property of MIMO detection problem. 
Furthermore, 
many spatial GNNs originate as the low order approximation of the graph convolution~\cite{hariri2026return}.
Graph convolution-based networks have not been fully studied for MIMO detection.

To improve both the efficiency and expressiveness of MIMO-specific GNNs, 
we propose a novel
graph convolution-enhanced EP estimator,
called GCEPNet,
which achieves state-of-the-art performance with fewer trainable parameters and lower inference complexity.


This chapter is organized as follows.
Section~\ref{sec:gecpnet_pre} introduces the preliminaries for this chapter.
Section~\ref{sec:gcepnet} presents the proposed GCEPNet, 
and Section~\ref{sec:gepnet-compare} analyzes its computational efficiency.
Section~\ref{sec:gcepnet_exp} reports the experimental results, 
including training details and performance comparisons with baseline methods.
Finally, Section~\ref{sec:conclusion} concludes this chapter.

\input{content/Paper03/chapter_preliminary}

\section{Efficient Graph Convolution-Based Estimator}

\label{sec:gcepnet}

In this section, we propose graph convolution-enhanced expectation propagation (GCEPNet), 
a novel GNN architecture with theoretical support for efficiently incorporating correlations among unknown transmitted symbols in MIMO detection.
The proposed method is designed to address two key limitations of existing approaches: 
the loss of correlation information in the original EP estimator and the computational bottleneck caused by the GNN module in state-of-the-art GNN-aided estimators.
The proposed method also provides a basis for developing more effective graph convolutional neural networks for MIMO detection in future research.

\begin{figure*}
\centering
\includegraphics[width=1.0\textwidth]{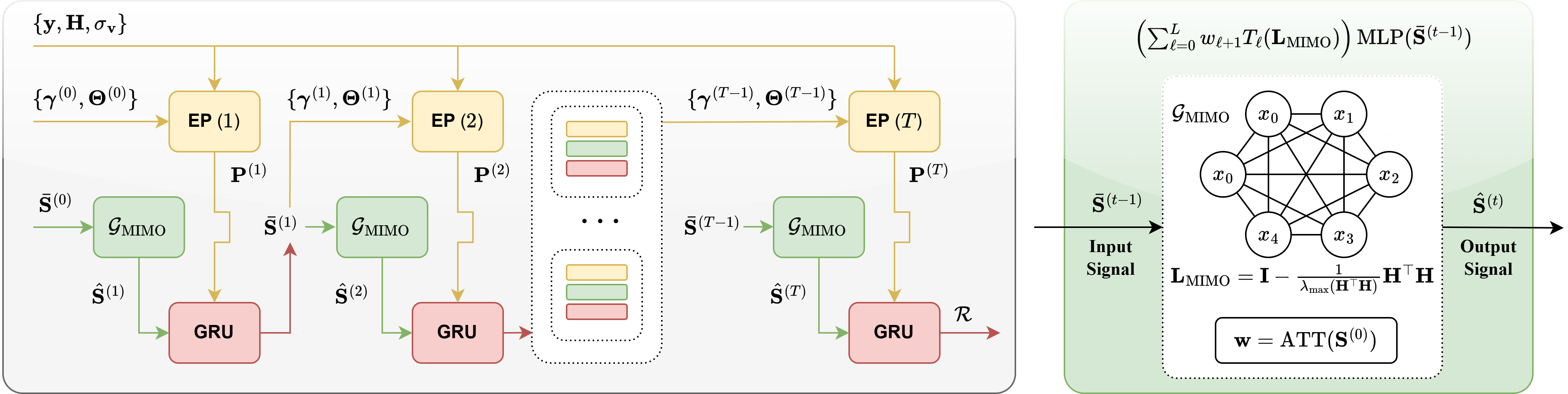}
\caption{The structure of GCEPNet.
The left panel shows the main workflow, where each iteration $t$ consists of EP calculation, graph convolution, and GRU gating modules. The arrows indicate the data flow.
The right panel illustrates the graph convolution process,
that computes \eqref{eq:hidden-feature} and \eqref{eq:gcepnet},
using $N_t = 3$ as an example.}  
\label{fig:GCEPNet}
\end{figure*}

\subsection{Graph Convolution Form of the System Model}
\label{sec:mimo_graph_cov}
In this section, we show that the real-valued MIMO system model can be represented in graph convolution form.

To learn representations of the unknown variables $x_i$ and capture correlations among them, 
we define, 
for each problem instance, 
a fully connected graph with self-loops
${\cal G}_{\mathrm{MIMO}}=(\mathcal{V},\mathcal{E})$.
Each node $i\in\mathcal{V}$ represents one unknown variable $x_i^*$.
In ${\cal G}_{\mathrm{MIMO}}$, 
an edge $e_{ij}$ is defined between every pair of nodes $(i, j)$, 
representing the potential interaction between $x_i^*$ and $x_j^*$.

The instance-specific quantities $\y$, $\H$, and $\sigma_{\noisevector}$ are encoded through a graph operator and graph signals defined on ${\cal G}_{\mathrm{MIMO}}$.
We define the following Laplacian-like operator:
\begin{equation*}
\mathbf{L}_{\mathrm{MIMO}} = \I - \alpha \H^\top \H := \I - \W, 
\qquad \alpha = 1/\lambda_{\text{max}}(\H^\top \H),
\end{equation*}
where $\W=\alpha\H^\top \H$ is the weight matrix of the graph
and $\W(i,j)$ is the weight associated with edge $e_{ij}$.

Like different variants of the graph Laplacian introduced in Section~\ref{sec:pre_grl}, 
$\mathbf{L}_{\text{MIMO}}$ has nonnegative eigenvalues since $\H$ has full column rank,
more specifically, $\lambda(\mathbf{L}_{\text{MIMO}}) \in [0, 1)$. 
While $\mathbf{L}_{\mathrm{MIMO}}$ resembles the graph Laplacian operators introduced in Section~\ref{sec:pre_grl}, 
it differs from standard graph Laplacians in several ways.
In particular, the off-diagonal weights $\W(i,j)$ can be negative for $i\neq j$, 
and the self-loop weights $\W(i,i)$ are not necessarily equal across nodes.
Thus, $\mathbf{L}_{\mathrm{MIMO}}$ should be viewed as a Laplacian-like operator tailored to the MIMO detection problem, 
rather than as a standard graph Laplacian.

Since in the real-valued system model for MIMO detection in equation~\eqref{eq:mimo-sys-model-real},
$\H$ is assumed to have full column rank,
multiplying both sides of equation \eqref{eq:mimo-sys-model-real} by $(\H^\top \H)^{-1} \H^\top$ gives
\begin{equation}
\x^* = (\H^\top \H)^{-1} (\H^\top\y - \H^\top\noisevector)  
  = (\alpha \H^\top \H)^{-1} (\alpha \H^\top \y - \alpha\H^\top\noisevector).
\label{eq:sys-model-real-scale} 
\end{equation}
Since $|\lambda(\mathbf{L}_{\text{MIMO}})|<1$,
$$
(\alpha \H^\top \H)^{-1}
= (\I-\mathbf{L}_{\text{MIMO}})^{-1}
= \sum_{\ell=0}^{\infty}  \mathbf{L}_{\text{MIMO}}^{\ell}.
$$
Then from \eqref{eq:sys-model-real-scale} we obtain
\begin{equation}
\x^*  
= \left( \sum_{\ell=0}^{\infty}  \mathbf{L}_{\text{MIMO}}^{\ell} \right) (\alpha \H^\top \y - \alpha\H^\top\noisevector).
\label{eq:sys-model-graph-convolution}
\end{equation}
This expression shows that the real-valued MIMO system model can be written in the form of a graph spectral convolution.
Specifically, 
$\x^*$ can be regarded as the result of graph convolution between the polynomial filter $\g(\lambda(\mathbf{L}_{\text{MIMO}})) = \sum_{\ell=0}^{\infty} \lambda(\mathbf{L}_{\text{MIMO}})^{\ell}$
and the graph signal $\alpha \H^\top \y + \alpha\H^\top\noisevector$ as shown in equation~\eqref{eq:ChebNetf}.
This graph convolution form motivates the use of graph convolution-based GNNs to learn representations for the unknown variables $x_i^*$ in the MIMO detection problem.

\subsection{Architecture of GCEPNet}

Based on the discussion in the previous sections, 
we propose graph convolution-enhanced expectation propagation (GCEPNet), 
where an efficient graph convolution module is used in each EP iteration to improve the estimation of the first two moments in~\eqref{eq:tilted_moments}.
The improved moment estimates are then used in subsequent EP iterations.

According to the graph convolution form of the system model in~\eqref{eq:sys-model-graph-convolution},
we initialize the input graph signal $\mathbf{S}^{(0)} \in \Rbb^{n\times 2}$ as
\begin{equation}
\mathbf{S}^{(0)} = [\alpha \H^\top \y, \alpha \sigma_{\noisevector} \H^\top\mathbf{1}],
\label{eq:feature-initial}
\end{equation}
where $\alpha\sigma_{\noisevector}\H^\top\mathbf{1}$ is used as an approximation to the unknown noise-related term $\alpha\H^\top\noisevector$ in~\eqref{eq:sys-model-graph-convolution}.
The $i^{th}$ row of $\mathbf{S}^{(0)}$ is the initial graph signal associated with node $i$, which corresponds to the unknown variable $x_i^*$.
The initial node representation is then computed as
\begin{equation}
\bar{\mathbf{S}}^{(0)} =
\mathbf{S}^{(0)}\mathbf{W}_0 + \mathbf{B}_0.
\label{eq:input-feature}
\end{equation}
Equation~\eqref{eq:input-feature} transforms the two-dimensional initial graph signal into a hidden representation.
Specifically, the trainable parameters $\mathbf{W}_0\in\Rbb^{2\times N_u}$ and $\mathbf{B}_0\in\Rbb^{n\times N_u}$ map $\mathbf{S}^{(0)}$ to $\bar{\mathbf{S}}^{(0)}\in\Rbb^{n\times N_u}$.
This transformation also helps mitigate the approximation error caused by replacing the unknown term $\alpha\H^\top\noisevector$ with $\alpha\sigma_{\noisevector}\H^\top\mathbf{1}$ in~\eqref{eq:feature-initial}.

Figure~\ref{fig:GCEPNet} presents the overall structure of GCEPNet.
The key operation in GCEPNet is to use graph convolution to extract useful information about the unknown vector $\x^*$ from the graph signal defined in~\eqref{eq:feature-initial}.
Motivated by~\eqref{eq:sys-model-graph-convolution}, 
we approximate the infinite graph filter by a finite Chebyshev polynomial filter of order $L$:
\begin{equation*}
\sum_{\ell=0}^{\infty} (\mathbf{L}_{\text{MIMO}})^{\ell} \approx \sum_{\ell=0}^{L} w_{\ell+1} T_{\ell}(\mathbf{L}_{\text{MIMO}}),
\end{equation*}
where $T_{\ell}(\cdot)$ denotes the Chebyshev polynomial of order $\ell$.
Next, we describe the forward process of the proposed GNN module, which is applied in each EP iteration.

At the $t^{th}$ EP iteration, 
we collect the cavity mean and variance of all variables into a matrix:
\begin{equation*}
\mathbf{P}^{(t)} =
[\mathbf{a}^{(t)}, \mathbf{b}^{(t)}],
\end{equation*}
where
\begin{equation*}
\mathbf{a}^{(t)} = [a_1^{(t)},\ldots,a_n^{(t)}]^\top,
\qquad
\mathbf{b}^{(t)} = [(b_1^{(t)})^2,\ldots,(b_n^{(t)})^2]^\top.
\end{equation*}
Here, $a_i^{(t)}$ and $(b_i^{(t)})^2$ are the mean and variance of the cavity distribution for $x_i$ defined in~\eqref{eq:cavity_density}.
For each EP iteration $t=1, \ldots, T$, 
the forward update via graph convolution module is defined as
\begin{align}
\widetilde{\mathbf{S}}^{(t)}&=
\operatorname{MLP}_{1}
\left(\bar{\mathbf{S}}^{(t-1)}\right),
\label{eq:hidden-feature}
\\
\hat{\mathbf{S}}^{(t)}&=
\sum_{\ell=0}^{L}
\operatorname{ATT}(\mathbf{S}^{(0)})_{\ell+1}
T_{\ell}(\mathbf{L}_{\mathrm{MIMO}})
\widetilde{\mathbf{S}}^{(t)},
\label{eq:gcepnet}
\\
\bar{\mathbf{S}}^{(t)}&=
\operatorname{GRU}
\left(\hat{\mathbf{S}}^{(t)},\mathbf{P}^{(t)}\right),
\label{eq:gru}
\\
\mathbf{P}_{\mathcal{G}}^{(t)}&=
R\left(\operatorname{MLP}_{2}
\left(\bar{\mathbf{S}}^{(t)}\right)
\right).
\label{eq:read-out}
\end{align}
We use $\operatorname{MLP}$ for a Multi-Layer Perceptron (MLP) module.
The module $\operatorname{MLP}_{1}$ maps the node representation from the previous EP iteration $\bar{\mathbf{S}}^{(t-1)}$ to the input signal $\widetilde{\mathbf{S}}^{(t)}$ for graph convolution.
The graph convolution in~\eqref{eq:gcepnet} then computes the output signal $\hat{\mathbf{S}}^{(t)}$ by applying a Chebyshev polynomial filter on ${\cal G}_{\mathrm{MIMO}}$.
The attention network $\operatorname{ATT}(\mathbf{S}^{(0)})$ produces an $(L+1)$-dimensional vector of data-dependent Chebyshev coefficients, 
where $\operatorname{ATT}(\mathbf{S}^{(0)})_{\ell+1}$ is the coefficient associated with $\ell^{th}$ order $T_{\ell}(\mathbf{L}_{\mathrm{MIMO}})$.

The gated recurrent unit (GRU)~\cite{cho2014learning} in~\eqref{eq:gru} integrates the graph convolution output $\hat{\mathbf{S}}^{(t)}$ with the cavity information stored in $\mathbf{P}^{(t)}$,
and computes the updated node representation.

Finally, the readout module $R$ in~\eqref{eq:read-out} computes the graph-enhanced estimates of the first two moments for each $x_i$. 
These estimates are then used to update the site parameters in the current EP iteration.

The following two subsections provide more details about the attention module and the readout module.

\subsection{Data-dependent Graph Convolution Coefficients}

The standard Chebyshev polynomial graph convolution is primarily designed for transductive learning on a fixed graph~\cite{defferrard2016convolutional}.
In this setting, graph Laplacian remains the same during training and testing.
The graph filter coefficients are learned during training and fixed.
The learned filter is then used to compute node representations for the testing nodes in the same graph during inference.

However, MIMO detection has an inductive structure.
Each problem instance has its own channel matrix $\H$, observation vector $\y$, and noise level $\sigma_{\noisevector}$.
Consequently, the corresponding graph ${\cal G}_{\mathrm{MIMO}}$ and graph operator $\mathbf{L}_{\mathrm{MIMO}}$ are instance-dependent.
Using a single set of fixed Chebyshev coefficients shared across all MIMO instances may therefore prevent the graph convolution module from adapting to instance-specific information.
 
Thus, in~\eqref{eq:gcepnet}, 
we replace the fixed Chebyshev coefficients with data-dependent coefficients generated by an attention module using $\mathbf{S}^{(0)}$. 
Since $\mathbf{S}^{(0)}$ is constructed from the instance-specific information $\{\y,\H,\sigma_{\noisevector}\}$, 
the attention module allows the graph convolution filter to adapt to each problem instance.

The attention module first computes a key matrix from the initial graph signal:
\begin{equation*}
\mathbf{K} = \operatorname{MLP}_{3}(\mathbf{S}^{(0)}),
\qquad \mathbf{K}\in\Rbb^{n\times (L+1)}.
\end{equation*}
The column $\mathbf{K}_{:, \ell}$ can be interpreted as a node-wise embedding associated with the $\ell^{th}$ Chebyshev coefficient.
Then, for each node $i$, an attention weight $\alpha_i$ is computed by comparing the $i^{th}$ row of $\mathbf{K}$ with an all-one query vector $\mathbf{1}\in\Rbb^{L+1}$ representing the default coefficients: 
\begin{equation}
\alpha_i =
\frac{\exp(\mathbf{K}_{i,:}\mathbf{1})}{ \sum_{j=1}^{n}
\exp(\mathbf{K}_{j,:}\mathbf{1})}.
\label{eq:atten-feature}
\end{equation}
The output of the attention module is then computed as a weighted average of the rows of $\mathbf{K}$:
\begin{equation}
\operatorname{ATT}(\mathbf{S}^{(0)}) =
\left( \sum_{i=1}^{n} \alpha_i \mathbf{K}_{i,:} \right)^\top .
\label{eq:atten-output}
\end{equation}
Therefore, $\operatorname{ATT}(\mathbf{S}^{(0)})\in\Rbb^{L+1}$ provides an instance-dependent vector of Chebyshev coefficients for the graph convolution in~\eqref{eq:gcepnet}.

\subsection{Readout Module for Graph-based Estimation}

This subsection describes how the graph convolution module produces the graph-enhanced moment estimates used in EP.
Since the node representations contain graph-enhanced information about the dependencies among the variables $x_i$, 
the readout module maps these representations to a discrete distribution over the constellation set for each $x_i$.
The first two moments of this distribution are then used to update the EP site parameters.

The $t^{th}$ EP iteration,
the readout module takes the hidden representation $\bar{\mathbf{S}}^{(t)}$ and outputs an $n\times M$ logit matrix where $M = |\calX_k|$:
\begin{equation*}
\mathbf{Z}_{\mathcal{G}}^{(t)} = \operatorname{MLP}_{2}\left(\bar{\mathbf{S}}^{(t)}\right),
\qquad
\mathbf{Z}_{\mathcal{G}}^{(t)}\in\Rbb^{n\times M}.
\end{equation*}
Here we denote $s_j$ as the $j^{th}$ smallest element in $\calX_k$.
The entry $\mathbf{Z}_{\mathcal{G}}^{(t)}(i,j)$ is the unnormalized score for assigning $x_i$ to $s_j$.
The readout function $R$ then applies softmax normalization defined in~\eqref{eq:softmax} to each row:
\begin{equation*}
\mathbf{P}_{\mathcal{G}}^{(t)}(i,j)
=\softmax(\mathbf{Z}_{\mathcal{G}}^{(t)}[i, :]^\top)_j
\end{equation*}
where $\mathbf{P}_{\mathcal{G}}^{(t)}(i,j)$ is the predicted probability that $x_i$ equals $s_j$.
Thus, the $i^{th}$ row of $\mathbf{P}_{\mathcal{G}}^{(t)}$ defines a graph-enhanced discrete distribution over the constellation set for $x_i$.
From this distribution, we compute the graph-enhanced soft estimate and variance:
\begin{equation*}
\hat{\mu}_{i,\mathcal{G}}^{(t)} =
\sum_{j=1}^{M}
s_j\,
\mathbf{P}_{\mathcal{G}}^{(t)}(i,j), \qquad
(\hat{\sigma}_{i,\mathcal{G}}^{(t)})^2 =
\sum_{j=1}^{M}
\left(
s_j-\hat{\mu}_{i,\mathcal{G}}^{(t)}
\right)^2
\mathbf{P}_{\mathcal{G}}^{(t)}(i,j).
\end{equation*}
These two moments define a graph-enhanced Gaussian approximation:
\begin{equation*}
q_{\mathcal{G}}^{(t)}(x_i)
=
N\left(
x_i:\hat{\mu}_{i,\mathcal{G}}^{(t)},
(\hat{\sigma}_{i,\mathcal{G}}^{(t)})^2
\right).
\end{equation*}

For $t=1,\ldots,T-1$, 
the graph-enhanced moments 
$\hat{a}_{i,\mathcal{G}}^{(t)}$ and $(\hat{b}_{i,\mathcal{G}}^{(t)})^2$
are used to update the EP site parameters.
Specifically, they replace the moments obtained from the standard EP tilted distribution 
when computing the site update.
For $x_i$, the graph-enhanced updated site parameters are given by
\begin{equation*}
\theta_{i,\mathcal{G}}^{(t)}
=
\frac{1}{(\hat{\sigma}_{i,\mathcal{G}}^{(t)})^2}
-
\frac{1}{(b_i^{(t)})^2},
\qquad
\gamma_{i,\mathcal{G}}^{(t)}
=
\frac{\hat{\mu}_{i,\mathcal{G}}^{(t)}}{(\hat{\sigma}_{i,\mathcal{G}}^{(t)})^2}
-
\frac{a_i^{(t)}}{(b_i^{(t)})^2}.
\end{equation*}
which replaces the standard updation defined in~\eqref{eq:site_update}.

At the final EP iteration $t=T$, 
the output probability matrix from the network,
$\mathbf{P}_{\mathcal{G}}^{(T)}$,
is used to compute the training loss and the final estimation results.
Let $\Z \in \Rbb^{n \times 2^k}$ be the label matrix, where $\Z(i, j) = 1$ if $x^*_i = s_j$ and $0$ otherwise.
The cross-entropy loss for one training sample, corresponding to one problem instance, is computed as
\begin{equation}
\mathcal{L} = - \text{trace}(\Z^\top \log \mathbf{P}_{\mathcal{G}}^{(T)}),
\label{eq:mimo_ce_sample_loss}
\end{equation}
where $\log(\cdot)$ is a component-wise logarithm operation.

At the inference stage, the soft estimate at the final iteration is used to obtain the final estimation result $\hat{\x}_{\mathcal{G}}$:
\begin{equation*}
\hat{x}_{\mathcal{G}} (i) = \text{arg} \: \underset{s \in \calX_k} {\text{min}} |\hat{\mu}_{i,\mathcal{G}}^{(T)} - s|, \quad \text{for} \quad i=1:n.
\end{equation*}

\section{Computational Complexity Analysis}
\label{sec:gepnet-compare}
This section compares the computational complexity of the GNN modules in GCEPNet and GEPNet.
Both methods introduce additional computations beyond the original EP, 
and these additional costs mainly come from their GNN modules.

In GCEPNet, 
the graph convolution module in~\eqref{eq:hidden-feature} and~\eqref{eq:gcepnet},
GCEPNet transform the input graph signal $\bar{\mathbf{S}}^{(t-1)}$ to the output signal $\hat{\mathbf{S}}^{(t)}$.
In contrast, GEPNet uses an edge-wise spatial GNN aggregation.
Specifically, GEPNet defines the edge feature between nodes $i$ and $j$ as
 as $\boldsymbol{\epsilon}_{ij} = [-\h_i^\top \h_j, \sigma_{\noisevector}^2]$,
and updates the node representation by
\begin{equation}
\hat{\s}_i^{(t)} = \sum_{j \neq i} \text{MLP}_{4} \left([\bar{\s}_i^{(t-1)}, \bar{\s}_j^{(t-1)}, \boldsymbol{\epsilon}_{ij}] \right).
\label{eq:gepnet-gnn}
\end{equation}
This aggregation requires evaluating $\operatorname{MLP}_{4}$ for every ordered pair of distinct nodes.

For comparison, we use MLPs with two hidden layers of sizes $N_{h_1}$ and $N_{h_2}$ in both GEPNet and GCEPNet, following the setting in~\cite{kosasih2022graph}.
The input dimension of $\operatorname{MLP}_{4}$ in~\eqref{eq:gepnet-gnn} is $2N_u+2$, and its output dimension is $N_u$.
Since $n=2N_t$, where $N_t$ is the number of transmit antennas, the total cost of computing the updated hidden representations for all variables $x_i$ in GEPNet  $\mathcal{C}_{1} =  n(n-1)[(2N_u + 2)N_{h_1} + N_{h_1} N_{h_2} + N_{h_2} N_u + N_u]$.
Under the setting $N_u=8$, $N_{h_1}=64$, and $N_{h_2}=32$ used in~\cite{kosasih2022graph}, 
the coefficient of the dominant $n^2$ term is $3464$, which is not negligible.

For GCEPNet, the main additional cost comes from the MLP transformation in~\eqref{eq:hidden-feature} and the Chebyshev graph convolution in~\eqref{eq:gcepnet}, 
whose coefficients are generated by the attention module in~\eqref{eq:atten-feature}--\eqref{eq:atten-output}.
The module $\operatorname{MLP}_{1}$ in \eqref{eq:hidden-feature} maps an input of dimension $N_u$ to an output of dimension $N_u$ through two hidden layers.
The $\operatorname{MLP}_{3}$ in the attention module~\eqref{eq:atten-feature} has an input of size $2$ and an output of size $L$.
The total cost for the GNN module in GCEPNet is $\mathcal{C}_{2} = n(N_u N_{h_1} + N_{h_1} N_{h_2} + N_{h_2} N_{u} + N_{h_2} L + L) + n^2N_uL$.
In $\mathcal{C}_{2}$, the last term $Ln^2N_u$ is the dominant cost of the Chebyshev graph convolution.

When $L<433$,
$\mathcal{C}_{2}$ has a smaller coefficient of $n^2$ than $\mathcal{C}_{1}$.
In practice, we use $L=3$,
so the graph convolution module in GCEPNet has a much smaller dominant coefficient than the edge-wise GNN aggregation in GEPNet.

\begin{figure}[t]
\centerline{\includegraphics[width=0.9\textwidth]{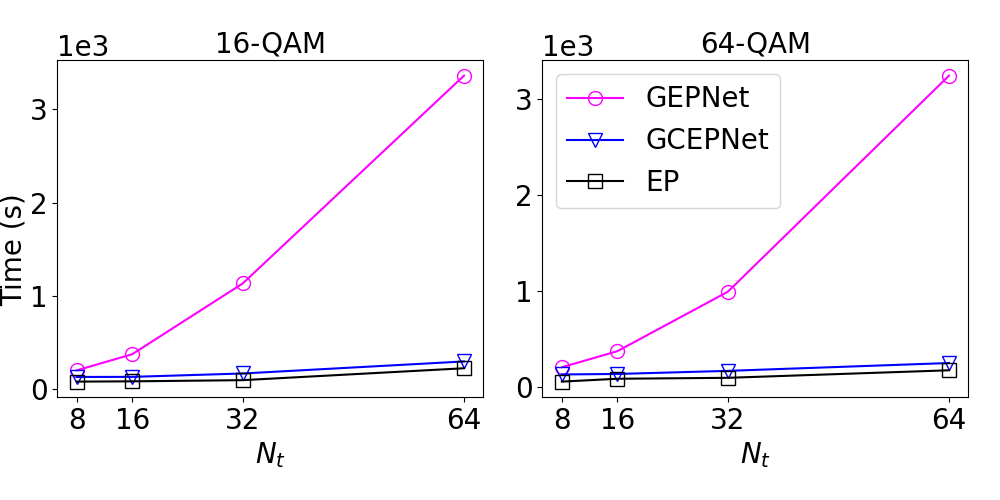}}
\caption{The total inference running time (in seconds) for $5 \times 10^4$ samples with $N_t = N_r$ and the same number of EP iterations ($T=10$). 
Both GCEPNet and GEPNet use 2 GNN layers in each EP iteration.
The implementation of GEPNet is from \cite{kosasih2022graph}.
All methods are under CUDA acceleration.}  
\label{fig:time_comparison}
\end{figure}

Figure~\ref{fig:time_comparison} compares the total inference running time.
Compared with EP, GEPNet\footnote{Following the implementations in \cite{kosasih2022graph}: https://github.com/GNN-based-MIMO-Detection/GNN-based-MIMO-Detection}  scales poorly with the problem size because of its expensive edge-wise GNN aggregation.
GCEPNet alleviates this computational bottleneck by replacing edge-wise message aggregation with the proposed graph convolution module.

\section{Numerical Results} 
\label{sec:gcepnet_exp}

This section compares the performance of GCEPNet with EP and GEPNet.
GEPNet outperforms other methods \cite{kosasih2022graph2},
except IEP-GNN \cite{liu2023model},
which improves the EP module and has the same complexity as GEPNet.
The proposed GNN module can be effortlessly combined with the EP module of IEP-GNN and such a combination may further improve both performance and efficiency.
However, because IEP-GNN is not open-sourced, it is not included in our numerical experiments.

In this test,
we first generate data from the complex-valued MIMO model and then transform it into the real-valued linear model according to~\eqref{def:ils-real}.
Thus, the resulting real-valued channel matrix preserves the block structure induced by the complex-to-real transformation.

All learning-based models are trained on i.i.d. Gaussian channels.
Specifically, during training,
the entries of $\H_c$ are generated independently from $\mathcal{CN}(0, \omega_c^2)$ with $\omega_c^2 = 1/m$.
The corresponding real-valued channel matrix $\H$ is then obtained using~\eqref{def:ils-real}.
The transmitted signal $\x^*$ is uniformly sampled from $\calX_k^{n}$.
Given the signal-to-noise ratio $\text{SNR}_{\text{dB}}$ defined in~\eqref{eq:snr}, 
we compute the corresponding noise standard deviation $\sigma_{\noisevector}$.
We then generate the noise vector $\noisevector$ from $\calN(\0, \sigma_{\noisevector}^2\I)$ and compute the observation $\y$.
The network input is $\{\y,\H,\sigma_{\noisevector}\}$, 
and $\x^*$ is used to compute the training loss and the symbol error rate (SER).

Each network is trained for 850 epochs.
The training loss is the average sample loss, where the loss for each sample is defined as the cross-entropy loss in~\eqref{eq:mimo_ce_sample_loss}.
Each epoch contains 100 iterations, and each iteration uses 100 training samples.
For each training sample, $\text{SNR}_{\text{dB}}$ is randomly sampled from $[25,50]$.
At each epoch,
the network is evaluated on a validation set containing 2000 samples for each $\text{SNR}_{\text{dB}} \in \{25, 26, \cdots, 50 \}$.
We use the Adam optimizer with an initial learning rate of $10^{-3}$.
The learning rate is adjusted every 100 epochs according to the validation loss using the ReduceLROnPlateau scheduler in PyTorch.
The network with the best mean SER on the validation set is selected for testing.
All networks are trained separately for each problem size and QAM configuration.
Testing uses $10^5$ samples for each $\text{SNR}_{\text{dB}}$.
For a fair comparison, each method uses 10 EP iterations with the updating parameter $\beta = 0.7$ in~\eqref{eq:beta_damping} at each iteration, 
while GCEPNet and GEPNet are configured with the same hidden size, 
as detailed in Section~\ref{sec:gepnet-compare}.

We evaluate the trained models under two testing settings.
First, in Section~\ref{sec:iid_channel},
the channels matrix of each testing problem instance is generated from the same i.i.d. Gaussian channel distribution used during training.
Second, in Section~\ref{sec:correlated_channel}, 
we evaluate the same trained models on Kronecker correlated channels.
This second setting tests whether models trained on i.i.d. Gaussian channels can generalize to channel distributions with spatial correlation.

\subsection{Performance Comparison on i.i.d. Gaussian Channels}
\label{sec:iid_channel}
\begin{figure}[t]
\centerline{\includegraphics[width=1.0\textwidth]{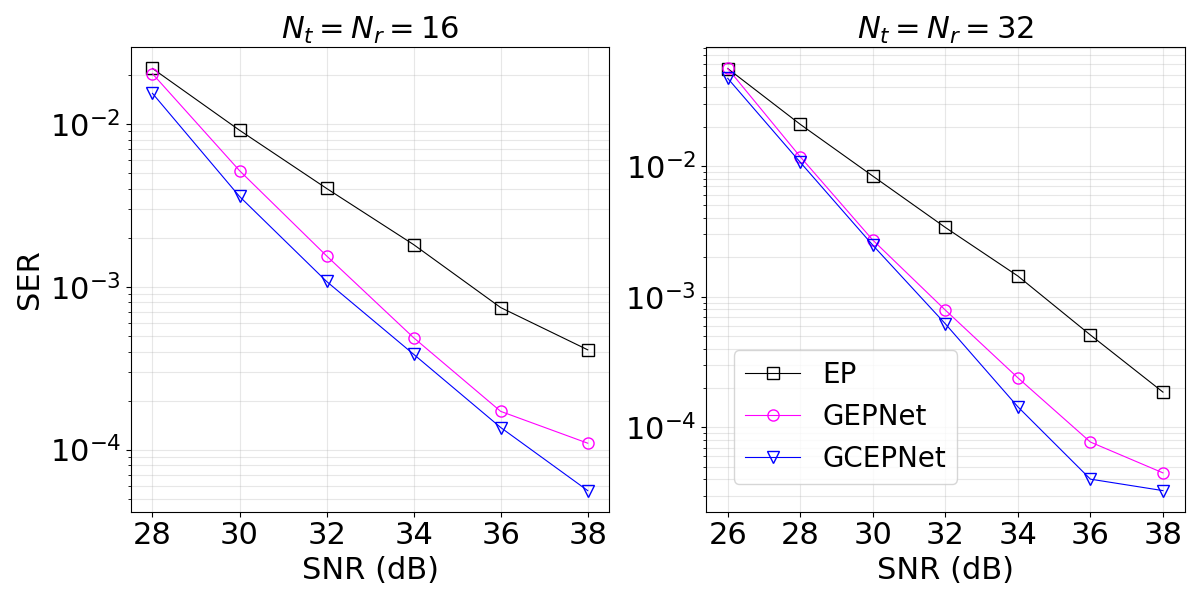}}
\caption{SER performance comparison for $64$-QAM on i.i.d. Gaussian channels.} 
\label{fig:results}
\end{figure}

Figure~\ref{fig:results} compares the average SER of GCEPNet, GEPNet~\cite{kosasih2022graph} and EP~\cite{cespedes2014expectation} for 
$N_t = N_r = 16$ and $N_t = N_r = 32$ with $64$-QAM,
where the testing channel matrices are generated from the same i.i.d. Gaussian distribution as the training channel matrices.

Note that although entries of the channel matrix are independent,
the posterior components $x^*_i$ are correlated after conditioning on $\y$.
Both GCEPNet and GEPNet improve the performance of EP by using their GNN modules to incorporate such correlation information. 
While GCEPNet is significantly more efficient than GEPNet as shown in Section~\ref{sec:gepnet-compare},
it consistently outperforms GEPNet as well.
Given that GEPNet achieves the best performance among existing sub-optimal estimators \cite{kosasih2022graph2},
these results show that GCEPNet provides a more accurate and efficient GNN-aided EP estimator.

\subsection{Robustness to Correlated Channel}
\label{sec:correlated_channel}

\begin{figure}[t]
\centerline{\includegraphics[width=1.0\textwidth]{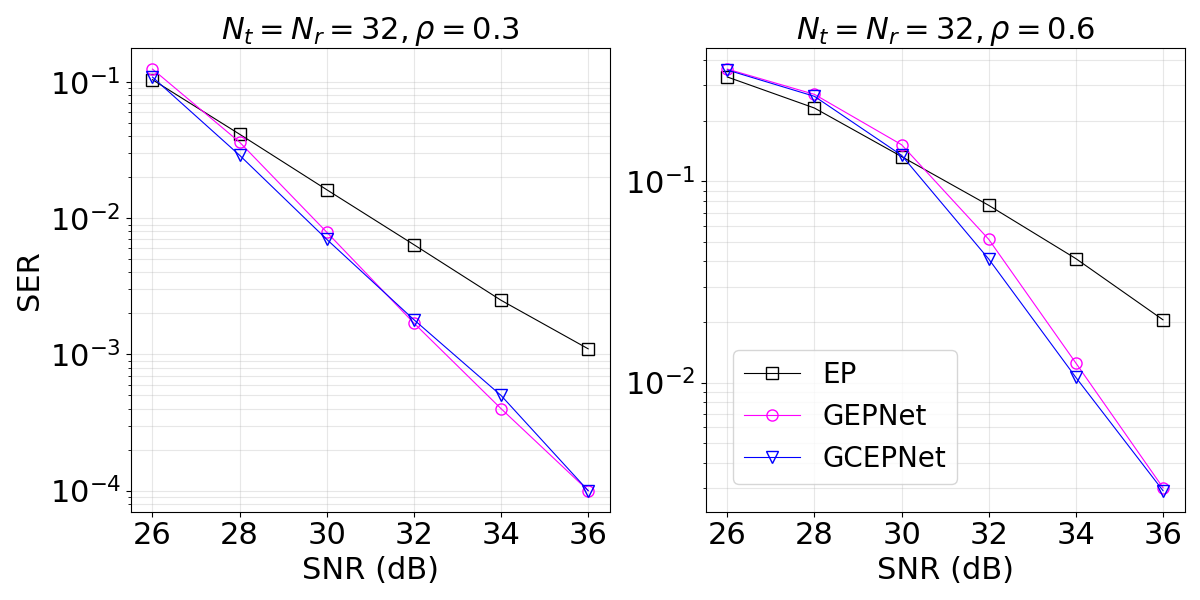}}
\caption{SER performance comparison for $64$-QAM on Kronecker correlated channels.} 
\label{fig:results_corr}
\end{figure}

We further evaluate the robustness of the trained models on correlated channels.
Importantly, the models used in this experiment are still trained only on i.i.d. Gaussian channels.
They are directly tested on correlated channels without retraining.

We use exponential receiver and transmitter correlation matrices, 
denoted by 
$\mathbf{R}_r\in\Rbb^{N_r\times N_r}$ and 
$\mathbf{R}_t\in\Rbb^{N_t\times N_t}$, respectively, 
to generate Kronecker correlated channels~\cite{ChuTKV02, ShiWC08, loyka2002channel}.
For a given correlation coefficient $\rho$, 
entries of these matrices are
$\mathbf{R}_r(i, j)= \rho^{\lvert i-j \rvert}$ and $\mathbf{R}_t(i, j)= \rho^{\lvert i-j \rvert}$.
We generate $\H_c$ with entries independently sampled from $\mathcal{CN}(0,\omega_c^2)$, where $\omega_c^2=1/m$.
The correlated channel is then constructed as
\begin{equation*}
\H_c^{\mathrm{corr}} =
\mathbf{R}_r^{1/2} \H_c \mathbf{R}_t^{1/2},
\end{equation*}
with $\mathbf{R}_r^{1/2}\mathbf{R}_r^{1/2}=\mathbf{R}_r$
and $\mathbf{R}_t^{1/2}\mathbf{R}_t^{1/2}=\mathbf{R}_t$.
The coefficient $\rho$ controls the correlation level.
A larger $\rho$ corresponds to stronger channel correlation and a larger deviation from the training distribution.
In our experiments, we evaluate $\rho=0.3$ and $\rho=0.6$.
After generating $\H_c^{\mathrm{corr}}$, 
we convert it to the corresponding real-valued channel matrix using~\eqref{def:ils-real}.
The procedures for generating $\x^*$ and $\y$ remain the same.

Figure~\ref{fig:results_corr} shows that both GCEPNet and GEPNet generalize well to problem instances with Kronecker correlated channels, even though they are trained only on instances with i.i.d. Gaussian channels.
For both correlation levels, $\rho=0.3$ and $\rho=0.6$, the two GNN-aided EP methods still outperform EP in the moderate-to-high SNR regime.
This robustness is important in practical MIMO systems, where the exact channel correlation structure or correlation coefficient is unknown.
Training a separate estimator for every possible channel distribution is impractical.
Therefore, an estimator trained on standard i.i.d. Gaussian channels while able to generalize to correlated channels is more desirable for deployment.
GCEPNet maintains this generalization ability while replacing the GNN module in GEPNet with a more efficient graph convolution-based module that has fewer parameters.


\section{Conclusion}
\label{sec:conclusion}
In this chapter,
we proposed GCEPNet,
a graph convolution-enhanced expectation propagation  for MIMO detection.
The proposed method incorporates an efficient graph convolution module into EP to improve the estimation of moment information used in site-parameter updates.
By replacing the less efficient GNN aggregation module in GEPNet with a graph convolution-based module, 
GCEPNet reduces the computational bottleneck of GNN-aided EP estimation while preserving the ability to exploit correlation information among transmitted symbols.

The numerical results show that GCEPNet consistently improves upon EP and outperforms GEPNet on i.i.d. Gaussian channels.
Moreover, although the learning-based methods are trained only on i.i.d. Gaussian channels, 
GCEPNet generalizes well to Kronecker correlated channels without retraining.

One limitation of the proposed method is that the graph convolution form of the system model is derived under the assumption that the channel matrix has full column rank.
When the linear system is underdetermined or the channel matrix is rank-deficient, 
this derivation no longer applies.
Another limitation is that GNN-aided EP estimators, including both GCEPNet and GEPNet, 
are typically trained end-to-end with a fixed number of stacked iterations.
This design makes the estimator less flexible when changing the number of iterations at inference time.
Such flexibility would be useful in high-SNR scenarios, where fewer iterations may be sufficient and could reduce computational cost.



%% file: content/Paper03/chapter_preliminary.tex
\section{Chapter Preliminaries}
\label{sec:gecpnet_pre}

\subsection{Expectation Propagation}
\label{sec:ep}
In this section, 
we introduce the expectation propagation (EP) estimator~\cite{cespedes2014expectation}.

To simplify notation, we denote $p(\x^*=\x\,|\,\y)$, $p(\y\,|\,\x^*=\x)$ and $p(\x^*=\x)$ 
by $p(\x\,|\,\y)$, $p(\y\,|\,\x)$ and $p(\x)$,
respectively.
For the real-valued MIMO system model defined in \eqref{eq:mimo-sys-model-real},
consider the posterior probability of $\x^*$ 
given by Bayes' rule: 
\begin{equation}
\hspace*{-2mm}
p(\x \,|\, \y ) \! \propto \! p(\y \,|\, \x ) p(\x) \!=\! N(\y: \H \x, \sigma_{\noisevector}^2 \I) \prod_{i=1}^{n} p(x_i),
\label{eq:sys-model-bayes}
\end{equation}
where the prior $p(x_i) \propto \mathbbm{1}\{x_i \in \calX_k\}$ denotes the uniform prior over the constellation set.
The EP estimator aims to approximate the intractable posterior $p(\x\,|\,\y)$ by a continuous Gaussian density function:
\begin{equation*}
q(\x\,|\,\y)=N(\x:\bmu_{\x},\bSigma_{\x}).
\end{equation*}
The final estimate $\hat{\x}$ is obtained by projecting each entry of the approximate posterior mean $\bmu_{\x}$ onto the constellation set:
\begin{equation*}
\hat{x}_i = \text{arg} \: \underset{s \in \calX_k} {\text{min}} |\bmu_{\x}(i) - s|, \quad \text{for} \quad i=1:n.
\end{equation*}

To obtain a Gaussian approximation to the posterior,
EP replaces each non-Gaussian prior factor $p(x_i)$ in~\eqref{eq:sys-model-bayes} with an unnormalized Gaussian site function $g_i(x_i)$:
\begin{equation*} 
\prod_{i=1}^{n} p(x_i) \sim \prod_{i=1}^{n} g_i(x_i) 
= \prod_{i=1}^{n} \exp\big\{\!-\frac{1}{2} \theta_{i} x_i^2 +  {\gamma}_i x_i  \big\} =\exp\big\{\!-\frac{1}{2} \x^\top \boldsymbol{\Theta} \x + \x^\top \boldsymbol{\gamma}  \big\},
\end{equation*}
where $\boldsymbol{\Theta}=\diag(\theta_1, \ldots, \theta_n)$ and
$\bgamma=[\gamma_1, \ldots, \gamma_n]^\top$.
The unknown parameters $\boldsymbol{\Theta}$ and $\boldsymbol{\gamma}$ are referred to as the site parameters.
The EP algorithm iteratively updates these site parameters,
starting from an initialization $\boldsymbol{\Theta}^{(0)} = \sigma^{-2}_{\x^*} \I$
with $\sigma_{\x^*}^2$ given in \eqref{eq:sigmax}
 and $\boldsymbol{\gamma}^{(0)} = \mathbf{0}$.

At the $t^{th}$ iteration, $t \geq 1$,
the posterior approximation to $p(\x | \y)$ is given by
\begin{align*}
q^{(t)}(\x \,|\, \y ) &= N(\x\!:\!\bmu_{\x}^{(t)}, \bSigma_{\x}^{(t)}) \nonumber \\
&\propto N(\y\!:\!\H \x, \sigma_{\noisevector}^2 \I) \exp\left\{-\frac{1}{2} \x^\top \boldsymbol{\Theta}^{(t-1)}\x + \x^\top \boldsymbol{\gamma}^{(t-1)} \right\} \\
&\propto \exp \left \{ -\frac{1}{2 \sigma_{\noisevector}^2} \| \y - \H \x \|_2^2 \right \} \exp \left \{-\frac{1}{2} \x^\top \boldsymbol{\Theta}^{(t-1)}\x + \x^\top \boldsymbol{\gamma}^{(t-1)} \right \}\\
&\propto \exp\left\{
-\frac{1}{2} \x^\top
\left(\sigma_{\noisevector}^{-2}\H^\top\H + 
\boldsymbol{\Theta}^{(t-1)}
\right) \x
+
\x^\top
\left( \sigma_{\noisevector}^{-2}\H^\top\y +
\boldsymbol{\gamma}^{(t-1)} \right)
\right\},
\end{align*}
which says
\begin{equation}
\bSigma_{\x}^{(t)} =
\left(
\sigma_{\noisevector}^{-2}\H^\top\H +
\boldsymbol{\Theta}^{(t-1)}
\right)^{-1},
\qquad
\bmu_{\x}^{(t)} =
\bSigma_{\x}^{(t)}
\left(
\sigma_{\noisevector}^{-2}\H^\top\y +
\boldsymbol{\gamma}^{(t-1)}
\right).
\label{eq:ep_posterior}
\end{equation}
The rest of this section explains how the parameters in each site function $g_i(x_i)$ are updated at the $t^{th}$ iteration.
According to~\eqref{eq:ep_posterior}, the marginal approximation for $x_i$ is
\begin{equation}
 q^{(t)}(x_i \,|\, \y ) = N (x_i \!:\! \bmu_{\x}^{(t)}(i), \bSigma_{\x}^{(t)}(i, i)).
\label{eq:iga-1}
\end{equation}
EP then removes the approximate site factor $g_i(x_i)$ from this marginal by dividing it out and obtain the cavity distribution:
\begin{equation}
q^{(t) \backslash i}(x_i) = N(x_i\!:\! a^{(t)}_i, (b_i^{(t)})^2) = \frac{q^{(t)}(x_i \,|\, \y )}{g_i(x_i)} = \frac{N (x_i \!:\! \bmu_{\x}^{(t)}(i), \bSigma_{\x}^{(t)}(i, i))}{\exp\{-\frac{1}{2}\theta_{i}^{(t-1)} x_i^2 + {\gamma}_i^{(t-1)} x_i \}},
\label{eq:cavity}
\end{equation}
where
\begin{equation}
(b_i^{(t)})^2 = \frac{\bSigma_{\x}^{(t)}(i, i)}{1 - \bSigma_{\x}^{(t)}(i, i) \theta_{i}^{(t-1)}}, 
\qquad
a^{(t)}_i = (b_i^{(t)})^2 \left( \frac{\bmu_{\x}^{(t)}(i)}{\bSigma_{\x}^{(t)}(i, i)} - {\gamma}_i^{(t-1)} \right).
\label{eq:cavity_density}
\end{equation}
Then the updated $\hat{\theta}^{(t)}_{i}$ and $\hat{\gamma}^{(t)}_i$ are computed by matching the first two moment of the tilted distribution
\begin{equation}
q^{(t)\setminus i}(x_i)
\mathbbm{1}\{x_i\in\calX_k\}
\label{eq:tilted}
\end{equation}
with those of
\begin{equation}
q^{(t) \setminus i}(x_i) \exp\{-\frac{1}{2} \hat{\theta}^{(t)}_{i} x_i^2 + \hat{\gamma}_i^{(t)} x_i \}.
\label{eq:updated}
\end{equation}
Specifically,
the tilted distribution~\eqref{eq:tilted} is supported only on the discrete points in $\calX_k$ and its first two moments are given by
\begin{equation}
\hat{\mu}_i^{(t)}
=
\sum_{s \in\calX_k}
s \,\tilde{p}^{(t)}_i(x_i=s), \qquad
(\hat{\sigma}_i^{(t)})^2
=
\sum_{s\in\calX_k}
s^2\,\tilde{p}^{(t)}_i(x_i=s)
-
\left(\hat{\mu}_i^{(t)}\right)^2,
\label{eq:tilted_moments}
\end{equation}
where
\begin{equation*}
\tilde{p}^{(t)}_i(x_i=s)
=
\frac{
N(s:a_i^{(t)},(b_i^{(t)})^2)
}{
\sum_{c\in\calX_k}
N(c:a_i^{(t)},(b_i^{(t)})^2)
},
\qquad s \in\calX_k.
\end{equation*}
Equating the first two moments of~\eqref{eq:tilted} and~\eqref{eq:updated} gives the updated site parameters of $g_i(x_i)$:
\begin{equation}
\hat{\theta}^{(t)}_{i} = \frac{1}{(\hat{\sigma}_i^{(t)})^2} - \frac{1}{(b_i^{(t)})^2}, \qquad
\hat{\gamma}_i^{(t)} = \frac{\hat{\mu}_i^{(t)}}{(\hat{\sigma}_i^{(t)})^2} - \frac{a_i^{(t)}}{(b_i^{(t)})^2}.
\label{eq:site_update}
\end{equation}

Note that $\hat{\theta}_i^{(t)}$ in~\eqref{eq:site_update} may be negative, 
which can lead to an invalid Gaussian approximation because the corresponding precision parameter should be nonnegative.
Therefore, if $\hat{\theta}_i^{(t)}<0$, EP keeps the previous site parameters by setting
$\hat{\theta}_i^{(t)}=\theta_i^{(t-1)}$
and
$\hat{\gamma}_i^{(t)}=\gamma_i^{(t-1)}$.
Finally, a damping update is applied to obtain the updated site parameters for $i=1:n$ by
\begin{equation}
\theta_i^{(t)}
\leftarrow
(1-\beta)\hat{\theta}_i^{(t)}
+
\theta_i^{(t-1)}, \qquad
\gamma_i^{(t)}
\leftarrow
(1-\beta)\hat{\gamma}_i^{(t)}
+
\beta \gamma_i^{(t-1)},
\label{eq:beta_damping}
\end{equation}
where $\beta\in[0,1]$ is a fixed damping factor.

The terms $\H^\top\H$ and $\H^\top\y$ in~\eqref{eq:ep_posterior} can be precomputed with total cost $O(mn^2)$.
In each EP iteration, 
the computational cost is dominated by computing the Gaussian posterior parameters in~\eqref{eq:ep_posterior}, 
which costs $O(n^3)$.
Once these posterior parameters are computed, 
the site parameters for each $g_i(x_i)$ are updated independently and in parallel.
The total cost of the site updates is $O(|\calX_k|n)$ per iteration.
Therefore, running the EP algorithm for $T$ iterations has total complexity $O(mn^2 + Tn^3 + T|\calX_k|n)$.

\subsection{Motivation for GNN-aided EP}
This section explains the motivation for incorporating GNNs into EP-based MIMO detection.

Although the EP estimator has been shown to outperform various conventional suboptimal estimators across different problem sizes and QAM orders~\cite{cespedes2014expectation}, 
its site-parameter updates are performed in a variable-wise manner.
Specifically, when updating the site parameters for $x_i$, EP uses only the local Gaussian marginal in~\eqref{eq:iga-1} to construct the cavity distribution in~\eqref{eq:cavity}.
This update depends on the marginal mean $\bmu_{\x}^{(t)}(i)$ and marginal variance $\bSigma_{\x}^{(t)}(i,i)$, 
but does not directly use the off-diagonal entries of $\bSigma_{\x}^{(t)}$, 
which encode correlations between different components of $\x$.

As a result, although the Gaussian posterior approximation in EP captures correlations through its full covariance matrix, 
the subsequent site updates are still based on local one-dimensional marginal information.
This variable-wise update may limit the ability of EP to exploit dependencies among transmitted symbols.
GEPNet is proposed to address this limitation by incorporating GNNs into EP.
Specifically, 
GNNs perform message passing on the MRF graph to capture synergy between the components of $x$.
The resulting node representations are then used to update the site parameters in each EP iteration~\cite{kosasih2022graph}.
Building on this GNN-aided EP framework, 
we propose an improved GNN module based on graph convolution in the following sections.

%% file: content/Paper04/Paper04.tex
\chapter{\rm\bfseries Graph-based Denoising Diffusion Method for Estimation}
\label{ch:Paper04}

Chapter~\ref{ch:Paper03} proposes GCEPNet,
which improves the efficiency of the GNN module in the state-of-the-art estimator GEPNet for MIMO detection using graph convolution.
However, 
both GCEPNet and GEPNet are end-to-end estimators that use a fixed number of sequential refinement steps.
This paradigm limits the ability to adapt the accuracy--runtime trade-off at inference time.

 
This limitation motivates the study of a diffusion-based estimation paradigm.
A diffusion-based estimator can use
the same trained denoising network with a flexible number of denoising steps at inference time.
This allows the estimation process to be stopped early for low-latency estimation or extended to obtain higher quality results, and the performance gains typically flatten after several denoising steps~\cite{song2020denoising, ma2025scaling}.
Such controllable inference time computation is a key advantage of the diffusion-based paradigm,
where the number of denoising steps naturally provides a mechanism for trading quality against runtime.
Recent diffusion-based methods for solving hard optimization problems further support the use of this paradigm for generating high quality approximate solutions under flexible inference budgets~\cite{sun2023difusco, sanokowski2024diffusion, li2024fast}

However, 
existing diffusion-based methods for MIMO detection still incur high latency~\cite{luong2025diffusion}. 
We also observe that they become less effective when the linear model is underdetermined.
In addition, the graph convolution form used in GCEPNet in Chapter~\ref{ch:Paper03} relies on the assumption that the channel matrix has full column rank.

To address these limitations, 
this chapter proposes GDIM, 
a graph-based denoising diffusion method that achieves a better accuracy--runtime trade-off for MIMO detection in both overdetermined and underdetermined settings.
Similar to Chapter~\ref{ch:Paper03}, 
GDIM employs graph representation learning to encode the information of problem instances and capture dependencies among transmitted symbols during inference.

Diffusion-based methods generate samples from an unknown target distribution through
a predefined forward corruption process that gradually transform the data into noise and a reverse denoising process to recover the original data.
For MIMO detection,
recent works have adopted score-based diffusion methods to sample from the intractable posterior distribution $p(\x^*=\x\,|\,\y)$~\cite{zilberstein2022annealed, he2024massive}.
Here, the score refers to the gradient of the log-posterior and a local direction for moving a candidate solution toward regions of higher posterior probability.
One representative diffusion-based MIMO detector is based on annealed Langevin dynamics (ALD)~\cite{zilberstein2022annealed}.
It computes a closed-form posterior score using the SNIPS framework, which relies on the singular value decomposition (SVD) of the channel matrix~\cite{kawar2021snips}.
More recently,
approximate diffusion detection (ADD) is proposed as an SVD-free alternative~\cite{he2024massive}.
Instead of deriving the posterior score analytically, ADD approximates it by combining an iterative detector with Tweedie's identity~\cite{he2024massive}.
Both ALD and ADD define the forward and reverse process in a relaxed continuous space.
As shown in our numerical results, 
both ALD and ADD suffer from two key limitations.
First, they require extensive sequential sampling iterations at inference time, 
which leads to high latency~\cite{luong2025diffusion}.
Second,
their detection performance degrades in underdetermined MIMO settings, 
where the posterior score becomes unreliable for guiding the reverse process.

These limitations motivate a shift from analytic score-based sampling to a learned reverse process parameterized by a denoising network. 
The denoising network refines a noisy candidate into a cleaner estimate of the transmitted signal at each step, 
providing stronger guidance than a reverse process driven solely by the posterior score in underdetermined settings.
In addition, the denoising network parameters are shared across all denoising steps, enabling fast inference with a small number of reverse updates and fewer network evaluations. 
Recent advances in denoising diffusion have shown strong performance in generative modeling over discrete domain,
such as text generation and combinatorial optimization~\cite{austin2021structured, sun2023difusco, lou2023discrete}.
To the best of our knowledge, this denoising diffusion paradigm has not yet been explored for MIMO detection.

In this chapter, 
we use the diffusion framework as a parameter sharing mechanism for iterative refinement.
A single denoising network is trained across different noise levels and can be repeatedly applied for a flexible number of reverse steps at inference time.
The injected noise creates a sequence of progressively harder reconstruction problems around the clean transmitted signal.
By training across these noise levels, 
the denoising network learns a reusable refinement operator that moves noisy candidates toward solutions consistent with the observed MIMO problem instance.
This design provides a natural mechanism for controlling the accuracy--runtime trade-off without training separate models for different numbers of refinement steps.

This chapter proposes GDIM, a graph-based denoising diffusion method for MIMO detection.
The proposed method considers both discrete and continuous diffusion formulations.
Section~\ref{sec:discrete_diffusion} introduces the discrete diffusion formulation, where noisy candidates remain constrained to the discrete constellation space $\calX_k^n$.
Section~\ref{sec:continuous_diffusion} introduces the continuous diffusion formulation, where the forward process corrupts the transmitted signal in a continuous space.
For both formulations, Section~\ref{sec:graph_denoising} presents the gated graph message passing-based denoising network used in GDIM for MIMO detection.
The denoising network encodes each problem instance $\{\y,\H,\sigma_{\noisevector}\}$ through graph features, allowing the denoising process to be conditioned on the problem instance.
Compared with the discrete formulation, the continuous diffusion formulation provides additional advantages by explicitly incorporating the likelihood of the linear model at each denoising step.
This design allows the denoising network to use fewer layers and lower computational complexity, while also achieving better performance.
Section~\ref{sec:optimization} describes the training objectives, and Section~\ref{sec:fast_inference} presents the inference strategies.
Section~\ref{sec:exp} reports numerical results comparing GDIM with classical methods, learning-based estimators, 
and existing diffusion-based methods.
The results demonstrate that GDIM achieves a more favorable trade-off between performance and runtime in both overdetermined and underdetermined settings, thereby addressing the limitations discussed above.

Note that GDIM adopts a different formulation from many diffusion-based methods for inverse problems. In many such methods, the signal prior is not available in closed form and is therefore learned from clean data using an unconditional diffusion model; the learned prior is then combined with the observation likelihood through guidance at inference time~\cite{chung2023diffusion, feng2023score}. In MIMO detection, by contrast, the prior over the transmitted symbols is explicitly known, typically as a uniform distribution over the finite constellation set. Therefore, GDIM does not use diffusion to learn an unknown signal prior. Instead, the main challenge is to approximate the computationally intractable posterior induced by the observation model. Accordingly, the GDIM denoising network is trained conditionally on the MIMO problem instance $\{\y,\H,\sigma_{\noisevector}\}$, so that the observation information is incorporated directly into the learned reverse process rather than introduced only through an external guidance term at inference time.

\section{Chapter Preliminaries}
\label{sec:pre}

In this chapter, 
to simplify the presentation of our method, 
we transform the real-valued MIMO system in~\eqref{eq:mimo-sys-model-real} so that the constraint set becomes $\{\pm \frac{1}{2^k - 1}, \pm \frac{3}{2^k - 1}, \ldots, \pm 1 \}$
whose elements are bounded within $[-1,1]$.
We define the transformed variables as
\begin{align*}
\x^*_2  = \frac{1}{2^k - 1}\x^*, \ \H_2 = (2^k - 1) \H, \ \calX_k^{(2)} = \{\pm \frac{1}{2^k - 1}, \pm \frac{3}{2^k - 1}, \ldots, \pm 1 \}.
\end{align*}
Then~\eqref{eq:mimo-sys-model-real} becomes
\begin{equation*}
\y = \H_2 \x^*_2 + \noisevector, \noisevector \sim \calN(\0, \sigma_{\noisevector}^2\I),
\end{equation*}
where $\x^*_2 \in \calX_k^{(2)}$.
For notational simplicity, 
we reuse $\x^*$, $\H$, 
and $\calX_k$ to denote the transformed variables 
$\x^*_2$, $\H_2$, 
and $\calX_k^{(2)}$, respectively, 
in the rest of this chapter.
We denote by $s_j$ the $j^{th}$ smallest element in $\calX_k$.
We also denote the mean symbol energy of $\calX_k$ as $\sigma_{\x^*}$ in this chapter.
To simplify notation, we denote $p(\x^*=\x\,|\,\y)$, $p(\y\,|\,\x^*=\x)$ and $p(\x^*=\x)$ 
by $p(\x\,|\,\y)$, $p(\y\,|\,\x)$ and $p(\x)$,
respectively.


\section{Diffusion-based Method in GDIM}
\label{sec:dm}

A diffusion model is characterized by a predefined forward corruption process and a learnable reverse denoising process~\cite{sohl2015deep, ho2020denoising, song2020score}.
For MIMO detection,
the target distribution is the unknown posterior $p(\x \,|\, \y)$. 
In our formulation,
the forward corruption process is independent of the problem instance. 
Specifically,
for a training sample with ground-truth transmitted signal $\x_0 = \x^*$, 
the forward process generates a sequence of latent variables $\x_{1:T} = (\x_1, \ldots, \x_T)$ with progressively increasing noise, such that the terminal distribution of $\x_T$ approaches a simple known distribution $p_T$.
The problem instance is incorporated into the reverse denoising process as input to the denoising network.
Inference starts from an initialization $\hat{\x}_T\sim p_T$,
which is then iteratively denoised by the trained denoising network to produce an estimate of $\x_0$ for the given problem instance.

This section introduces the discrete and continuous diffusion-based methods used in GDIM.
The discrete formulation keeps the latent variables $\x_t$ in the constrained discrete space $\calX_k^n$, 
whereas the continuous formulation relaxes this constraint.
For both formulations, 
the reverse denoising problem is reduced to predicting the clean transmitted signal $\x_0$ at a given noise level for the given problem instance.


\subsection{Discrete Diffusion-based Method}
\label{sec:discrete_diffusion}


In this subsection,
we consider a discrete diffusion-based method for MIMO detection.
We adopt discrete diffusion models with multinomial noise~\cite{ho2020denoising, austin2021structured, hoogeboom2021argmax},
where the forward Markov process corrupts the data by applying a transition matrix at each step.
We represent the categorical distribution of the $n$-dimensional variable $\x_t \in \calX_k^n$ by a probability matrix 
$\P(\x_t) \in \Rbb^{n\times 2^k}$, 
where $\P_{ij} = \Pr(\x_t(i) = s_j)$.
Given $\x_{t-1}$, 
$\x_t$ is sampled as
\begin{equation*}
\x_t \sim \mathrm{Cat}\left(\x_t; \1_{\calX_k}(\x_{t-1}) \Q_t\right),
\end{equation*}
where the $\1_{\calX_k}(\x_{t-1})$ is the one-hot encoding of $\x_{t-1}$ defined in~\eqref{eq:one-hot},
and $\Q_t \in \Rbb^{2^k \times 2^k}$ is the transition matrix at step $t$,
where $\Q_t(a, b)$ denotes the probability of transitioning from state $s_a$ at step $t-1$ to state $s_b$ at step $t$.
Equivalently, $\x_t$ can be sampled directly from $\x_0$ as 
\begin{equation}
\x_t \sim \mathrm{Cat}\left(\x_t; \1_{\calX_k}(\x_0) \bar{\Q}_t\right), \quad
\bar{\Q}_t = \Q_1\cdots\Q_{t-1}\Q_t=\bar{\Q}_{t-1}\Q_t.
\label{eq:forward-multi}
\end{equation}
As the off-diagonal mass of the cumulative transition matrix $\bar{\Q}_t$ increases,
the noise level in $\x_t$ also increases. 
We choose $\Q_t$ so that the noise level introduced by $\bar{\Q}_t$ increases gradually with $t$,
and the terminal distribution satisfies $\x_T \sim \mathrm{Uniform}(\calX_k)^{n}$. 
In this way,
$\x_t$ becomes increasingly different from $\x_0$ as
the diffusion step $t$ increases,
while the final $\x_T$ follows the uniform distribution from which reverse sampling can be initialized.
In practice, we construct $\Q_t$ using discretized Gaussian probabilities that account for the ordinal nature of elements in $\calX_k$,
which also improves the training.
The detailed configuration of $\Q_t$ are provided in Section~\ref{sec:exp}. 


For discrete reverse denoising,
we consider a transition from a higher noise level $t_2$ to a lower noise level $t_1$, where $0 \leq t_1 < t_2 \leq T$.
Define
\begin{equation*}
\bar{\Q}_{t_1\rightarrow t_2} = \Q_{t_{1} + 1} \Q_{t_{1} + 2} \cdots \Q_{t_2}.
\end{equation*}
By Bayes' rule, the exact reverse distribution $\P(\x_{t_1} \,|\, \x_{t_2}, \x_0)$ could be derived from the forward process if the clean signal $\x_0$ is known:
\begin{align}
\P(\x_{t_1} \,|\, \x_{t_2}, \x_0) 
&\propto \, \P(\x_{t_2} \,|\, \x_{t_1}, \x_0) \odot \P(\x_{t_1} \,|\, \x_0) = \P(\x_{t_2} \,|\, \x_{t_1}) \odot \P(\x_{t_1} \,|\, \x_0) \nonumber \\
&\propto \, \left(\1_{\calX_k}(\x_{t_2}) \bar{\Q}_{t_1\rightarrow t_2}^\top \right) \odot 
\left(  \1_{\calX_k}(\x_0) \bar{\Q}_{t_1} \right).
\label{eq:d3pm_reverse}
\end{align}
The right-hand side is normalized row-wise so that each row defines a valid categorical distribution over $\calX_k$.
However, $\x_0$ is unknown at inference time.
Therefore, the reverse distribution does not have a closed-form expression and is therefore parameterized and approximated using a neural network.
Following \cite{ho2020denoising, hoogeboom2021argmax, austin2021structured},
we train a denoising network to predict the clean transmitted signal $\x_0$ from the noisy latent variable $\x_t$ for the problem instance.
The network parameterized by $\btheta$ outputs the conditional distribution $\P_{\btheta}(\hat{\x}_0 | \x_t, \y) \in \Rbb^{n \times 2^k}$,
where the $i^{th}$ row gives the predicted categorical distribution of $\x_0(i)$ over $\calX_k$.
Given this prediction, 
we construct the reverse distribution by replacing the unknown one-hot representation $\1_{\calX_k}(\x_0)$ in~\eqref{eq:d3pm_reverse} with the network prediction $\P_{\btheta}(\hat{\x}_0 | \x_{t_2}, \y)$ and obtain
\begin{equation}
\P_{\btheta}(\x_{t_1} \,|\, \x_{t_2}, \y) \propto \,  \left(\1_{\calX_k}(\x_{t_2}) \bar{\Q}_{t_1\rightarrow t_2}^\top \right) \odot 
\left( \P_{\btheta}(\hat{\x}_0 | \x_{t_2}, \y) \bar{\Q}_{t_1} \right),
\label{eq:reverse}
\end{equation}
and the latent variable corresponding to $t_1$ is sampled as
\begin{equation*}
\x_{t_1} \sim \mathrm{Cat}\left( \x_{t_1}; \P_{\btheta}(\x_{t_1} \,|\, \x_{t_2}, \y) \right). 
\end{equation*}
Repeating this reverse denoising process from $\x_T$ yields the final estimate of $\x_0$.


\subsection{Continuous Diffusion-based Method}
\label{sec:continuous_diffusion}

For the continuous formulation,
we adopt the denoising diffusion probabilistic models (DDPM) framework~\cite{ho2020denoising}.
After the transformation in Section~\ref{sec:pre}, 
each entry of $\x_0$ lies in the discrete set $\{-1, -\frac{2^k-3}{2^k-1}, \ldots, \frac{2^k-3}{2^k-1}, 1\}$.
The forward process generates latent variables by adding continuous Gaussian noise to $\x_0$.
We denote these latent variables by $\{ \tilde{\x}_t \}_{t=1:T}$ to distinguish them from the latent variables in the discrete diffusion in Section~\ref{sec:discrete_diffusion}.

In the forward process, we treat $\tilde{\x}_0 = \x_0$ as a continuous variable.
Given $\tilde{\x}_{t-1} \in \Rbb^n$, 
$\tilde{\x}_t \in \Rbb^n$ is sampled as
\begin{equation}
\tilde{\x}_t \sim p(\tilde{\x}_t \;|\; \tilde{\x}_{t-1})  = N(\tilde{\x}_t \!:\! \sqrt{1 - \beta_t} \tilde{\x}_{t-1}, \beta_t \I),   
\label{eq:ddpm_forward}
\end{equation}
where the hyperparameter $\beta_t$ is chosen such that $\prod_{t=1}^T (1-\beta_t) \approx 0$, 
so that the terminal distribution is close to $N(\0, \I)$~\cite{ho2020denoising}.
Let $\alpha_t = \prod_{\ell=1}^t (1-\beta_{\ell})$,
then~\eqref{eq:ddpm_forward} is equivalent to directly sampling $\tilde{\x}_t$ via
\begin{equation}
\tilde{\x}_t \sim p(\tilde{\x}_t \;|\; \x_0) = N(\tilde{\x}_t \!:\! \sqrt{\alpha_t} \x_0, (1 - \alpha_t) \I).
\label{eq:ddpm_forward2}
\end{equation}

For reverse denoising,
we adopt the deterministic denoising diffusion implicit models (DDIM) update~\cite{song2020denoising}, 
which is compatible with the forward process defined in~\eqref{eq:ddpm_forward2}.
Similar to the discrete formulation,
here we also consider a transition from a higher noise level $t_2$ to a lower noise level $t_1$, where $0\leq t_1<t_2\leq T$.
Given a point estimate $\hat{\x}_0$ of the clean signal predicted from $\tilde{\x}_{t_2}$ and the problem instance, 
the reverse update for obtaining the latent variable at a lower noise level $t_1$ is
\begin{align*}
\tilde{\x}_{t_1} &= \sqrt{\alpha_{t_1}} \hat{\x}_0 + \sqrt{1 - \alpha_{t_1}} \frac{\tilde{\x}_{t_2} - \sqrt{\alpha_{t_2}} \hat{\x}_0}{\sqrt{1-\alpha_{t_2}}} 
\end{align*}

Based on~\eqref{eq:ddpm_forward2},
we further derive an approximate posterior for estimating the original clean signal $\x_0 \in \calX_k^n$ from the noisy sample $\tilde{\x}_t$ and the problem instance $\{ \y, \H, \sigma_{\noisevector} \}$.
Using the approximation
$p(\x_0) \approx N(\0, \sigma_{\x^*}^{2} \I)$
and the conditional independence of $\y$ and $\tilde{\x}_t$ given $\x_0$, we have
\begin{align*}
p(\x_0 \;|\; \tilde{\x}_t, \y) &\propto \, p(\y, \tilde{\x}_t \;|\; \x_0)p(\x_0) = p(\y \;|\; \x_0) p(\tilde{\x}_t \;|\; \x_0) p(\x_0) \nonumber\\
&\approx \, N(\y \!:\! \H \x_0, \sigma_{\noisevector}^2 \I) \, N(\tilde{\x}_t \!:\! \sqrt{\alpha_t} \x_0, (1 - \alpha_t) \I) \, N(\x_0 \!:\! \0, \sigma_{\x^*}^{2} \I) \nonumber\\
&\propto \, 
\exp \left \{ -\frac{1}{2 \sigma_{\noisevector}^2} \| \y - \H \x_0 \|_2^2 \right \} 
\exp \left \{ -\frac{1}{2 (1 - \alpha_t)} \| \tilde{\x}_t - \sqrt{\alpha_t} \x_0 \|_2^2 \right \}
\exp \left \{ -\frac{1}{2 \sigma_{\x^*}^2}  \| \x_0  \|_2^2 \right \} \nonumber\\
&\propto \,  \exp \left \{ -\frac{1}{2} \x_0^\top \left( 
\sigma_{\noisevector}^{-2} \H^\top \H + \frac{\alpha_t}{1-\alpha_t} \I + \sigma_{\x^*}^{-2} \I \right) \x_0 
+ \x_0^\top \left( \sigma_{\noisevector}^{-2} \H^\top \y + \frac{\sqrt{\alpha_t}}{1-\alpha_t} \tilde{\x}_t \right) \right \} \nonumber\\
&=N(\x_0 \!:\! \bmu_{t}, \bSigma_t), \nonumber
\end{align*}
where
\begin{equation}
\bSigma_t = \left( \sigma_{\noisevector}^{-2} \H^\top \H + \frac{\alpha_t}{1-\alpha_t} \I + \sigma_{\x^*}^{-2} \I \right)^{-1}, \qquad  
\bmu_{t} = \bSigma_t \left( \sigma_{\noisevector}^{-2} \H^\top \y + \frac{\sqrt{\alpha_t}}{1-\alpha_t} \tilde{\x}_t \right).
\label{eq:ddpm_x0_pred}
\end{equation}
Therefore, in the denoising network, whose training goal is to predict $\x_0 \in \calX_k^n$ from $\tilde{\x}_t$ and the problem instance,
we use the corresponding $\{\bmu_{t}, \bSigma_t \}$ in~\eqref{eq:ddpm_x0_pred} as the input to the network.

\section{Gated Graph Message Passing-based Denoising Network for GDIM}
\label{sec:graph_denoising}

In this section, 
we propose a gated graph message passing-based denoising network for diffusion-based MIMO detection.
This design is inspired by graph-based denoising networks used in diffusion methods for solving the traveling salesman problem~\cite{joshi2022learning,qiu2022dimes,sun2023difusco}.
In GDIM, the denoising network aims to predict the clean transmitted signal $\x_0$ at each noise level for a given MIMO problem instance.

We denote the denoising network by 
$f_{\btheta}(\x_{n}, \mathbf{n}, \mathcal{G}_{\mathrm{MIMO}}) \in \Rbb^{n \times 2^k}$,
where $\x_{n}$ and $\mathbf{n}$ denote $n$-dimensional vectors representing the noisy state and the noise level of each element to be denoised, respectively,
and $\mathcal{G}_{\mathrm{MIMO}}$ is a graph constructed from the problem instance $\{ \y, \H, \sigma_{\noisevector} \}$.
The output of $f_{\btheta}$ is a categorical probability matrix over $\calX_k$, whose $i^{th}$ row gives the predicted distribution of $x_0(i)$.

The network architecture is the same for both the discrete and continuous diffusion formulations introduced in Section~\ref{sec:dm}, 
but the two formulations use separately trained network parameters and different inputs to represent the noisy state and noise level.
To avoid confusion, we denote the parameters of the discrete denoising network by $\btheta_{\mathrm{d}}$ and those of the continuous denoising network by $\btheta_{\mathrm{c}}$.
Specifically, 
the discrete diffusion performs denoising via $f_{\btheta_{\mathrm{d}}}(\x_t, t\mathbf{1}, \mathcal{G}_{\mathrm{MIMO}})$,
where the noisy state is the discrete latent variable $\x_t \in \calX_k^n$, 
and the noise level is represented by the diffusion step $t$.
For the continuous diffusion,
the continuous latent variable $\tilde{\x}_t$ and the step $t$ are not directly used as network inputs.
Instead, we propose to use $f_{\btheta_{\mathrm{c}}}(\bmu_{t}, \diag(\bSigma_t), \mathcal{G}_{\mathrm{MIMO}})$ where
$\bmu_{t}$, $\diag(\bSigma_t)$ are defined in~\eqref{eq:ddpm_x0_pred}.
In this case, $\bmu_t$ provides a likelihood-informed estimate of the clean signal based on $\tilde{\x}_t$, 
while $\diag(\bSigma_t)$ provides element-wise uncertainty information.
Compared with using $\tilde{\x}_t$ and $t$ directly, 
$\bmu_t$ and $\diag(\bSigma_t)$ provide a more informative input for predicting $\x_0$.
The rest of this section describes the architecture of the denoising network.

\paragraph{MIMO graph construction}
While the forward diffusion process corrupts each entry $x^*(i)$ independently,
the true posterior couples all symbols through the received signal $\y$ and the channel matrix $\H$.
To capture these dependencies,
we represent each problem instance as a fully connected graph,
where nodes correspond to the entries of $\x^*$,
and node and edge features are defined using the instance information $\{ \y, \H, \sigma_{\noisevector} \}$.
The network then performs message passing on this graph.
This graph-based formulation allows the denoising process to effectively exploit the conditional information.

For a given instance $\{\y,\H,\sigma_{\noisevector}\}$, 
we define a fully connected graph with self-loops,
$\mathcal{G}_{\mathrm{MIMO}}=(\mathcal{V},\mathcal{E})$, 
where each node $i\in\mathcal{V}$ represents the transmitted symbol $x_i^*$.
In $\mathcal{G}_{\mathrm{MIMO}}$, an edge $e_{ij}$ exists between each pair of nodes $(i,j)$.
To inject problem-specific inductive bias into the denoising network,
we follow the pairwise MRF factorization of the posterior~\cite{scotti2020graph} and initialize the node feature $\noisevector_i \in \Rbb^{3}$, 
and edge feature $\e_{ij} \in \Rbb^{2}$ as defined in~\eqref{eq:mrf_mimo_graph_feature}. 
The node features encode local evidence associated with symbol $x_i^*$, 
while the edge features encode pairwise relationships induced by the channel matrix.
Unlike the graph defined in Section~\ref{sec:mimo_graph_cov} of Chapter~\ref{ch:Paper03}, 
this form does not rely on the full column rank assumption of $\H$.

\paragraph{Node and edge representation learning} 
In addition to the instance graph, 
the denoising network must also incorporate the current noisy state.
To this end, 
we encode the noisy state $\x_n$ and the noise level $\n$ as graph signals using the sinusoidal embeddings~\cite{vaswani2017attention}.
Specifically, define the sinusoidal map $\f(\cdot; \tau, H): \Rbb \rightarrow \Rbb^{H}$ by
\begin{equation*}
\f(x; \tau, H)_{2j-1} = \mathrm{sin} \left( x / (\tau)^{2j/H} \right), \quad
\f(x; \tau, H)_{2j} = \mathrm{cos} \left( x / (\tau)^{2j/H} \right), 
\end{equation*}
where the temperature $\tau$ and hidden dimension $H$ are hyperparameters, and $j=1:\frac{H}{2}$.
The state embedding 
$\X \in \Rbb^{H \times n}$ and noise level embedding
$\mathbf{N} \in \Rbb^{H \times n}$ are defined as
\begin{equation*}
\X_{:, i} = \f(x_n(i); 10000, H), \quad \N_{:, i} = \f(n(i); 10000, H).
\end{equation*}

We next describe the gated graph message passing at each layer of the denoising network.
For each node $i$ and edge $(i, j)$, 
we first compute the initial node embedding $\noisevector_i^{(1)}$ and the edge embedding $\e_{ij}^{(1)}$ as
\begin{align*}
&\noisevector_i^{(1)} = \mathrm{MLP}_1\left( [\mathrm{MLP}_2(\noisevector_i); \X_{:, i}] \right) \in \Rbb^{H}, 
\\
&\e_{ij}^{(1)} = \mathrm{MLP}_{\mathrm{e}}\left( \e_{ij} \right) \in \Rbb^{H}.
\end{align*}
At layer $\ell$,
a relational embedding $\mathbf{r}_{ij}^{(\ell)}$ is computed for each edge $(i, j)$:
\begin{align*}
\mathbf{r}_{ij}^{(\ell)} = \mathrm{MLP}_r^{(\ell)}\left([\e_{ij}^{(\ell)}; \noisevector_i^{(\ell)}; \noisevector_j^{(\ell)}] \right) \in \Rbb^{H}.
\end{align*}
This relational embedding determines how strongly node $j$ should influence node $i$,
and allows the aggregation weights to depend on both endpoints and their interaction.
Adopting edge gating mechanisms~\cite{bresson2018experimental, joshi2022learning}, 
the gated neighborhood aggregation is defined as
\begin{equation*}
\hat{\noisevector}_i^{(\ell)} = \mathrm{MLP}_3^{(\ell)} (\noisevector_i^{(\ell)}) + \sum_j \sigma \left( \mathbf{r}_{ij}^{(\ell)} \right) \odot \mathrm{MLP}_4^{(\ell)} (\noisevector_j^{(\ell)}),
\end{equation*}
where $\sigma(\cdot)$ denotes the element-wise sigmoid function.
Finally,
the edge and node embeddings are updated using residual connections~\cite{he2016deep} and noise level conditioning:
\begin{align*}
&\noisevector_i^{(\ell+1)} = \noisevector_i^{(\ell)} + \mathrm{ReLU} \left( \hat{\noisevector}_i^{(\ell)} \right) + \mathrm{MLP}_{n}^{(\ell)}( \N_{:, i} ), 
\\
&\e_{ij}^{(\ell+1)} = \e_{ij}^{(\ell)} + \mathbf{r}_{ij}^{(\ell)}.
\end{align*}
The term $\mathrm{MLP}_{n}^{(\ell)}(\N_{:, i})$ injects element-wise noise level information into each layer. 
This graph message passing on $\mathcal{G}_{\mathrm{MIMO}}$ enables the network to jointly exploit the problem instance $\{\y, \H, \sigma_{\noisevector}\}$ and the current diffusion state $\{\x_n, \n\}$.

\paragraph{Readout layer for ordinal discrete distribution} 
At the output layer, 
the node embedding $\noisevector_i^{(L)}$ parameterizes the predicted distribution 
$[p_1, \ldots, p_{2^k}]^\top$ for $x_0(i)$ over $\calX_k$, 
where
$p_c = \Pr(x_0(i)=s_c)$,
$c = 1:2^k$. 
We now describe the process to obtain $\{ p_c \}_{c=1}^{2^k}$.
Since the values in $\calX_k$ are ordinal,
we incorporate this structure by modeling the output distribution with a discretized 
truncated logistic distribution~\cite{austin2021structured}.
Specifically, 
for each $x_0(i)$, 
the output layer computes the location $\mu_i$ and the scale $\sigma_i$ of a logistic distribution via
\begin{align}
&(\hat{\mu}_i, \: \sigma_i) = \mathrm{CNN}\left(\mathrm{ReLU}\left(\noisevector_i^{(L)} \right) \right), \nonumber \\
&\mu_i = \mathrm{tanh}\left(\hat{\mu}_i + x_n(i) \right), 
\label{eq:loc}
\end{align}
where $\mathrm{CNN}$ is a convolutional layer with two output channels.
In~\eqref{eq:loc}, 
the model learns a residual correction around the noisy state $x_n(i)$,
while $\mathrm{tanh}(\cdot)$ activation keeps $\mu_i$ within $[-1, 1]$.
Given the location $\mu_i$ and scale $\sigma_i$, 
we discretize the logistic distribution 
using $2^k$ evenly spaced bin centers:
\begin{equation*}
b_c = -1 + \frac{2 (c-1)}{2^k - 1},\quad c = 1, \ldots, 2^k,
\end{equation*}
with bin width $\Delta = \frac{2}{2^k - 1}$.
Accordingly, the bin boundaries for class $c$ are
\begin{equation*}
b_c^{-} = b_c - \frac{\Delta}{2}, \quad b_c^{+} = b_c + \frac{\Delta}{2},
\end{equation*}
matching the $c^{th}$ smallest constellation point in $\calX_k$ defined in Section~\ref{sec:pre}.
The probability for class $c$ is then computed as $p_c = F(b_c^{+}; \mu_i, \sigma_i) - F(b_c^{-}; \mu_i, \sigma_i)$, where $F$ denotes the CDF of the logistic distribution.
The resulting probabilities $\{ p_c \}_{c=1}^{2^k}$ concentrate on classes near $\mu_i$ and decay smoothly as the classes move farther away.
This design assigns higher probability to ``near-miss" predictions than to distant ones,
which is consistent with the ordinal structure of $\calX_k$ and leads to improved model performance.

\begin{algorithm}[t]
\caption{Training of the GDIM denoising network $f_{\btheta_{\mathrm{d}}}$ for discrete diffusion}
\label{alg:gd4_training_discrete}
\begin{algorithmic}[1]
\FOR{each training iteration}
\STATE Generate a problem instance $\{ \H,\y,\sigma_\noisevector, \x_0=\x^* \}$
\STATE Sample $t \sim \mathrm{Uniform}(\{1,\ldots,T\})$
\STATE Sample $\x_t \sim \P(\x_t \,|\, \x_0)$ via~\eqref{eq:forward-multi}
\STATE Compute $\P_{\btheta_{\mathrm{d}}}(\hat{\x}_0 \,|\, \x_t,\y) = f_{\btheta_{\mathrm{d}}}(\x_t, t\mathbf{1}, \mathcal{G}_{\mathrm{MIMO}})$
\STATE Compute $\P_{\btheta_{\mathrm{d}}}(\x_{t-1} \,|\, \x_t,\y)$ via~\eqref{eq:reverse}
\STATE Take gradient descent step on $\nabla_{\btheta_{\mathrm{d}}} \mathcal{L}$ where
\begin{align*}
&\mathcal{L}_{\mathrm{vb}}
=
\sum_i
\mathrm{KLD}\!\left(
\P(\x_{t-1} |  \x_t,\x_0)_{i,:}
\,\middle\|\,
\P_{\btheta_{\mathrm{d}}}(\x_{t-1} | \x_t,\y)_{i,:}
\right)\\
&\mathcal{L}_{\mathrm{ce}}
=
-\mathrm{trace}\!\left(
\1_{\calX_k}(\x_0)^\top
\log \P_{\btheta_{\mathrm{d}}}(\hat{\x}_0 \,|\, \x_t,\y)
\right)\\
&\mathcal{L} = \mathcal{L}_{\mathrm{vb}} + \mathcal{L}_{\mathrm{ce}}
\end{align*}
\ENDFOR
\end{algorithmic}
\end{algorithm}

\begin{algorithm}[t]
\caption{Training of the GDIM denoising network $f_{\btheta_{\mathrm{c}}}$ for continuous diffusion}
\label{alg:gd4_training_continuous}
\begin{algorithmic}[1]
\FOR{each training iteration}
\STATE Generate a problem instance $\{ \H,\y,\sigma_{\noisevector},\x_0=\x^* \}$
\STATE Sample $t \sim \mathrm{Uniform}(\{1, \ldots, T\})$
\STATE Sample $\tilde{\x}_t \sim p(\tilde{\x}_t \,|\, \x_0)$ using~\eqref{eq:ddpm_forward2}
\STATE Compute $\bmu_t$ and $\bSigma_t$ using~\eqref{eq:ddpm_x0_pred}
\STATE Compute $\P_{\btheta_{\mathrm{c}}}(\hat{\x}_0 \,|\, \tilde{\x}_t,\y) =
f_{\btheta_{\mathrm{c}}}(\bmu_t,\diag(\bSigma_t),\mathcal{G}_{\mathrm{MIMO}})$
\STATE Take a gradient descent step on $\nabla_{\btheta_{\mathrm{c}}}\mathcal{L}_{\mathrm{c}}$, where
\begin{equation*}
\mathcal{L}_{\mathrm{c}}
=
-\operatorname{trace}\!\left(
\1_{\calX_k}(\x_0)^\top
\log \P_{\btheta_{\mathrm{c}}}(\hat{\x}_0 \mid \tilde{\x}_t,\y)
\right).
\end{equation*}
\ENDFOR
\end{algorithmic}
\end{algorithm}

\section{Optimization}
\label{sec:optimization}
Algorithms~\ref{alg:gd4_training_discrete} and~\ref{alg:gd4_training_continuous} summarize the training procedures for the discrete and continuous GDIM denoising networks, respectively.
Both formulations train the denoising network to predict the clean transmitted signal $\x_0$ at a randomly sampled noise level.

For the discrete diffusion formulation, the training objective contains the variational loss and the cross-entropy loss~\cite{austin2021structured, sun2023difusco}.
The variational loss $\mathcal{L}_{\mathrm{vb}}$ minimizes the Kullback--Leibler divergence between the true reverse posterior induced by the forward process and the model-predicted reverse distribution.
The exact reverse posterior $\P(\x_{t-1}\,|\, \x_t, \x_0)$ is available during training because the clean signal $\x_0$ is known and can be computed via~\eqref{eq:d3pm_reverse}.
The cross-entropy loss $\mathcal{L}_{\mathrm{ce}}$ encourages accurate prediction of the clean signal $\x_0$ from the noisy state $\x_t$ and the problem instance.

For the continuous diffusion formulation,
a common training objective is to predict the clean data $\x_0$ from the corrupted data $\tilde{\x}_t$ at diffusion step $t$. 
Denote this prediction as $\hat{\x}_0 (\tilde{\x}_t, t)$,
then the loss is defined as $\mathbb{E}_{t, \x_{0}} [\|\x_0 - \hat{\x}_0 (\tilde{\x}_t, t) \|_2^2]$~\cite{salimans2022progressive, ho2020denoising}.
For MIMO detection, since $\x_0$ is constrained to $\calX_k^n$, 
the denoising network for GDIM treats the $\x_0$ prediction problem as a classification problem and is therefore trained using the cross-entropy loss,
which directly accounts for the discrete constraint on $\x_0$.


\section{Fast Inference Strategy}
\label{sec:fast_inference}
To reduce inference cost, 
we consider step-skipped reverse inference~\cite{song2020denoising, austin2021structured} to reduce the number of evaluating the denoising network,
which applies to both the discrete and continuous diffusion formulations.
It reduces the number of denoising steps by moving directly between a selected subset of noise levels.
Instead of traversing the full reverse chain at noise levels corresponding to diffusion steps $T, T-1, \ldots, 1$,
step-skipped inference uses $K << T$ reverse steps along a shortened reverse chain $t_K, \ldots, t_1$, with $t_K = T$ and $t_1 = 0$.

\section{Numerical Results}
\label{sec:exp}

This section evaluates the performance of the proposed GDIM.
Section~\ref{sec:setup} describes the experimental setup,
and Section~\ref{sec:runtime} provides the complexity and runtime comparison.
In Section~\ref{sec:gdim_compare_baselines},
we first compare the discrete and continuous formulations of GDIM, 
showing that the continuous formulation achieves better detection performance with lower computational cost.
We then compares the continuous GDIM with 
classical baselines including the $K$-best Klain-Babai point~\cite{Wan25},
and EP~\cite{cespedes2014expectation},
and diffusion-based methods including ALD~\cite{zilberstein2022annealed} and ADD~\cite{he2024massive},
and learning-based methods, 
including GEPNet~\cite{kosasih2022graph} and OAMPNet~\cite{he2018model}.
The evaluation covers both overdetermined and underdetermined settings.
For learning-based methods, the networks are trained on overdetermined problem instances and are also tested on underdetermined instances to evaluate their generalization ability.


\begin{table}[t]
\caption{Average detection runtime per problem instance ($N_t = 32$, $N_r = 32$, $16$-QAM)}
\begin{center}
\resizebox{\textwidth}{!}{
\begin{tabular}{l l l}
\toprule
Method & Runtime (seconds) & Complexity \\ 
\toprule
10-best Klein-Babai & $1.66 \times 10^{-3}$ & $\mathcal{O}(mn^2 + K n^2)$, $K=10$\\ \midrule
EP & $9.40 \times 10^{-4}$ & $\mathcal{O}(T n^3)$, $T=10$\\ \midrule
ALD & $2.35 \times 10^{-1}$ & \makecell[l]{$\mathcal{O}(mn^2 + L T (n^2 + | \calX_{k} | n))$, $L=20$, $T=70$} \\ \midrule
ADD & $1.65 \times 10^{-2}$ & \makecell[l]{$\mathcal{O}(mn^2 + T (mn + T_{\text{iter}} n^2 + | \calX_{k} | n)$, $T = 70$, $T_{\text{iter}} = 20$} \\ \midrule
OAMPNet & $1.30 \times 10^{-3}$ & \makecell[l]{$\mathcal{O}(T (m^3 + n^3 + m^2 n + n^2 m))$, $T=10$} \\ \midrule
GEPNet, $T = 5$ & $2.60 \times 10^{-3}$ & \multirow{2}{*}{\makecell[l]{$\mathcal{O}(mn^2 + T n^3 + T L n^2 (N_u N_{h_1} + N_{h_1} N_{h_2} + N_{h_2} N_u))$, \\ $L = 2$, $N_u = 8$, $N_{h_1}=32$, $N_{h_2}=16$}} \\
GEPNet, $T = 10$ & $4.32 \times 10^{-3}$ & \\ \midrule
GDIM continuous, $T = 5$ & $7.73 \times 10^{-4}$ & \multirow{2}{*}{\makecell[l]{$\mathcal{O}(mn^2 + T n^3 + T L n^2 H^2)$, $L = 2$, $H = 32$}}\\
GDIM continuous, $T = 10$ & $1.57 \times 10^{-3}$ & \\ \midrule
GDIM discrete, $T = 10$ & $5.12 \times 10^{-3}$ & \makecell[l]{$\mathcal{O}(mn^2 + T n^3 + T L n^2 H^2)$, $L = 12$, $H = 32$} \\
\bottomrule
\end{tabular}
}
\label{tab:runtime}
\end{center}
\end{table}

\begin{figure}[t]
\centering
\centering 
\includegraphics[width=1\linewidth]{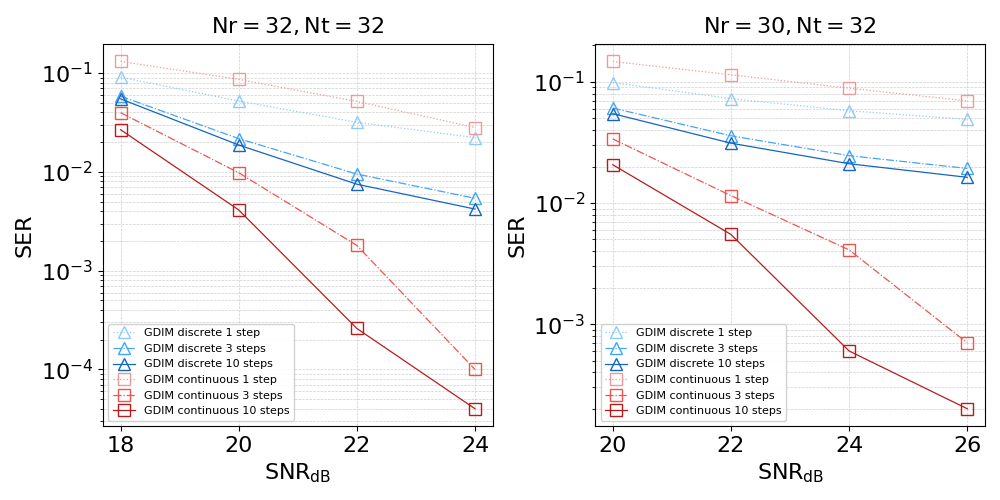} 
\caption{SER performance of discrete and continuous GDIM with different numbers of reverse denoising steps under overdetermined ($N_t = N_r = 32$) and underdetermined ($N_t = 32$, $N_r = 30$) settings with i.i.d. Gaussian channels and $16$-QAM.}
\label{fig:compare_discrete_continuous} 
\end{figure}

\begin{figure}[t]
\centering
\includegraphics[width=1\linewidth]{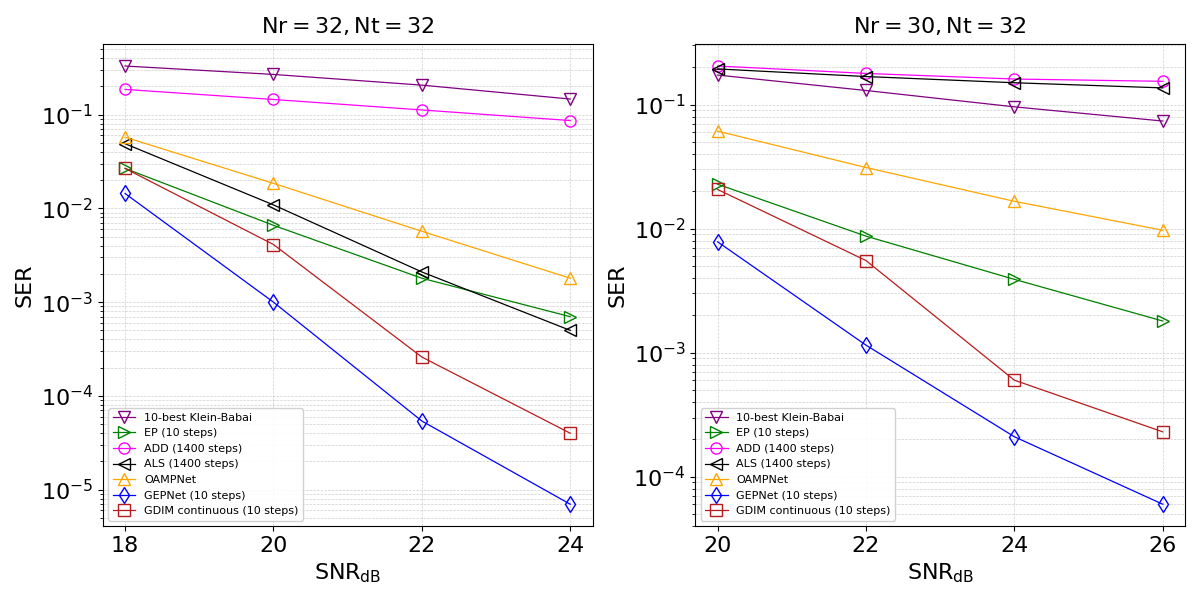}
\caption{SER performance comparison between continuous GDIM (10 denoising steps) and baseline methods for overdetermined ($N_t = N_r = 32$) and underdetermined ($N_t = 32$, $N_r = 30$) settings with i.i.d. Gaussian channels and $16$-QAM.} 
\label{fig:compare_baselines}
\end{figure}

\begin{figure}[t]
\centering
\includegraphics[width=0.9\linewidth]{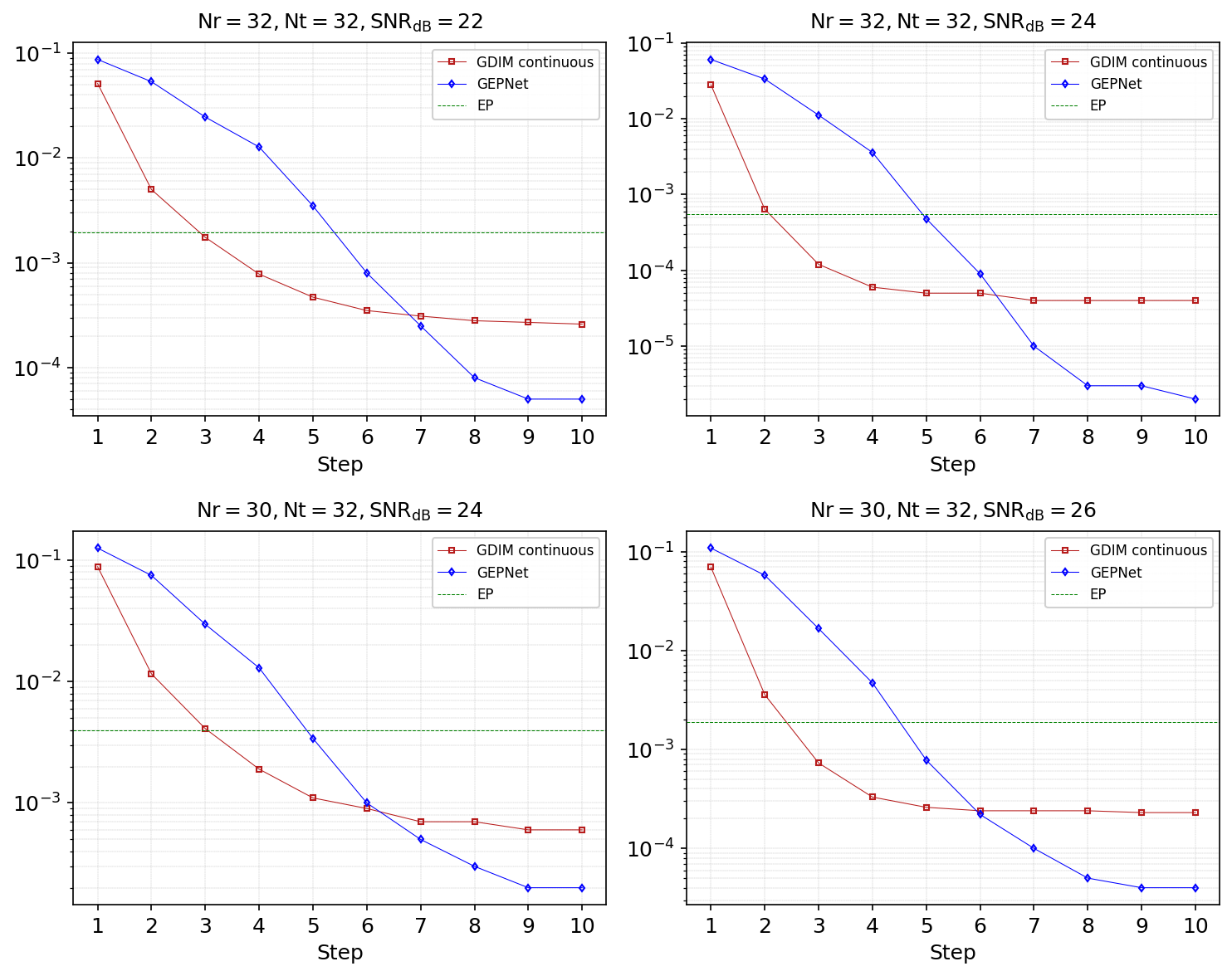}
\caption{SER performance of continuous GDIM and GEPNet at each sequential inference steps for overdetermined ($N_t = N_r = 32$) and underdetermined 
($N_t = 32$, $N_r = 30$) settings with i.i.d. Gaussian channels and $16$-QAM. 
The horizontal dashed line indicates the SER achieved by EP.} 
\label{fig:compare_steps}
\end{figure}

\subsection{Experimental Setup}
\label{sec:setup}
For each problem instance,
the entries of $\H_c$ are drawn independently from $\mathcal{CN}(0, 1/N_{r})$.
The transmitted signal $\x_{c}^{*}$ is sampled uniformly from $\widetilde{\calX}_k$ in~\eqref{eq:mimo_const}.
We set $k=2$ for $16$-QAM.
Given a target $\mathrm{SNR}_{\mathrm{db}}$ defined in~\eqref{eq:snr},
we use the corresponding $\sigma_{\noisevector_c}$ to generate $\noisevector_c$ and compute $\y_c$.
The real-valued $\x^{*}$, $\y$, $\H$ are then obtained according to the transformation described in Section~\ref{sec:pre}.

For GDIM, we train the denoising network using randomly generated problem instances. 
For the discrete formulation, 
given $\x_0 = \x^{*}$ and a diffusion step $t$, 
we use a discretized Gaussian transition matrix $\Q_t$~\cite{austin2021structured} to sample $\x_t$ according to~\eqref{eq:forward-multi}, 
where
\begin{equation*}
\Q_t(i,j) = \begin{cases} 
\frac{\exp\left(- 4|i-j|^2 / (2^k-1)^2 \beta_t \right)}{\sum_{c=1}^{2^k} \exp \left( -4c^2 / (2^k-1)^2 \beta_t \right)}, & i\neq j, \\
1 - \sum_{c \neq i} \Q_t(i,c), & i=j.
\end{cases}
\end{equation*}
We use a linear noise schedule $\{ \beta_t \}_{t=1}^T = \mathrm{linspace}(\beta_1, \beta_T, T)$ with $\beta_1 = 8.5\times 10^{-3}$, $\beta_T = 1.7\times10^{-2}$, $T=1000$.

For training the denoising network in GDIM, 
each iteration uses 32 randomly generated instances,
with $\mathrm{SNR}_{\mathrm{dB}}$ sampled uniformly from $[30, 40]$.
Training uses the Adam optimizer with learning rate $10^{-4}$ and weight decay $5 \times 10^{-5}$.
For discrete GDIM, we use hidden size $H = 32$ and $L=12$ layers.
For continuous GDIM, 
we use $H = 32$ and $L=2$ layers.
The discrete formulation requires significantly more layers in the denoising network for effective training,
since the network is conditioned only on the noisy discrete state at a given diffusion step.
In contrast, 
the continuous formulation uses the likelihood-informed inputs $\bmu_t$ and $\diag(\bSigma_t)$, 
which provide a refined estimate of the clean signal with coordinate-wise uncertainty.
Thus, the denoising network in continuous GDIM starts from a more informative representation, 
allowing it to use a less complex architecture with fewer layers.

We next describe the implementation details of the baseline methods.
For classical estimators,
we use $K$-best Klein-Babai with $K=10$ and EP with $10$ iterations.
ALD~\cite{zilberstein2022annealed}, 
a massive MIMO diffusion detector based on annealed Langevin dynamics,
uses 20 noise levels and 70 sampling steps per noise level, 
resulting in 1400 sequential sampling iterations in total, 
following~\cite{zilberstein2022annealed}.
We also use 1400 sequential sampling iterations for ADD, 
which is an approximate diffusion detector proposed as an SVD-free alternative to ALD~\cite{he2024massive}.
For OAMPNet, 
we use 10 layers and follow the training procedure proposed in~\cite{he2018model}.
For GEPNet, 
we use 10 layers as proposed in~\cite{kosasih2022graph},
with hidden size $32$, 
which is the same hidden size used in GDIM.

For both GDIM and GEPNet, 
we train the networks for 12 hours on the Rorqual cluster of The Digital Research Alliance of Canada using one H100 MIG GPU instance with 20GB GPU memory.
For GEPNet, we use a validation set with 2000 samples to select the best network.
For GDIM, we do not observe overfitting and therefore use the network from the last training iteration for evaluation.

For testing, 
all methods are evaluated on $10^5$ randomly generated instances for each $\mathrm{SNR}_{\mathrm{dB}}$.
In the underdetermined case, 
the $K$-best randomized Babai estimator is applied to the regularized ILS problem~\eqref{eq:ils-transformed-l2-reg}, 
as regularization is required for this method.
The proposed GDIM and other baseline methods are applied without adding regularization in all cases.
For methods that require training, 
including OAMPNet, GEPNet, and GDIM,
training is conducted on the overdetermined setting with $N_t = N_r = 32$,
and testing is conducted on both overdetermined ($N_t = N_r = 32$) and underdetermined settings ($N_t = 32$, $N_r = 30$).

\subsection{Complexity and Runtime Analysis}
\label{sec:runtime}
Table~\ref{tab:runtime} reports the average per-instance detection runtime and the corresponding computational complexity, 
where $m = 2 N_r$, $n = 2 N_t$.
All methods are tested on an NVIDIA GeForce RTX 4090 GPU with CUDA acceleration.
For the $K$-best Klein-Babai, 
the $K$ candidates are produced in parallel.
In all experiments, 
we use a batch size of $32$, and the samples within each batch are processed in parallel.

For GEPNet and the proposed GDIM, 
we report the runtime using both 5 and 10 steps. 
Here, the term ``step'' follows the definition of each model. 
For GEPNet, one step corresponds to one GNN-aided EP iteration. 
For GDIM, one step corresponds to one reverse denoising step. 
Table~\ref{tab:runtime} shows that continuous GDIM achieves a lower average per-step runtime than GEPNet.


\subsection{Performance Comparison}
\label{sec:gdim_compare_baselines}




We first compare the discrete and continuous formulations of the proposed GDIM.
Figure~\ref{fig:compare_discrete_continuous} compares their SER performance under different numbers of reverse denoising steps.
In both overdetermined and underdetermined settings under varying SNR conditions,
continuous GDIM with 3 denoising steps achieves lower SER than discrete GDIM with 10 steps.
Since discrete GDIM also evaluates a substantially deeper denoising network at each step, 
as described in Section~\ref{sec:setup}, 
continuous GDIM achieves better performance with lower inference cost.

Since the denoising networks in both discrete and continuous GDIM are trained to predict the clean signal $\x_0$ satisfying the discrete constraint $\x_0\in\calX_k^n$, 
the main difference between the two formulations lies in whether the noisy state are constrained to remain in the discrete space during the denoising process.
These results suggest that,
for GDIM, constraining the noisy state to remain in the discrete constraint space is not necessary for effective denoising.
Instead, the continuous forward process enables a closed-form Gaussian approximation of the posterior distribution of the clean signal given the noisy latent state and the MIMO observation.
The resulting likelihood-informed statistics provide a more informative representation for the denoising network, 
making $\x_0$ prediction easier than using the noisy discrete state directly.
Therefore, the subsequent comparisons focus only on continuous GDIM.

Next, Figure~\ref{fig:compare_baselines} compares continuous GDIM with 10 denoising steps against the baseline methods.
We make the following key observations.
(1) GDIM and GEPNet achieve the best overall detection performance among all tested methods, while EP ranks third.
This indicates that learned graph-based refinement is effective for MIMO detection.
GEPNet achieves better performance than GDIM when both methods use 10 steps and have comparable runtime.
This advantage can be attributed to the difference in training objectives.
GEPNet is trained end-to-end with 10 sequential refinement steps and optimizes the loss over the full refinement trajectory.
This allows the network to learn step-to-step dependencies and coordinate its updates globally across all steps.
In contrast, as a diffusion-based method,
GDIM trains the denoising network to predict the clean signal from noisy state information at individual diffusion steps.
It does not directly supervise how errors made at earlier denoising steps should be corrected in subsequent steps.
Nevertheless, GDIM offers a favorable accuracy--runtime trade-off over GEPNet, as discussed later.
(2) The existing diffusion-based methods, 
namely ALD and ADD, require many sequential iterations and have substantially longer runtime, 
as shown in Table~\ref{tab:runtime}.
However, their detection performance falls behind EP.
Furthermore, both ALD and ADD exhibit performance degradation in the underdetermined setting, whereas the proposed GDIM remains robust.


To compare GDIM and GEPNet,
Figure~\ref{fig:compare_steps} traces the SER performance achieved after each sequential inference step under each configuration,
while Table~\ref{tab:runtime} shows that 
continuous GDIM has a lower average runtime per step than GEPNet.
Together, these results show how efficiently each method converts additional sequential computation into detection performance gains.
Figure~\ref{fig:compare_steps} shows that GDIM achieves better detection performance than GEPNet in the early iterations.
Moreover, GDIM at the $5^{th}$ denoising steps already outperforms EP while requiring lower runtime, 
in each configuration.
In contrast, GEPNet requires more iterations at the cost of higher inference time to surpass EP.
Thus, although GEPNet achieves stronger final-step performance, 
GDIM provides a more favorable accuracy--runtime trade-off by outperforming the baseline methods with lower runtime.

\section{Conclusion}

This chapter proposed GDIM, 
a graph-based denoising diffusion method for MIMO detection.
We investigated both the discrete diffusion with multinomial noise and the continuous diffusion with Gaussian noise for GDIM.
The discrete one keeps the latent variables constrained to the constellation set,
while the continuous formulation relaxes this constraint.
With the continuous formulation,
likelihood-informed posterior statistics can be derived from the continuous latent state and the problem instances.
These statistics provide a more informative input to the denoising network and make the clean-signal prediction task easier.
The numerical results show that continuous GDIM outperforms discrete GDIM while requiring a shallower denoising network and lower inference cost.

Compared with baseline methods, 
continuous GDIM achieves strong performance in both overdetermined and underdetermined settings, addressing some limitations of existing methods.
Unlike existing diffusion-based methods that rely on analytic posterior score-based sampling, 
GDIM parameterizes the reverse process with a graph-based denoising network and generalizes better to underdetermined MIMO detection tasks.
Continuous GDIM also improves the accuracy--runtime trade-off by allowing a flexible number of denoising steps during inference.

%% file: content/Conclusions.tex
\chapter{\rm\bfseries Conclusion and Future Work}
\label{ch:conclusions}

\section{Conclusion}

This thesis investigates graph representation learning for node classification,
learning to optimize approaches to integer-valued parameter estimation in linear models,
and the intersection of these two areas.

First, Chapter~\ref{ch:Paper02} proposes a parameterized graph Laplacian that enables flexible graph message passing scopes.
Based on this Laplacian,
PD-GCN and PD-GAT are developed, 
together with a topology-guided rewiring strategy.
The proposed methods outperform SOTA baselines on six real-world benchmark datasets. 

Second, using machine learning,
this thesis proposes methods that closely approximate quantities that are computationally infeasible to obtain in integer-valued parameter estimation problems.
Chapter~\ref{ch:Paper01} extends the Babai estimator and learns a structured latent space to estimate and optimize its success probability efficiently.
For the estimation problem,
this thesis further leverages the underlying densely connected graph structure and applies graph representation learning techniques to address limitations of existing estimators.
Based on a graph convolution form of the linear model,
Chapter~\ref{ch:Paper03} proposes GCEPNet,
which significantly reduces the computational bottleneck caused by the GNN module in the SOTA estimator GEPNet,
while achieving comparable or better performance.
Chapter~\ref{ch:Paper04} proposes a graph-based denoising diffusion method that offers a better accuracy--runtime trade-off and addresses the performance degradation observed in existing diffusion-based methods for underdetermined estimation problems.

\section{Future Directions}

This thesis addresses some key limitations in graph representation learning 
and develops learning-based approaches for optimization problems arising in integer valued parameter estimation in linear models,
with particular emphasis on graph representation learning-based methods.
Despite these advances,
several important questions remain unanswered.
In this section,
we outline future research directions that remain open for further exploration.

First, 
new methods that leverage the parameterized normalized Laplacian proposed in Chapter~\ref{ch:Paper02} deserve further investigation.
Recent studies have shown new progress in spectral GNNs, 
where the graph Laplacian plays a central role~\cite{xu2024revisiting, zheng2025dynamic, ezoe2025unfolded, hariri2026return, yang2025disentangled, azizi2026spectral}.
Methods based on spectral graph theory
have been developed for a wide range of tasks,
including long-range information modeling~\cite{hariri2026return},
link prediction~\cite{azizi2026spectral},
graph anomaly detection~\cite{xu2024revisiting, zheng2025dynamic},
representation learning on time varying graphs~\cite{ezoe2025unfolded},
and graph domain adaptation~\cite{yang2025disentangled}.
These methods are built on commonly used fixed graph Laplacian matrices, as introduced in Section~\ref{sec:pre_grl}.
Therefore, a promising future direction is to extend them using the proposed parameterized Laplacian.

Second, for learning-based approaches to approximating optimal solutions to NP-hard estimation problems,
although methods presented in Chapters~\ref{ch:Paper01},~\ref{ch:Paper03}, and~\ref{ch:Paper04} improve existing approaches from various perspectives,
the optimization problems studied in these chapters follow those considered in previous research and are relatively simple compared with real-world problems.
Specifically, 
for the integer parameter estimation problem in a linear model defined in~\eqref{eq:linear-model},
we consider a setting where the model matrix $\H$ is known and the noise follows a known Gaussian distribution $\calN(\0,\sigma_{\noisevector}^2\I)$.
While this setting provides a useful starting point for studying the broader potential of learning-based optimization and allows for straightforward comparison with existing works,
it remains to be explored how these methods can be extended to more complex settings,
where the noise may be non-Gaussian or correlated~\cite{zhang2000adaptive},
or where the model matrix may contain errors~\cite{truong2013effects}.